\documentclass{article} 
\usepackage{iclr2027_conference,times}

\usepackage{amsmath,amsfonts,bm}

\def\eqref#1{equation~\ref{#1}}

\def\1{\bm{1}}

\def\ra{{\textnormal{a}}}

\def\rx{{\textnormal{x}}}

\def\rva{{\mathbf{a}}}

\def\erva{{\textnormal{a}}}

\def\ervx{{\textnormal{x}}}

\def\rmA{{\mathbf{A}}}

\def\vmu{{\bm{\mu}}}
\def\vtheta{{\bm{\theta}}}
\def\va{{\bm{a}}}

\def\ve{{\bm{e}}}

\def\vx{{\bm{x}}}

\def\eva{{a}}

\def\mA{{\bm{A}}}

\def\mH{{\bm{H}}}
\def\mI{{\bm{I}}}
\def\mJ{{\bm{J}}}

\def\mX{{\bm{X}}}

\def\mSigma{{\bm{\Sigma}}}

\DeclareMathAlphabet{\mathsfit}{\encodingdefault}{\sfdefault}{m}{sl}
\SetMathAlphabet{\mathsfit}{bold}{\encodingdefault}{\sfdefault}{bx}{n}
\newcommand{\tens}[1]{\bm{\mathsfit{#1}}}
\def\tA{{\tens{A}}}

\def\tX{{\tens{X}}}

\def\gG{{\mathcal{G}}}

\def\sA{{\mathbb{A}}}
\def\sB{{\mathbb{B}}}

\def\sS{{\mathbb{S}}}

\def\emA{{A}}

\newcommand{\etens}[1]{\mathsfit{#1}}

\def\etA{{\etens{A}}}

\newcommand{\E}{\mathbb{E}}

\newcommand{\R}{\mathbb{R}}

\newcommand{\KL}{D_{\mathrm{KL}}}
\newcommand{\Var}{\mathrm{Var}}

\newcommand{\Cov}{\mathrm{Cov}}

\newcommand{\normltwo}{L^2}
\newcommand{\normlp}{L^p}

\newcommand{\parents}{Pa} 

\usepackage{hyperref}
\usepackage{comment}
\usepackage{url}
\usepackage{graphicx}
\usepackage{booktabs}
\usepackage{amsthm}
\usepackage{amsfonts}
\usepackage{amsmath,amssymb}
\usepackage{tikz-cd}
\usepackage{multicol}
\definecolor{orangehighlight}{RGB}{255, 230, 200} 

\usepackage{wrapfig}
\usepackage{graphicx}
\usepackage{subcaption}
\usepackage{mathtools}
\usepackage{enumitem}
\usepackage[utf8]{inputenc}
\usepackage[T1]{fontenc}
\usepackage{amsmath, amssymb}
\usepackage[skins]{tcolorbox}
\tcbuselibrary{skins}
\usepackage{graphicx}
\usepackage{forest}
\usepackage{bm}
\usepackage[skins]{tcolorbox}
\usepackage{tikz}
\usetikzlibrary{arrows.meta} 
\usetikzlibrary{decorations.pathmorphing, backgrounds, calc, shapes.arrows}
\definecolor{elegantBlue}{RGB}{65, 105, 225} 
\definecolor{elegantOrange}{RGB}{237, 125, 49} 
\definecolor{acadblue}{HTML}{1F4E79}
\definecolor{tealcyan}{HTML}{00A0B0}

\usepackage[utf8]{inputenc} 
\usepackage[T1]{fontenc}    
\usepackage{hyperref}       
\usepackage{url}            
\usepackage{booktabs}       
\usepackage{amsfonts}       
\usepackage{nicefrac}       
\usepackage{microtype}      
\usepackage{xcolor}         
\usepackage{tcolorbox}
\tcbuselibrary{skins} 

\title{Sampling Meets Interaction: Sequentially-Controlled Multi-Particle Flow-Maps for efficient Inference-time Search}

\author{
    Binglin Ji$^{1}$\thanks{Authors Made Equal and Core Contributions to This Paper.} ,~~~Anindya Sarkar$^{1}$\footnotemark[1] ,~~~~Hengchang Lu$^{1}$,~~~~Jens Sjölund$^{2}$,~~~~Yevgeniy Vorobeychik$^{1}$ \\
    \texttt{\{binglin.j,~anindya,~yvorobeychik\}@wustl.edu,}\\${}^1$Department of CSE, Washington University in St.Louis, USA\\
    ${}^2$Department of Information Technology, Uppsala University, Sweden}

\iclrfinalcopy 
\begin{document}

\maketitle

\begin{abstract}
While generative models have enabled training-free reward alignment, existing particle-based methods are fundamentally constrained by their propensity 
to local exploration within narrow regions of the underlying distribution, severely restricting sample diversity.
 This limitation becomes especially acute under tight reward-feedback budgets, where effective search demands broad, strategic exploration to uncover high-utility regions. To address this, we propose Sequentially-Controlled \textbf{I}nteractive \textbf{M}ulti-\textbf{P}article \textbf{F}low-\textbf{M}aps (\textbf{IMPFM}), a framework for feedback-efficient search. IMPFM progressively transports a group of interactive particles toward the target distribution, maintaining the broad coverage essential for heterogeneous preference alignment. IMPFM leverages a principled and efficient posterior sample-sharing mechanism across particles powered by flow maps. By correcting individual particle drift with the collective value gradient from the entire ensemble's posterior samples at each correction step, the framework maximizes sample utility to enable global exploration while actively mitigating reward over-optimization, typical of standard control frameworks. 
  Paired with a principled exploration-exploitation reweighting mechanism involving multi-particle interaction, this sequentially corrected multi-particle dynamics explicitly preserves structural diversity and overcomes the weight degeneracy inherent to standard Sequential Monte Carlo (SMC) samplers.
  Crucially, we prove that the resulting sampling framework yields a multi-particle interaction-aware Feynman-Kac corrector that progressively steers the multi-particle system toward a KL-tilted target distribution, facilitating global exploration and preventing mode collapse. Extensive empirical evaluations and ablations across diverse search and alignment tasks confirm the efficacy of IMPFM over existing baselines.
\end{abstract}
\vspace{-10pt}
\vspace{-2pt}
\section{Introduction}
\vspace{-10pt}
Across diverse scientific and engineering disciplines, the pace of discovery is fundamentally constrained by the high cost of evaluation. This bottleneck gives rise to the ubiquitous challenge of feedback-efficient search, where finding an optimal solution requires iteratively proposing candidates and adapting based on sequential feedback. Whether designing life-saving therapeutics or optimizing interactive visual recommendation systems, these real-world applications often rely on costly physical experiments or human engagement.
Consequently, minimizing the number of such interactions is a central requirement.
Accelerating breakthroughs in these domains demands a robust sampling framework capable of strategic, global exploration across complex, high-dimensional spaces—such as natural images. 

Existing methods remain fundamentally ill-equipped for this challenge. For instance, online RL fine-tuning of diffusion models~\citep{uehara2024feedback} relies on a mode-seeking KL-divergence objective that inherently fails to capture diverse, high-utility regions. Furthermore, these approaches depend on online-trained reward models, whose early-stage biases frequently misdirect fine-tuning and degrade sample quality. Conversely, recent Sequential Monte Carlo (SMC) inference-time search methods~\citep{singhal2025general,kim2025test} avoid fine-tuning by leveraging fixed generative priors. 
Yet, they suffer from severe weight degeneracy, leading to proposal collapse and diminished diversity~\citep{lee2025debiasing}. 
Moreover, existing SMC-based samplers rely rigidly on prior generative dynamics, failing to actively correct search trajectories using sequential feedback. 
This passive exploration inevitably yields suboptimal performance. Recently, Feynman-Kac correctors~\citep{skreta2025feynman} (FKCs) have emerged as a powerful alternative—progressively correcting dynamics at every resampling step—making them a highly promising paradigm for rapid, feedback-driven adaptation.
\begin{wrapfigure}{r}{0.61\textwidth}
    \centering
    \includegraphics[width=0.61\textwidth]{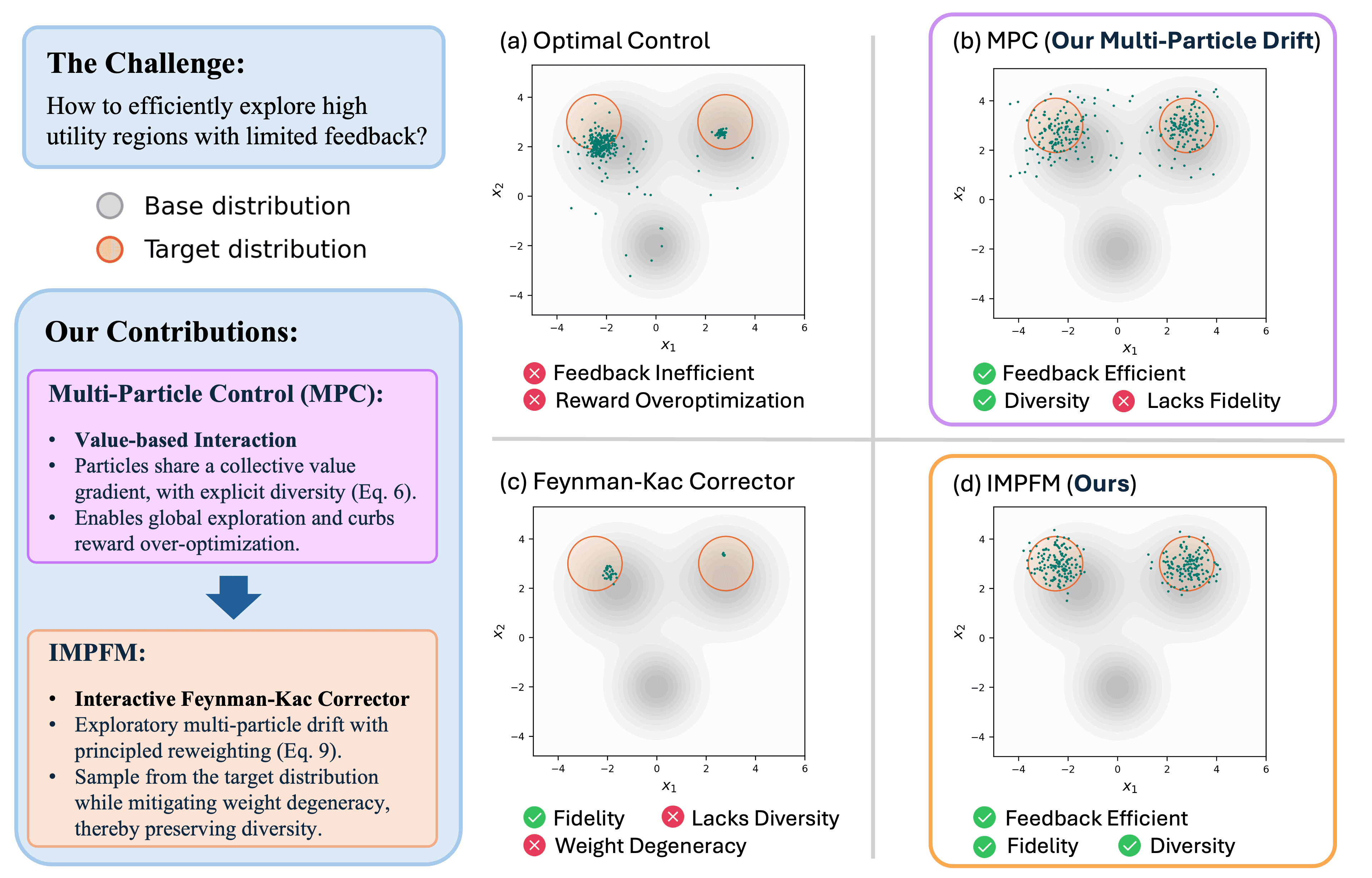}
    \vspace{-5pt}
    \caption{Conceptual Overview of IMPFM.}
    \label{fig:intro-overview}
    \vspace{-10pt}
\end{wrapfigure}
However, these approaches rely on reward values or gradients at every resampling step; in practice, these signals are unavailable and must be approximated, introducing systematic bias into the sampler and degrading sample quality. 
  Crucially, standard FKC samplers transport particles in strict isolation. This lack of inter-particle interaction severely bottlenecks broad strategic exploration.
To overcome these bottlenecks, we introduce \emph{Interactive Multi-Particle
Flow-Maps (IMPFM)}, a principled, feedback-efficient sampling framework that ensures diverse coverage across high-reward regions. 
Figure~\ref{fig:intro-overview} provides an overview of IMPFM.

\vspace{-6pt}
\begin{tcolorbox}[colback=blue!5!white, colframe=gray!80!black, title={\textbf{Key Idea behind IMPFM}}]
IMPFM is built on the premise that search thrives not in isolation, but through collective intelligence. It recognizes that collaboration is the ultimate engine for expanding information diversity and unlocking truly sample-efficient search.
\end{tcolorbox}
\vspace{-5pt}
At the heart of IMPFM is a value-based particle interaction mechanism that drives a dual-force exploration dynamic, combining an attractive pull toward high-utility regions with a repulsive push that prevents mode collapse. Powered by Flow Maps, IMPFM enables each particle to exploit the ensemble’s collective posterior to optimally correct its drift at every step, and this information reuse mechanism facilitates highly informed, feedback-efficient search. 
IMPFM’s power is substantially enhanced when paired with a principled reweighting mechanism at each resampling step. This reweighting encodes rich value-based multi-particle interactions and enforces a dual-force dynamic—attracting particles to high-density targets while simultaneously exerting a repulsive push to prevent mode collapse and tame reward overoptimization, the classic pitfall of reweighting-based samplers. Critically, we show that IMPFM induces interaction-aware Feynman–Kac correctors that rapidly correct the search dynamics based on sequential feedback and drive the particle system toward the KL-tilted marginal distribution, with the true target as its reference measure.
These findings show that, \emph{rather than treating samples independently, it is more powerful to reason about the particle ensemble as a whole and minimize a distributional distance. This ensemble-level view provides a broader picture, promotes global convergence, and yields substantially wider coverage.}
\vspace{-7pt}
\begin{tcolorbox}[colback=blue!5!white, colframe=gray!80!black, title={\textbf{In summary, we make the following contributions:}}]
\begin{itemize}[itemsep=0pt,topsep=0pt, leftmargin=*]
\item \textbf{Interactive FKC Sampler}: We introduce IMPFM, a Multi-Particle Interaction-aware Feynman–Kac Corrector for feedback-efficient search.
\item \textbf{Flow-Map-Driven Information Reuse:} 
IMPFM leverages a collective posterior-sharing mechanism powered by Flow Maps, maximizing the utility of every drawn sample, transforming isolated particle updates into a highly informed, globally coordinated search. 
\item \textbf{Rapid Adaptation and Coverage}: Driven by value-based dual-force (attractive-repulsive) exploration dynamics, and by coupling collaborative drift correction with dynamic reweighting, IMPFM enables expansive search-space coverage. 
\item \textbf{Rigorous Empirical Validation}: We validate the effectiveness of each component of IMPFM through comprehensive quantitative and qualitative ablation studies.
\end{itemize}
\end{tcolorbox}
\vspace{-15pt}

\vspace{-1pt}
\section{Related Work}\label{sec:rel_work}
\vspace{-4pt}
\paragraph{Multi-Particle Interaction based Transport}
Stein Variational Gradient Descent (SVGD)~\citep{liu2016stein,liu2017stein} transports interacting particles to match a target distribution. However, it relies on iterative score function evaluations for each particle; approximating these scores introduces biases at every iteration that severely limit its application in high-dimensional settings. Furthermore, SVGD is not designed for reward-tilted distributions, which would additionally require value gradient evaluation of each constituent particle at every step. In stark contrast, our approach entirely bypasses score computation. We introduce a purely value-based particle interaction mechanism—efficiently computed via flow maps—that uses particle interaction to augment exploitative particle drift with broad exploration. While our interactive dynamics target a Boltzmann distribution, we seamlessly recover the true reward-tilted distribution through a principled reweighting mechanism.

\vspace{-6pt}
\paragraph{Inference-time Scaling for Reward Alignment:}
Inference-time scaling has garnered significant attention recently, with particle-based SMC emerging as a highly versatile approach that maintains a population of samples to approximate the target distribution.
SMC~\citep{del2006sequential,phillips2024particle,cardoso2023monte,singhal2025general,kim2025test} leverages potential functions, which usually approximate the soft value function, to assign weights to particles and resample them at every step. While numerous adaptations have been developed specifically for diffusion model alignment~\citep{dou2024diffusion,cardoso2023monte,kim2025test,trippe2022diffusion,wu2023practical}, practical implementations face significant hurdles. Although SMC provides theoretical guarantees of exact sampling in the infinite-particle limit, practical implementations are hindered by weight degeneracy and inaccurate value estimates. These issues often induce mode collapse and compromise sample diversity.
Recently, \citep{holderrieth2026diamond} proposed Diamond Maps, which utilizes sufficient statistics~\citep{holderrieth2025glass} to emulate DDPM-style stochasticity within an ODE-based flow model, thereby facilitating an SMC-like sampling procedure. However, this method inherits the same fundamental limitations as traditional SMC.
Recently, several FKC techniques~\citep{skreta2025feynman,lee2025debiasing} have been introduced to alleviate particle weight degeneracy and debias particle-based guidance. However, these methods struggle in high-dimensional domains like images. Because they rely on reward gradients at every resampling step—which are typically unavailable and require approximation—they inadvertently inject systematic bias into the sampling process.
Although numerous gradient-free alignment methods avoid computing reward gradients by employing BON sampling~\citep{stiennon2020learning,beirami1879theoretical,nakano2022webgpt}, this generate-and-filter approach introduces substantial computational overhead. Furthermore, because these methods only intervene post-hoc at the final output stage, they cannot actively guide the intermediate generative trajectory or fundamentally alter the underlying dynamics.
An extended discussion on Related work is provided in the Appendix~\ref{sec:ex-rel}.
\vspace{-6pt}
\section{Method: IMPFM Framework}
\vspace{-10pt}
\textbf{Background:} The cornerstone of feedback-efficient search is a policy capable of efficiently sampling from the look-ahead value-tilted distribution. Formally, under optimal control, the marginal density $p_t^*$ of $X_t^*$ at timestep $t$ satisfies:
\vspace{-5pt}
\begin{equation}
    \small{p_t^*(x) \propto p_t(x) e^{V_t(x)}, \quad \text{where} \quad V_t(x) = \log \mathbb{E} \left[ e^{r(X_1)} \mid X_t = x \right]}
\end{equation}
Here, $p_t$ denotes the marginal density associated with the unconditional prior process. This formulation ensures that the terminal marginal distribution, $p_1^*$, aligns exactly with the target reward-tilted distribution $p_{\text{reward}}(x) \propto p_1(x) e^{r(x)}$. Leveraging Doob’s $h$-transform \citep{denker2024deft,dai1991stochastic}, the optimal drift $\smash{b_t^*}$ for a diffusion schedule $\sigma_t$ can be expressed as:
\vspace{-5pt}
\begin{equation}\label{eq:oc}
    \small{\dot{x}_t^* = \underbrace{b_t(x_t^*) + \frac{\sigma_t^2}{2} \nabla V_t(x_t^*)}_{b_t^*(x_t^*)}, \quad \text{with} \quad \text{Law}(x_t^*) = p_t^*, \quad x_0^* \sim p_0}
\end{equation}
While $b_t$ is directly accessible from the pre-trained Flow Matching model, the primary algorithmic bottleneck lies in efficiently estimating the value function gradient, $\nabla V_t(x)$. A consistent Monte Carlo estimator of $\nabla V_t(x)$ is: 
\vspace{-5pt}
\begin{equation}\label{eq:grad_v}
    \small{\nabla_{x_t} V_t(x_t) = \nabla_{x_t} \log \left( \frac{1}{N} \sum_{i=1}^N \exp \left[ r(\hat{x}^i_1) \right] \right), \quad \tcbox[on line, boxsep=0pt, left=2pt, right=2pt, top=2pt, bottom=2pt, colback=blue!10, colframe=gray!50, arc=2pt]{$\hat{x}^i_1 \overset{\text{iid}} {\sim} p_{1|t}(\cdot|x_t)$}}
\end{equation}
Hence, the primary challenge boils down to efficiently drawing samples from $p_{1|t}(\cdot|x_t)$ via an ODE-based flow model.
Conditioned on $x_t$, one can define an ODE that transports the prior $p_0$ directly to the posterior $p_{1|t}(\cdot|x_t)$. The associated drift field, $b_s(\cdot;x_t)$, is derived via standard flow matching~\citep{lipman2022flow}, with the crucial distinction that the objective targets the conditional posterior $p_{1|t}(\cdot|x_t)$ rather than the marginal data distribution $p_1$:
\vspace{-5pt}
$$\small{{b}_s(x; x_t) = \mathbb{E}\left[\dot{\alpha}_sx_1 + \dot{\beta}_sx_0 \mid x_s = x\right], \quad x_s = \alpha_sx_1 + \beta_sx_0, \quad x_0 \sim \mathcal{N}(0, I_d) , \quad x_1 \sim p_{1|t}(\cdot|x_t)} $$
Consequently, the probability flow associated with 
$b_s(\cdot; x_t)$ satisfies:
\vspace{-4pt}
\begin{equation}
\small{\frac{d}{ds}x_s = b_s(x_s; x_t), \quad x_0 \sim \mathcal{N}(0, I_d) \implies \text{Law}(x_1) = p_{1|t}(\cdot|x_t)}
\label{eq:ode_cond}
\end{equation}
 Where $\alpha_t$ and $\beta_t$ denote time-dependent coefficients with boundary conditions $\alpha_0 = \beta_1 = 0$ and $\alpha_1 = \beta_0 = 1$. Under the assumption of a Gaussian prior $p_0$, the conditional drift $b_s$ admits an analytical derivation~\citep{holderrieth2025glass} via a reparameterization of the unconditional drift $b_t$:
\vspace{-2pt}
$$\small{b_s(x_s; x_t) = w_1x_s + w_2 b_{t^*}(S(x_s, x_t)) + w_3 x_t}$$
Where, $w_1, w_2,$ and $w_3$ represent scalar weights. Furthermore, $S$ acts as a linear \emph{sufficient statistic}, while $t^*$ represents a reparameterized time variable:
\vspace{-3pt}
\begin{equation}
\small{S_{s,t}(x_s, x_t) = \frac{\alpha_s \sigma^2_t x_s + \alpha_t \sigma^2_s x_t}{\sigma^2_t \alpha^2_s + \alpha^2_t \sigma^2_s}, \qquad t^*(s, t) = g^{-1}\left( \frac{\sigma^2_t \sigma^2_s}{\sigma^2_t \alpha^2_s + \alpha^2_t \sigma^2_s} \right), \quad g(t) = \frac{\sigma^2_t}{\alpha^2_t}}
\label{eq:combined_reparam}
\end{equation}
While this reparameterization renders the conditional drift $b_s$ tractable, integrating the ODE trajectories as defined in~\ref{eq:ode_cond} to draw posterior samples from $p_{1|t}(\cdot|x_t)$ remains a significant computational bottleneck.
To bypass this computational bottleneck, a standard strategy is \emph{teacher-distillation}, which leverages the exact analytical field $b_s(x_s; x_t)$ as the teacher for the distillation process. The student velocity model, $\hat{v}$, is optimized via a joint objective function. The first component is an \emph{instantaneous loss} that minimizes the mean squared error between the student's predicted velocity, $\hat{v}_{s,s}$, and the exact analytical teacher field:
\vspace{-5pt}
$$\small{\mathcal{L}_{\text{inst}}^{\text{distill}}(\hat{v}) := \int_0^1 \int_0^1 \mathbb{E} ||\hat{v}_{s,s}(x; x_t) - b_s(x; x_t)||^2 ds dt}$$
This is paired with a \emph{consistency loss}~\citep{boffi2025build} ($\mathcal{L}^{\text{distill}}_{\text{cons}}$) to ensure the student reliably integrates this velocity across arbitrary time intervals, defined mathematically as:
\vspace{-5pt}
\[
    \small{\int_{0}^{1} \int_{0}^{u} \mathbb{E} \left\| \hat{v}_{s,u}(I_{s}; I'_t) - \text{sg}\left( b_u (\hat{X}_{s,u}(I_{s}; I'_t)); I'_t \right) - (u - s)\partial_u \hat{v}_{s,u}(I_{s}; I'_t) \right\|^2 ds \, du}
\]
Here, $\hat{X}_{s,u}(\cdot \, ; x_t) : \mathbb{R}^d \to \mathbb{R}^d$ denotes the solution operator for the context-dependent ODE defined in~\ref{eq:ode_cond}. This distillation process effectively collapses the iterative, multi-step ODE integration into a single amortized map via $\hat{v}$, \emph{enabling computationally efficient, one-step sampling directly from the conditional posterior} $p_{1|t}(\cdot|x_t)$, thereby streamlining the evaluation of $\nabla V_t(x)$ as defined in~\ref{eq:grad_v}. 
\vspace{-4pt}
\paragraph{Background on FKC:} We present the background on the FKC sampler in the Appendix~\ref{sec:fkc-back}.

\textbf{Proposed Approach:} Merely repeatedly simulating the optimal drift $b_t^*$ is inadequate for efficient feedback-driven search: solely value-guided trajectories are inherently exploitative and lack an explicit exploratory mechanism, causing them to miss disjoint, high-utility modes within the target landscape. Furthermore, this approach is highly vulnerable to reward over-optimization unless stabilized by an impractically large ensemble of Monte Carlo posterior samples, forcing an untenable trade-off under strict sampling budgets. In particular,~\cite{potaptchik2026meta} shows that the distance to the true target distribution is bounded by the inverse of the number of Monte Carlo posterior samples utilized to compute $\nabla V_t$ at each drift correction step.
To resolve this, motivated by~\citep{liu2016stein}, we propose an interactive multi-particle drift correction mechanism that integrates the value gradients from the posterior samples across an ensemble. This introduces an explicit exploration mechanism to augment the purely exploitative optimal drift, $b_t^*$. By pooling this global information to calibrate individual particle drift, our framework drives broad exploration and prevents reward over-optimization. Specifically, the dynamics of the $i$-th particle are steered by the following rule:
\begin{equation}\label{eq:dynamics}
    \small{\dot{x}_t^i = b_t(x_t^{i}) + \underbrace{\frac{\sigma_t^2}{2} \nabla_{x^i_t} V_t(x_t^{i})}_{\text{\textbf{Exploitation}}} + \tcbox[on line, boxsep=0pt, left=2pt, right=2pt, top=5pt, bottom=5pt, colback=blue!10, colframe=gray!50, arc=2pt]{$\displaystyle \underbrace{\frac{\sigma_t^2}{2}\frac{1}{n-1} \sum_{j=1, j \neq i}^n \left[ \underbrace{k(x^j_t, x^{i}_t) \nabla_{x^j_t} V(x^j_t)}_{\text{Pulling towards high utility region}} + \underbrace{\nabla_{x^j_t} k(x^j_t, x^{i}_t) }_{\text{Repulsive force induce diversity}}\right]}_{\text{\textbf{Exploration}: Interaction between particles encourage broad coverage of high utility regions}}$}}
\end{equation}
Here, $n$ represents the number of particles, $k(x^i_t, x^j_t)$ denotes the Radial Basis Function (RBF) kernel, defined as $k(x^i_t, x^j_t) = \exp \left( -\frac{1}{h} \| x^i_t - x^j_t \|_2^2 \right)$, where $h$ is the bandwidth parameter, and $\nabla V_t(x)$ is efficiently computed via Flow-Map following ~\ref{eq:grad_v}. We refer to the resulting framework as Multi-Particle Control (i.e., MPC). In Proposition 4.2 below, we demonstrate that the interplay between the exploitation and exploration terms in Eq.~\ref{eq:dynamics} induces a joint steering of the particle dynamics that achieves a steepest descent in the KL divergence between the empirical particle distribution and the Boltzmann distribution, where the value $V(x)$ serves as the energy function.  
\begin{tcolorbox}[colback=blue!5!white, colframe=gray!80!black, title={\textbf{Proposition 4.2} (Exploration-Augmented Drift Correction via Interactive Particle Dynamics)}]
Let $q(x)$ denote the empirical distribution induced by the interacting particles, and define the Boltzmann distribution as $p_{\text{boltz}}(x) \propto e^{V(x)}$. The steepest descent direction that minimizes the Kullback-Leibler divergence, i.e., $\nabla_x D_{\text{KL}}(q \parallel p_{\text{boltz}})$ within the unit ball of a Reproducing Kernel Hilbert Space (RKHS), is given by:
\vspace{-5pt}
\begin{equation}\label{eq:stein_}
    \small{\mathbb{E}_{x^i \sim q} \left[ \nabla_{x^i} V(x^{i}) + \frac{1}{n-1} \sum_{j=1, j \neq i}^n \left[ k(x^j, x^{i}) \nabla_{x^j} V(x^j) + \nabla_{x^j} k(x^j, x^{i}) \right] \right]}
\end{equation}
\vspace{-5pt}
\end{tcolorbox} 
Note that the exploration mechanism is driven by a dual-force mechanism. The first component executes a kernel-weighted value-gradient ascent, steering the ensemble toward high-utility manifolds. Complementing this, the second component exerts a repulsive force that prevents particle collapse into local modes, ensuring the ensemble maintains a broad, diverse coverage. 
Equation~\ref{eq:dynamics} also reveals a powerful collaborative mechanism: the drift of the $i$-th particle is calibrated by the value gradients of its peers. Since computing the value gradient fundamentally requires drawing multiple Monte Carlo samples from a particle's corresponding posterior. Hence, by aggregating Monte Carlo posterior samples across the entire ensemble, the framework effectively pools global information to refine individual trajectories. This collective sharing transforms the estimation of $\nabla V_t$ from a local, high-variance task into a robust, ensemble-wide operation, ensuring highly informed corrections. 
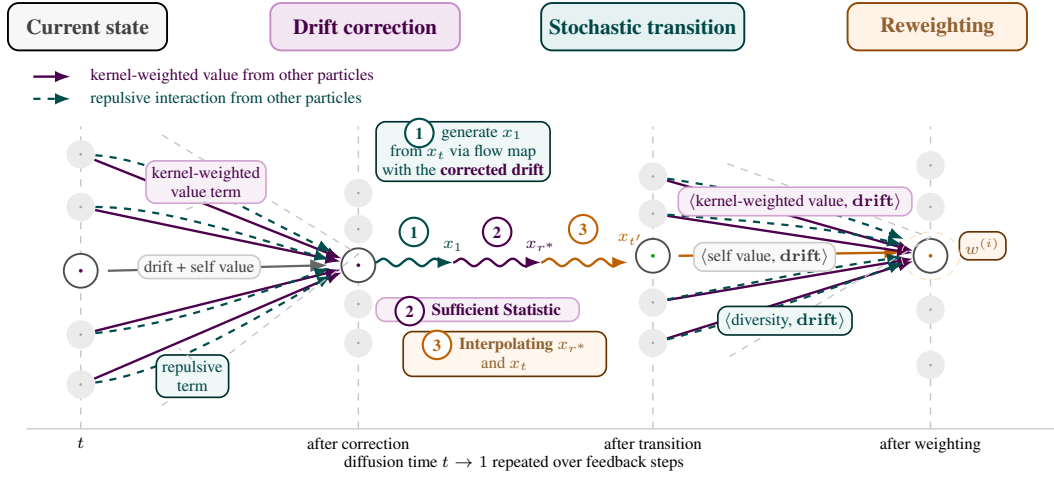
\begin{figure*}[t]
\centering
\resizebox{\textwidth}{!}{%
\begin{tikzpicture}[
    x=1cm,y=0.82cm,
    >=Latex,
    font=\small,
    particle/.style={circle, draw=black!70, fill=white, minimum size=5.0mm, inner sep=0pt, line width=0.8pt},
    faintparticle/.style={circle, draw=none, fill=gray!16, minimum size=4.2mm, inner sep=0pt},
    stageA/.style={draw, rounded corners=1.6mm, line width=0.8pt, fill=gray!6, text=black!75, minimum height=7mm, inner sep=2.5pt, font=\bfseries\footnotesize},
    stageB/.style={draw, rounded corners=1.6mm, line width=0.8pt, fill=violet!10, draw=violet!45, text=violet!55!black, minimum height=7mm, inner sep=2.5pt, font=\bfseries\footnotesize},
    stageC/.style={draw, rounded corners=1.6mm, line width=0.8pt, fill=teal!10, draw=teal!45!black, text=teal!55!black, minimum height=7mm, inner sep=2.5pt, font=\bfseries\footnotesize},
    stageD/.style={draw, rounded corners=1.6mm, line width=0.8pt, fill=orange!10, draw=orange!55!black, text=brown!75!black, minimum height=7mm, inner sep=2.5pt, font=\bfseries\footnotesize},
    noteV/.style={draw, rounded corners=1.4mm, line width=0.7pt, fill=violet!6, draw=violet!35, text=violet!55!black, inner sep=2.5pt, align=center, font=\scriptsize},
    noteT/.style={draw, rounded corners=1.4mm, line width=0.7pt, fill=teal!6, draw=teal!35!black, text=teal!55!black, inner sep=2.5pt, align=center, font=\scriptsize},
    noteO/.style={draw, rounded corners=1.4mm, line width=0.7pt, fill=orange!7, draw=orange!40!black, text=brown!75!black, inner sep=2.5pt, align=center, font=\scriptsize},
    noteG/.style={draw, rounded corners=1.4mm, line width=0.7pt, fill=gray!5, draw=black!20, text=black!75, inner sep=2.5pt, align=center, font=\scriptsize},
    mainarrow/.style={-{Latex[length=2.6mm]}, line width=0.95pt, black!60},
    share/.style={-{Latex[length=2.3mm]}, line width=1pt, violet!60!black},
    repel/.style={-{Latex[length=2.3mm]}, line width=0.95pt, teal!60!black, dashed},
    flow/.style={-{Latex[length=2.3mm]}, line width=1pt, green!55!black},
    weighta/.style={-{Latex[length=2.3mm]}, line width=1pt, orange!75!black},
    cycle/.style={-{Latex[length=2.6mm]}, line width=0.85pt, dashed, black!35},
    guide/.style={black!22, dashed, line width=0.6pt}
]

\node[stageA, minimum width=2.35cm] at (1.9,8.25) {Current state};
\node[stageB, minimum width=2.8cm] at (6.0,8.25) {Drift correction};
\node[stageC, minimum width=3.0cm] at (10.1,8.25) {Stochastic transition};
\node[stageD, minimum width=2.65cm] at (14.45,8.25) {Reweighting};

\draw[black!28] (1.0,1.0) -- (16.3,1.0);
\foreach \x/\lab in {1.8/$t$,5.9/after correction,10.25/after transition,14.35/after weighting} {
  \draw[guide] (\x,1.0) -- (\x,6.45);
  \node[below, font=\scriptsize] at (\x,1.0) {\lab};
}
\node[font=\scriptsize] at (8.2,0.38) {diffusion time $t \rightarrow 1$ repeated over feedback steps};

\node[faintparticle] (p1) at (1.8,5.95) {};
\node[faintparticle] (p2) at (1.8,5.0) {};
\node[particle] (hero) at (1.8,3.85) {};
\node[faintparticle] (p4) at (1.8,2.7) {};
\node[faintparticle] (p5) at (1.8,1.8) {};
\fill[violet!75!black] (hero.center) circle (0.78pt);
\fill[black!35] (p1.center) circle (0.55pt);
\fill[black!35] (p2.center) circle (0.55pt);
\fill[black!35] (p4.center) circle (0.55pt);
\fill[black!35] (p5.center) circle (0.55pt);

\node[faintparticle] at (5.9,5.25) {};
\node[faintparticle] at (5.9,4.6) {};
\node[particle] (corrp) at (5.9,3.95) {};
\node[faintparticle] at (5.9,3.25) {};
\node[faintparticle] at (5.9,2.55) {};
\fill[black!35] (5.9,5.25) circle (0.55pt);
\fill[black!35] (5.9,4.6) circle (0.55pt);
\fill[violet!75!black] (corrp.center) circle (0.78pt);
\fill[black!35] (5.9,3.25) circle (0.55pt);
\fill[black!35] (5.9,2.55) circle (0.55pt);
\draw[mainarrow] (2.2,3.85) -- (5.45,3.95);

\draw[share] (p1) -- (corrp);
\draw[share] (p2) -- (corrp);
\draw[share] (p4) -- (corrp);
\draw[share] (p5) -- (corrp);
\draw[repel] (p1) .. controls (3.0,5.85) and (4.25,4.95) .. (5.52,4.18);
\draw[repel] (p2) .. controls (3.0,5.0) and (4.2,4.5) .. (5.48,4.05);
\draw[repel] (p4) .. controls (3.0,2.85) and (4.18,3.32) .. (5.48,3.9);
\draw[repel] (p5) .. controls (3.0,2.0) and (4.15,2.82) .. (5.52,3.78);
\draw[guide] (3.0,6.3) -- (5.9,4.15) -- (3.0,1.42);

\node[noteV] at (3.65,5.45) {kernel-weighted\\value term};
\draw[guide] (3.65,5.1) -- (3.35,4.95);
\node[noteT] at (3.45,1.95) {repulsive\\term};
\draw[guide] (3.45,2.25) -- (3.25,2.45);
\node[noteG] at (3.55,3.92) {drift + self value};

\draw[-{Latex[length=2.3mm]}, line width=1pt, teal!60!black, decorate, decoration={snake, amplitude=0.55mm, segment length=3.8mm}] (6.15,4.0) -- (7.3,4.0);
\draw[-{Latex[length=2.3mm]}, line width=1pt, violet!60!black, decorate, decoration={snake, amplitude=0.55mm, segment length=3.8mm}] (7.3,4.0) -- (8.6,4.0);
\draw[-{Latex[length=2.3mm]}, line width=1pt, orange!75!black, decorate, decoration={snake, amplitude=0.55mm, segment length=3.8mm}] (8.6,4.0) -- (9.88,4.0);
\node[draw=teal!60!black, circle, line width=0.9pt, minimum size=4.2mm, inner sep=0pt, fill=white, text=teal!60!black, font=\scriptsize\bfseries] at (6.72,4.55) {1};
\node[draw=violet!60!black, circle, line width=0.9pt, minimum size=4.2mm, inner sep=0pt, fill=white, text=violet!60!black, font=\scriptsize\bfseries] at (7.95,4.55) {2};
\node[draw=orange!75!black, circle, line width=0.9pt, minimum size=4.2mm, inner sep=0pt, fill=white, text=orange!75!black, font=\scriptsize\bfseries] at (9.22,4.60) {3};
\node[noteT, anchor=west, minimum width=2.15cm] at (6.15,6.0) {\raisebox{0pt}[0pt][0pt]{\tikz[baseline=-0.55ex]\node[draw=teal!60!black, circle, line width=0.9pt, minimum size=4.2mm, inner sep=0pt, fill=white, text=teal!60!black, font=\scriptsize\bfseries]{1};}\hspace{1mm}generate $x_1$\\from $x_t$ via flow map\\with the \textcolor{violet!55!black}{\textbf{corrected drift}}};
\node[font=\scriptsize, text=teal!55!black] at (7.30,4.30) {$x_1$};
\node[noteV, anchor=west, minimum width=3.0cm] at (6.15,3.15) {\raisebox{0pt}[0pt][0pt]{\tikz[baseline=-0.55ex]\node[draw=violet!60!black, circle, line width=0.9pt, minimum size=4.2mm, inner sep=0pt, fill=white, text=violet!60!black, font=\scriptsize\bfseries]{2};}\hspace{1mm}\textbf{Sufficient Statistic}};
\node[font=\scriptsize, text=violet!55!black] at (8.60,4.30) {$x_{r^*}$};
\node[noteO, anchor=west, minimum width=3.0cm] at (6.55,2.35) {\raisebox{0pt}[0pt][0pt]{\tikz[baseline=-0.55ex]\node[draw=orange!75!black, circle, line width=0.9pt, minimum size=4.2mm, inner sep=0pt, fill=white, text=orange!75!black, font=\scriptsize\bfseries]{3};}\hspace{1mm}\textbf{Interpolating} $x_{r^*}$\\and $x_t$};
\node[faintparticle] (xt1) at (10.25,5.55) {};
\node[faintparticle] (xt2) at (10.25,4.88) {};
\node[particle] (xtp) at (10.25,4.12) {};
\node[faintparticle] (xt4) at (10.25,3.28) {};
\node[faintparticle] (xt5) at (10.25,2.55) {};
\fill[black!35] (10.25,5.55) circle (0.55pt);
\fill[black!35] (10.25,4.88) circle (0.55pt);
\fill[green!65!black] (xtp.center) circle (0.78pt);
\fill[black!35] (10.25,3.28) circle (0.55pt);
\fill[black!35] (10.25,2.55) circle (0.55pt);
\node[font=\scriptsize, text=orange!75!black] at (9.92,4.42) {$x_{t'}$};

\node[faintparticle] at (14.35,5.75) {};
\node[faintparticle] at (14.35,5.00) {};
\node[particle] (xw) at (14.35,4.12) {};
\node[faintparticle] at (14.35,3.15) {};
\node[faintparticle] at (14.35,2.15) {};
\fill[black!35] (14.35,5.75) circle (0.55pt);
\fill[black!35] (14.35,5.00) circle (0.55pt);
\fill[orange!75!black] (xw.center) circle (0.78pt);
\fill[black!35] (14.35,3.15) circle (0.55pt);
\fill[black!35] (14.35,2.15) circle (0.55pt);
\draw[weighta] (10.63,4.12) -- (13.90,4.18);
\draw[share] (xt1) -- (xw);
\draw[share] (xt2) -- (xw);
\draw[share] (xt4) -- (xw);
\draw[share] (xt5) -- (xw);
\draw[repel] (xt1) .. controls (11.35,5.35) and (12.80,5.00) .. (13.96,4.42);
\draw[repel] (xt2) .. controls (11.35,4.78) and (12.78,4.66) .. (13.94,4.32);
\draw[repel] (xt4) .. controls (11.35,3.55) and (12.78,3.90) .. (13.94,4.10);
\draw[repel] (xt5) .. controls (11.35,2.88) and (12.80,3.48) .. (13.96,4.00);
\draw[guide] (11.35,5.78) -- (14.35,4.45) -- (11.35,2.28);
\node[noteG] at (11.90,4.10) {$\langle\text{self value, }\mathbf{drift}\rangle$};
\node[noteO] at (15.12,4.30) {$w^{(i)}$};
\draw[orange!28, dashed] (14.35,4.12) circle (0.43);
\node[noteV, minimum width=3.45cm] at (12.35,5.10) {$\langle\text{kernel-weighted value, }\mathbf{drift}\rangle$};
\node[noteT] at (12.20,2.95) {$\langle\text{diversity, }\mathbf{drift}\rangle$};


\draw[share] (1.1,7.35) -- +(0.55,0);
\node[anchor=west, font=\scriptsize, text=violet!55!black] at (1.82,7.35) {kernel-weighted value from other particles};
\draw[repel] (1.1,6.95) -- +(0.55,0);
\node[anchor=west, font=\scriptsize, text=teal!55!black] at (1.82,6.95) {repulsive interaction from other particles};
\end{tikzpicture}%
}
\caption{\small{Sampling Mechanism of IMPFM.}}
\label{fig:impfm-mechanism}
\vspace{-8pt}
\end{figure*}
However, our objective is to sample from a value-tilted distribution ($p_1(x) e^{r(x)}$) rather than a standard Boltzmann distribution ($e^{V(x)}$). Because the Boltzmann distribution lacks explicit constraints on distributional coverage, relying on it risks sampling outside the true data manifold. To resolve this, we introduce a reweighting mechanism that guarantees our interactive multi-particle system converges strictly to the desired target distribution while preserving enhanced exploration capabilities of MPC. 

To construct a principled reweighting mechanism, we begin by recognizing that while highly efficient, the deterministic nature of ODE sampling inherently restricts exploration, necessitating stochastic transitions for effective search. 
To achieve this, we first sample standard noise $\epsilon \sim \mathcal{N}(0, I)$ and push it forward through the distilled flow map $\hat{v}(\epsilon; x_t)$. By definition, this directly yields exact samples from the true posterior, $x_1 \sim p_{1|t}(x_1 | x_t)$. 
With the trajectory's boundary points ($\epsilon$ and $x_1$) firmly anchored, we induce a DDPM-style dynamic duration transition via the Sufficient Statistic~\citep{holderrieth2025glass}. Specifically, we perform the following steps:
\vspace{-5pt}
\begin{equation}
\begin{aligned}
    & \small{\epsilon \xrightarrow{\hat{v}(\epsilon; x_t)} x_1 \sim p_{1|t}(x_1 | x_t) \xrightarrow{\text{Interp.}} x_{r^\ast} = \bar\alpha_{r^\ast} x_1 + \bar\sigma_{r^\ast} \varepsilon_1 \xrightarrow{\text{Trans.}} x_{t'} = \alpha_{t'} \frac{ \bar{\alpha}_{r^*} \sigma_t^2 x_{r^*} + \alpha_t \bar{\sigma}_{r^*}^2 x_t}{\bar{\alpha}_{r^*}^2 \sigma_t^2 + \alpha_t^2 \bar{\sigma}_{r^*}^2},} \\[-6pt]
    & \small{r^*(t, t') = g^{-1}\left( \frac{g(t)g(t')}{g(t) - g(t')} \right), \quad g(t) = \frac{\sigma^2_t}{\alpha^2_t}, \quad \varepsilon \sim \mathcal{N}(0, I), \quad \varepsilon_1 \sim \mathcal{N}(0, I).}
\end{aligned}
\label{eq:bss}
\end{equation}
$\bar{\alpha}$ and  $\bar{\sigma}^2$ are defined in the Appendix~\ref{sec:ss-comp}. Here, the sampled $x_1$ serves as fixed boundary points to compute the stochastic interpolant $x_{r^*}$, which acts as the essential foundation for constructing the final DDPM-style transition sample $x_{t'}$.

Armed with DDPM-style transitions, the interactive drift-correction mechanism naturally facilitates a FKC-style reweighting scheme, further amplifying the sampler’s capability. While standard value-based reweighting greedily preserves high-reward trajectories at the expense of diversity—inevitably leading to weight degeneracy and mode collapse—we introduce a principled, interaction-aware alternative. By leveraging ensemble-wide multi-particle dynamics, our proposed reweighting mechanism actively preserves the structural diversity of high-utility samples, encouraging broad exploration and global convergence. Specifically, we introduce the following scoring function to weight each particle 
\vspace{-6pt}
\begin{tcolorbox}[colback=blue!5!white, colframe=gray!80!black, title={(Proposed Particle Reweighting Mechanism Involving Multi-particle Interaction)}]
\begin{equation}\label{eq:weight}
\small{\underbrace{\bm{\langle} \nabla_{x_t^i}  V(x_t^{i}), \bm{b_t(x_t^i)} \bm{\rangle} + \frac{1}{n-1} \sum_{j=1, j \neq i}^n \left[ \bm{\langle} k(x_t^j, x_t^{i}) \nabla_{x_t^j} V(x_t^j), \bm{b_t(x_t^i)} \bm{\rangle} + \bm{\langle} \nabla_{x_t^j} k(x_t^j, x_t^{i}), \bm{b_t(x_t^i)} \bm{\rangle}  \right]}_{\textbf{weight of particle $i$ (i.e. $w^{(i)}_t$)}}}
\end{equation}
Intuitively,~\ref{eq:weight} prevents off-manifold degradation by projecting the value and diversity gradients onto the prior model dynamics $b_t(x_t^i)$. A particle is weighted highly only if its pursuit of high reward—both locally and via the ensemble—aligns with the base model dynamics, preventing it from straying into low-likelihood, off-manifold regions. Similarly, the repulsive interaction term guarantees that the drive for sample diversity does not compromise sample quality, as exploratory trajectories are strictly penalized if they violate the prior model dynamics.
\end{tcolorbox}
To maintain sampling efficiency, we trigger a standard resampling mechanism whenever the effective sample size (ESS) falls below a defined threshold:
$\text{ESS}_t^n = \frac{\left( \sum_{i=1}^n w_t^{(i)} \right)^2}{\sum_{i=1}^n \left( w_t^{(i)} \right)^2} \in [1, n]$.
Crucially, this resampling procedure leverages importance sampling to ensure the resulting samples remain asymptotically unbiased.
Next, we demonstrate that coupling the interactive drift correction (Eq.~\ref{eq:dynamics}) with our reweighting mechanism (Eq.~\ref{eq:weight}) yields an \emph{interaction-aware Feynman-Kac corrector}, as formalized in Proposition 4.3.
\begin{tcolorbox}[colback=blue!5!white, colframe=gray!80!black, title={\textbf{Proposition 4.3} (IMPFM Yields Feynman-Kac Corrector to KL-Tilted Target)}]\label{prop:3}
Consider a Flow model with drift $b_t(x)$ trained to sample from $q_t(x)$. Sampling from the KL-tilted marginals $p_t^{\text{kl}}(x) \propto q_t(x) \exp\Big( -D_{\text{KL}}\big( q_t(x) \parallel p_{\text{boltz}}(x) \big) \Big)$ is performed by an weighted SDE with an equivalent corresponding weighted ODE:
\begin{equation}
    \small{dx^i_t = b_t(x^i_t)\>dt  + \frac{\sigma_t^2}{2}\frac{1}{n} \sum_{j=1}^n \left[ k(x^j_t, x^{i}_t) \nabla_{x^j_t} V(x^j_t) \> dt + \nabla_{x^j_t} k(x^j_t, x^{i}_t) \> dt \right]}
\end{equation}
\vspace{-15pt}
\begin{equation}
    \small{dw^i_t = \frac{1}{n} \sum_{j=1}^n \left[ \bm{\langle} k(x_t^j, x_t^{i}) \nabla_{x_t^j} V(x_t^j), b_t(x_t^i) \bm{\rangle} \> dt + \bm{\langle} \nabla_{x_t^j} k(x_t^j, x_t^{i}), b_t(x_t^i) \bm{\rangle}\> dt \right]} 
\end{equation}
\end{tcolorbox} 
Thus, as shown in 
Proposition 4.3,
our framework jointly steers the particle ensemble toward the true reward-tilted distribution while maintaining high likelihood under the prior to preserve intrinsic sample quality. By leveraging multi-particle interactions to explicitly minimize the KL divergence to the target, this holistic approach fundamentally circumvents the severe vulnerabilities of reward over-optimization and mode collapse inherent in isolated trajectory optimization. We term our framework Sequentially Controlled Interactive Multi-Particle Flow-Maps (IMPFM) (See Fig.~\ref{fig:impfm-mechanism}). 

To broaden the applicability of the IMPFM to settings lacking an off-the-shelf stochastic flow map, we seamlessly adapt our approach using a standard deterministic flow matching model—
paired with a consistency model enabling a single-step sample (i.e., $x_1$) generation from any $x_t$. 
However, the inherent determinism of consistency models precludes sampling multiple $x^{(i)}_1$ from the posterior $p_{1|t}(\cdot|x_t)$, a critical requirement for estimating the value gradient $\nabla V_t$ and key to computing the corrected dynamics (Eq.~\ref{eq:dynamics}) and reweighting scores (Eq.~\ref{eq:weight}). 
To bypass this limitation, we introduce an \emph{iterative posterior sampling scheme: we alternate between predicting $x_1$ from $x_t$ via the consistency model, and injecting noise via a stochastic interpolant to produce a perturbed $x_t$. Repeating this sequence effectively yields valid samples (i.e., $x_1^{(1)}, x_1^{(2)}, \ldots x_1^{(K),} $) from the desired posterior}, as depicted below (where $\epsilon_1, \epsilon_2, \ldots, \epsilon_k \stackrel{\text{i.i.d.}}{\sim} \mathcal{N}(0, I)$):
\begin{equation}
    x_t^{(0)} \xrightarrow{\text{Consistency}} \textcolor{red}{x_1^{(1)}} \xrightarrow{x_t^{(1)} = \alpha_t x_1^{(1)} + \sigma_t \epsilon_1} \textcolor{red}{x_t^{(1)}} \xrightarrow{\text{Consistency}} \textcolor{blue}{x_1^{(2)}} \xrightarrow{x_t^{(2)} = \alpha_t x_1^{(2)} + \sigma_t \epsilon_2} \textcolor{blue}{x_t^{(2)}} \cdots \xrightarrow{\text{Repeat}} \textcolor{purple}{x_1^{(K)}}
\label{eq:explicit_iterative}
\end{equation}
\vspace{-28pt}






\section{Experiments and Results}
\begin{figure*}[h]
\centering

\newcommand{\legenditem}[4]{%
  \begin{tikzpicture}[baseline=-0.6ex]
    \draw[#1, line width=1.8pt, #2] (0,0) -- (0.9,0);
    \node[text=#1, fill=white, inner sep=0.6pt] at (0.45,0) {\scriptsize $#3$};
  \end{tikzpicture}\,#4%
}

\newcommand{\methodlegend}{%
  {\footnotesize
  \setlength{\tabcolsep}{4pt}%
  \begin{tabular}{@{}cccccc@{}}
    \legenditem{tealcyan}
    {dashed}{\boldsymbol{+}}{FKC} &
    \legenditem{acadblue}
    {solid}{\circ}{DAS} &
    \legenditem{red!75!black}{dashed}{\square}{FKS} &
    \legenditem{orange!90!black}{dash dot}{\triangle}{BoN} &
    \legenditem{purple!80!black}{dotted}{\diamond}{MFM} &
    \legenditem{green!60!black}{solid}{\blacktriangledown}{\textbf{IMPFM (Our)}}
  \end{tabular}%
  }%
}

\newcommand{\barpanel}[2]{%
  \begin{minipage}[t]{0.185\textwidth}
    \centering
    \includegraphics[width=\linewidth]{#1}\par
    \vspace{-0.3ex}
    {\scriptsize #2}
  \end{minipage}%
}

\methodlegend
\vspace{0.1ex}
\noindent\rule{\textwidth}{0.5pt}
\vspace{0.3ex}
\barpanel{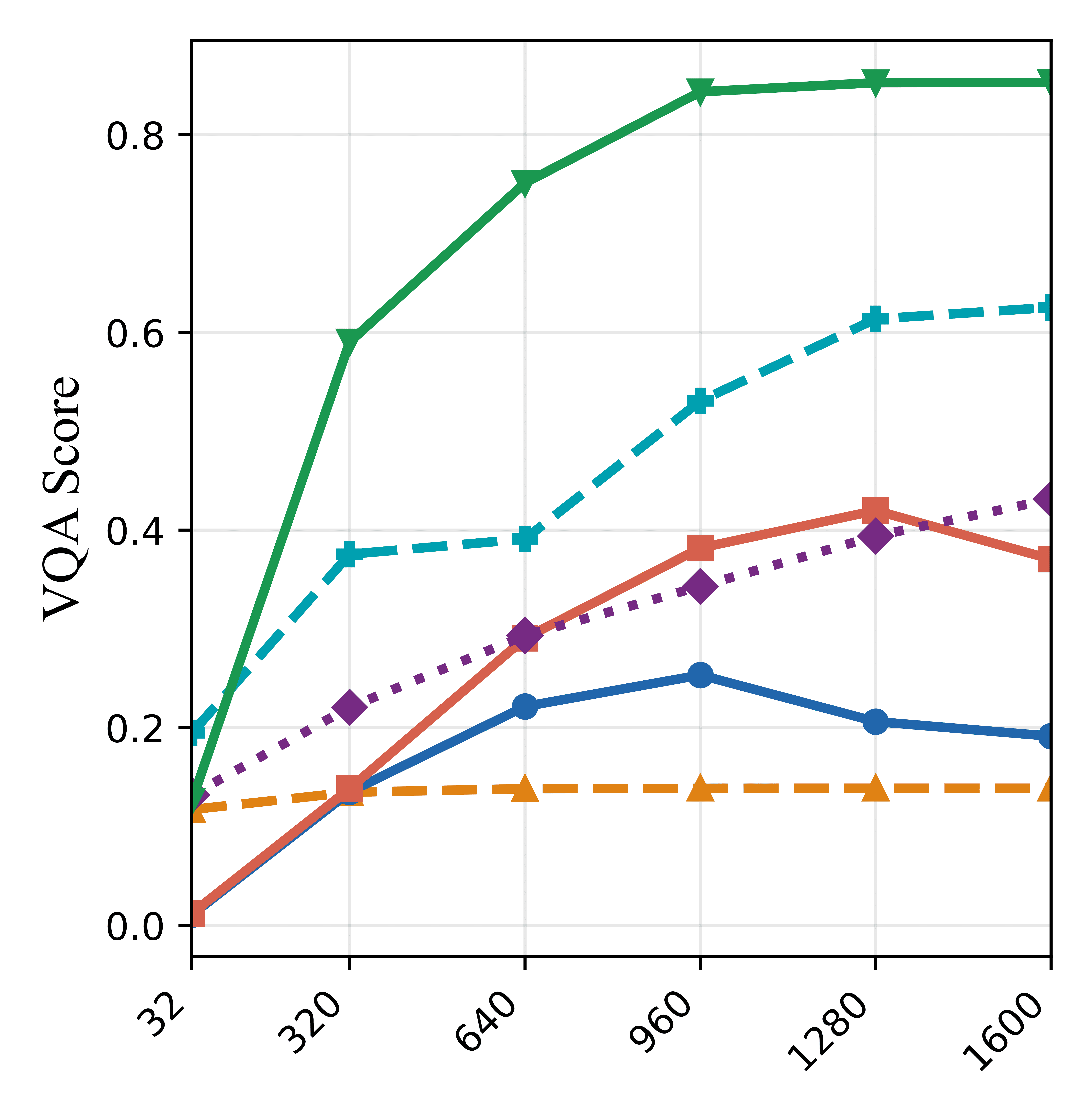}{Feedback Budgets}\hspace{0.004\textwidth}%
\barpanel{figs/Pick_score.png}{Feedback Budgets}\hspace{0.004\textwidth}%
\barpanel{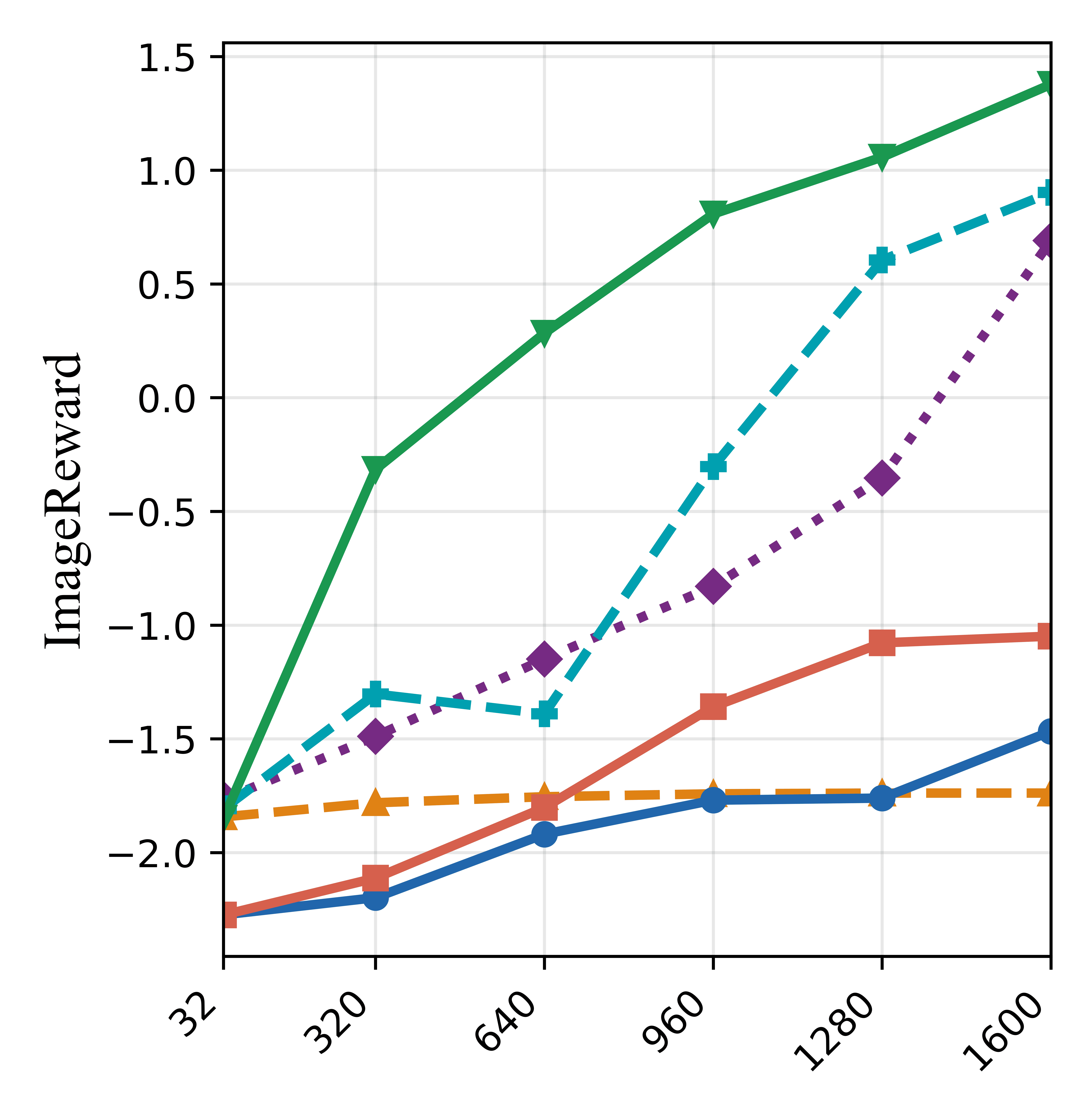}{Feedback Budgets}\hspace{0.004\textwidth}%
\barpanel{figs/Inst.png}{Feedback Budgets}\hspace{0.004\textwidth}%
\barpanel{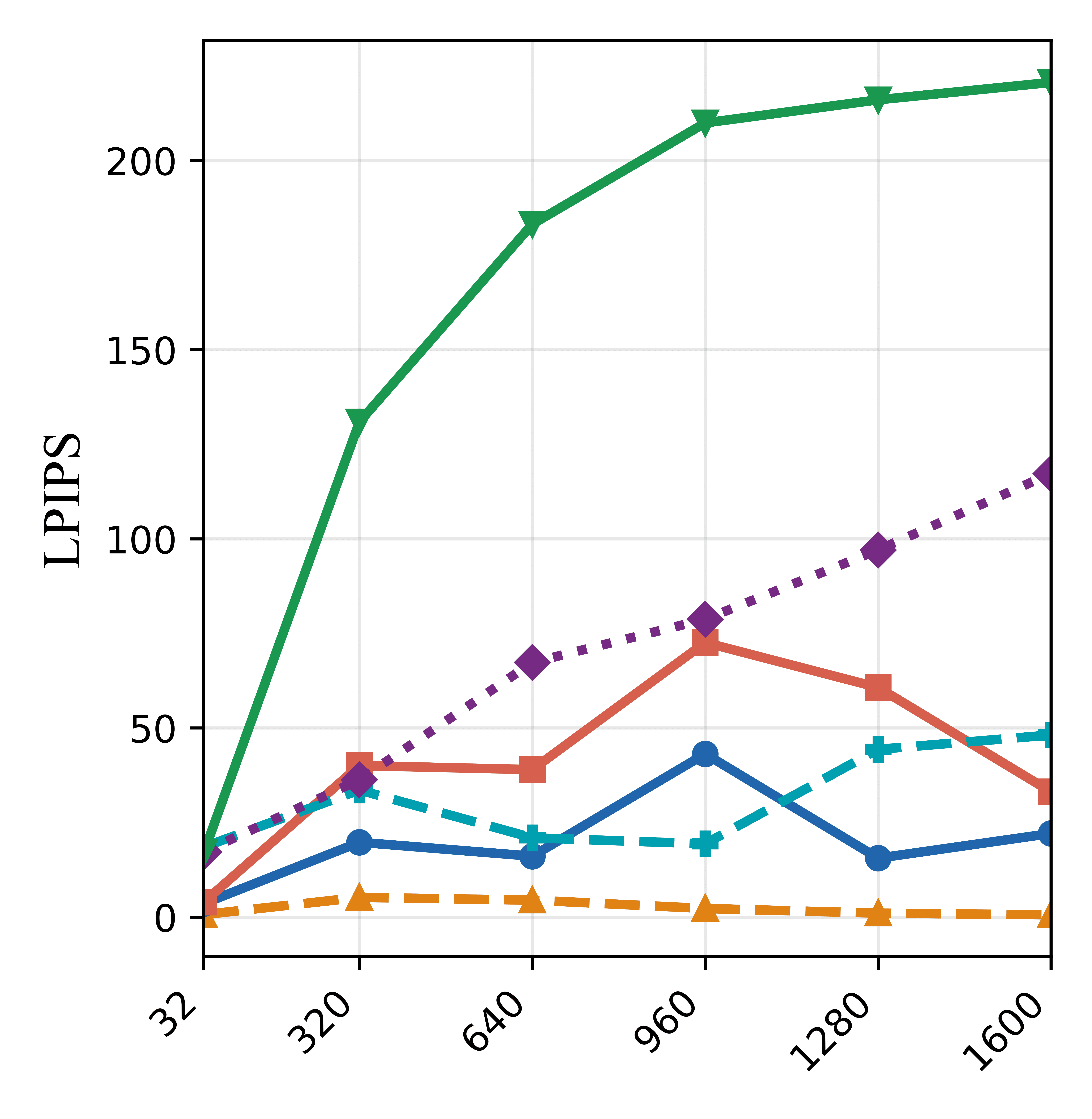}{Feedback Budgets}
\vspace{-7pt}
\caption{\small{Search Performance Analysis of Competitive Approaches on ImageNet Target Classes.}}
\label{fig:search-result}
\end{figure*}
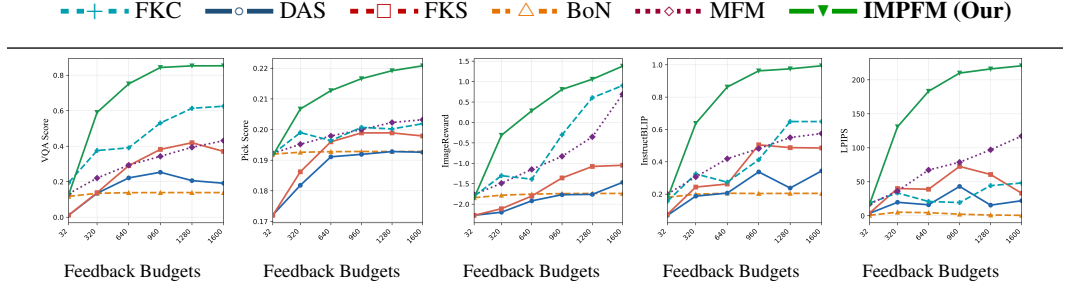

\vspace{-6pt}
\vspace{-3pt}
We evaluate IMPFM across two settings.
In the \emph{feedback-efficient search} setting, a target-class-conditioned prior operates without a priori knowledge of the fine-grained target prompt, relying entirely on iterative feedback to progressively steer generated samples toward the target specification revealed only through sequential interaction.
In the \emph{feedback-driven alignment} setting, a prior model, conditioned on the target prompt, leverages sequential feedback to iteratively refine its outputs toward satisfying complex, fine-grained constraints — including compositional and quantitative reasoning.
We benchmark IMPFM against a diverse set of competitive, state-of-the-art baselines. These include FKC~\citep{skreta2025feynman}, Feynman-Kac Steering (FKS)~\citep{singhal2025general}, the SMC-based approach DAS~\citep{kim2025test}, Best-of-N, and Meta-Flow-Maps with gradient-based estimator (MFM-G, hereafter referred to as MFM)~\citep{potaptchik2026meta}. To simulate online feedback, we employ widely adopted reward models, i.e., VQA score measured with CLIP-FlanT5~\citep{lin2024evaluating}, and reward score using InstructBLIP~\citep{dai2023instructblip}, ImageReward~\citep{xu2023imagereward}, and Pick-Score~\citep{kirstain2023pick}. Sample diversity is evaluated using LPIPS~\citep{kim2025test} score. 
\vspace{-11pt}
\paragraph{Search Experimental Setting:} To evaluate the search capabilities of IMPFM, we utilize ImageNet-1K. For the prior, we employ a distilled flow map model optimized jointly using the $\mathcal{L}^{\text{distill}}_{\text{inst}}$ and $\mathcal{L}^{\text{distill}}_{\text{const}}$. We evaluate on 200 randomly selected target classes, revealing their fine-grained specifications strictly via iterative feedback (see Appendix~\ref{sec:se-class} for example class and prompt details).  %
\vspace{-10pt}
\paragraph{Search Result:} Fig.~\ref{fig:search-result} reports the VQA scores, reward scores, and diversity evaluated across different feedback budgets ($\mathcal{B}$). Experimental results demonstrate that \emph{IMPFM achieves higher reward scores with considerably fewer feedback than competing approaches}. This highlights its efficacy for feedback-driven search, a setting where maximizing reward while ensuring sample efficiency is paramount. 
IMPFM achieves superior sample efficiency through a highly effective information-sharing mechanism. Unlike prior approaches, the posterior samples generated by any single particle are shared across the entire ensemble for drift correction and reweighting, maximizing information utilization and significantly boosting the framework's overall sample efficiency.
\begin{figure*}[h]
\centering
\newcommand{\stacktriplepanel}[3]{%
  \setlength{\tabcolsep}{2pt}%
  \begin{tabular}{@{}c@{\hspace{4pt}}c@{\hspace{4pt}}c@{}}
    \includegraphics[width=0.31\linewidth]{#1} &
    \includegraphics[width=0.31\linewidth]{#2} &
    \includegraphics[width=0.31\linewidth]{#3}
  \end{tabular}%
}
\newcommand{\stackpanelblock}[4]{%
  \begin{minipage}[t]{\linewidth}
    \centering
    \stacktriplepanel{#2}{#3}{#4}
    \par\vspace{0.4ex}
    {\itshape ``#1''\par}
  \end{minipage}%
}
\newcommand{\stacktopheaders}{%
  {\fontsize{9.5}{10.5}\selectfont
  \begin{tabular}{@{}c@{\hspace{4pt}}c@{\hspace{4pt}}c@{}}
    \makebox[0.31\linewidth][c]{\textbf{FKC}} &
    \makebox[0.31\linewidth][c]{\textbf{MFM}} &
    \makebox[0.31\linewidth][c]{\textbf{IMPFM}}
  \end{tabular}%
  }
}
\begin{minipage}[t]{0.49\textwidth}
\centering
\stacktopheaders\par
\vspace{-1.0ex}
\noindent\rule{\linewidth}{0.6pt}\par
\vspace{0.6ex}
\stackpanelblock{\textcolor{purple}{a red daisy.}}{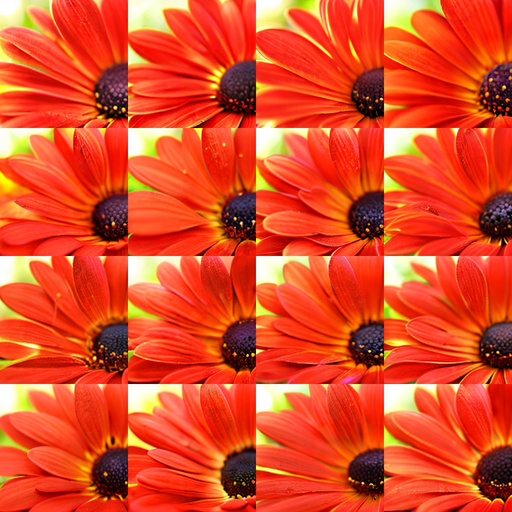}{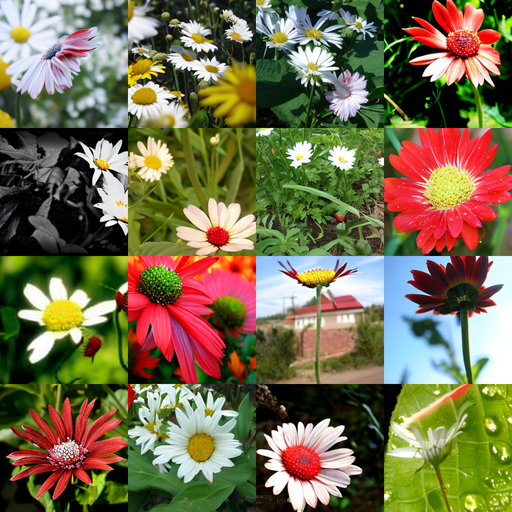}{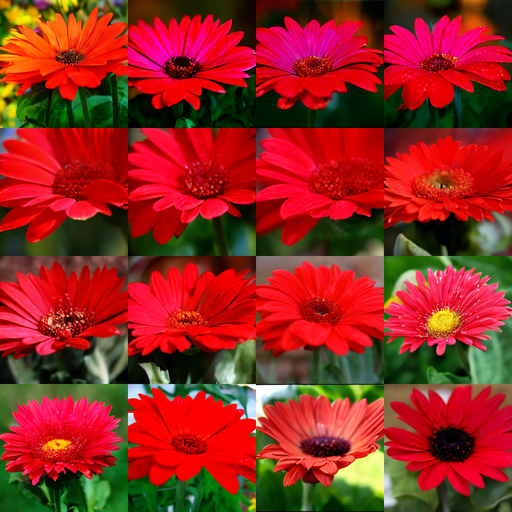}
\end{minipage}%
\hspace{2pt}%
\begin{minipage}[t]{0.49\textwidth}
\centering
\stacktopheaders\par
\vspace{-1.0ex}
\noindent\rule{\linewidth}{0.6pt}\par
\vspace{0.6ex}
\stackpanelblock{\textcolor{purple}{a dog in the snow.}}{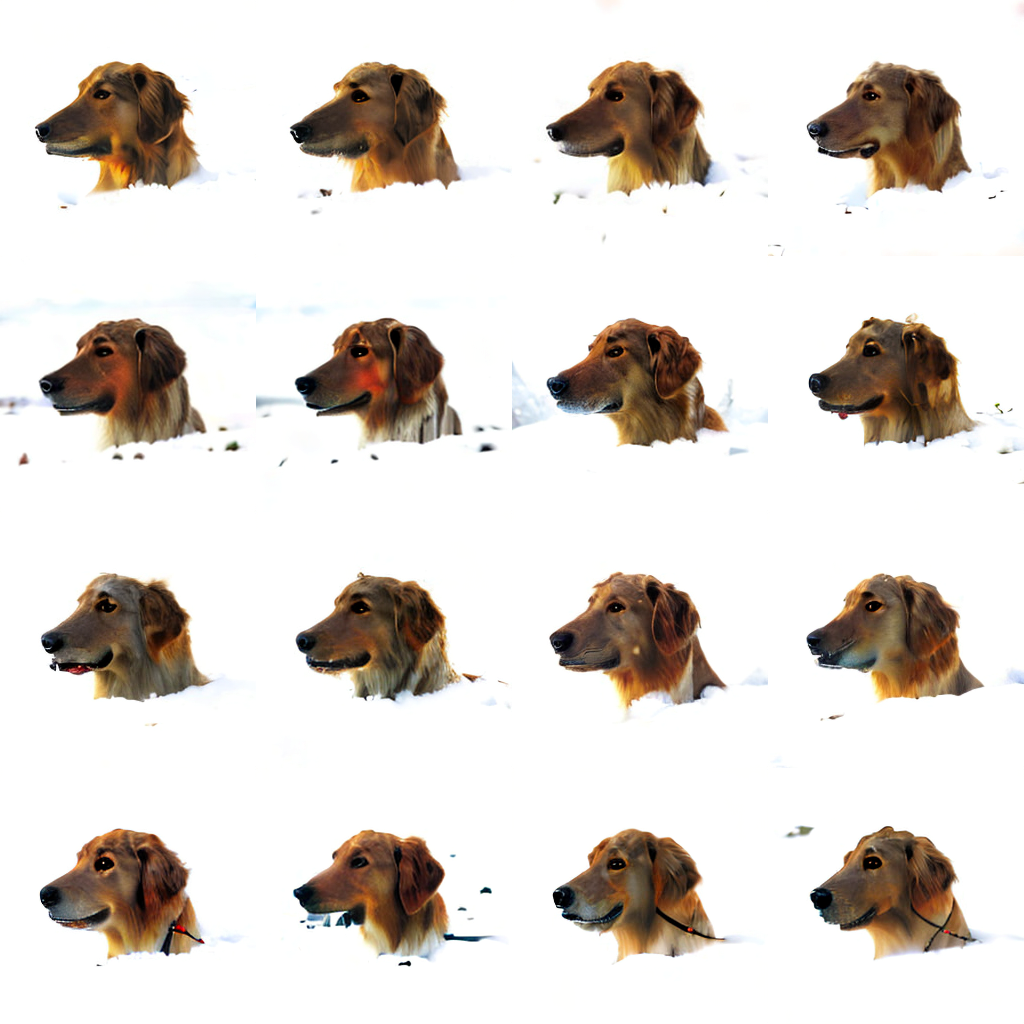}{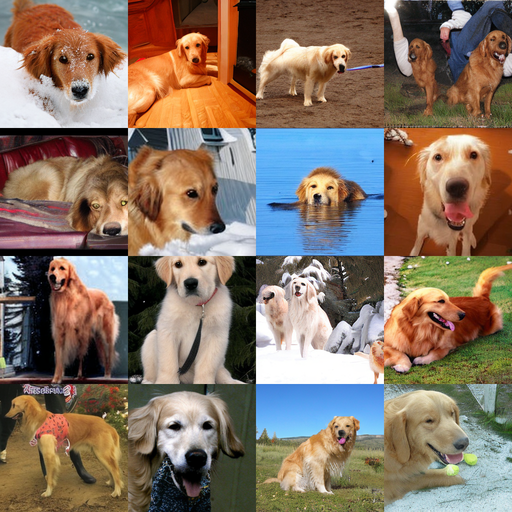}{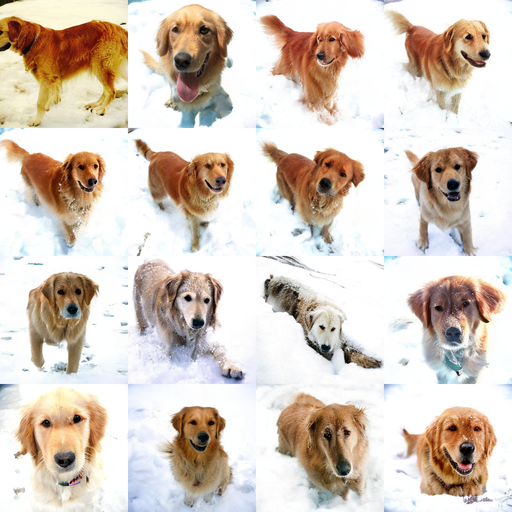}
\end{minipage}
\vspace{0.6ex}
\caption{\small{Search Visualizations.}}
\label{fig:search-visu}
\vspace{-10pt}
\end{figure*}
\vspace{-8pt}

Notably, IMPFM demonstrates superior resistance to reward over-optimization. While both IMPFM and the MFM baseline utilize ImageReward to steer the dynamics, IMPFM consistently outperforms the baselines across held-out evaluation reward models, e.g., Pick-Score~\citep{kirstain2023pick}. 
We attribute this robustness to our ensemble-based reward guidance mechanism. Rather than evaluating each particle against local reward signals in isolation, IMPFM aggregates reward feedback across the full ensemble's posterior samples. This global aggregation effectively smooths the reward landscape, establishing a broader consensus for the guidance that prevents individual particles from collapsing into the narrow, high-reward local maxima that typically emerge under purely local evaluation.
Crucially, IMPFM unlocks a striking increase in semantic diversity among the generated samples, as reflected in the LPIPS score. This underscores the value of our explicit, diversity-inducing repulsive force—demonstrating that actively enforcing particle separation for reweighting and drift correction is key for diversity.
We visualize the generated samples in Fig.~\ref{fig:search-visu} (See Appendix~\ref{sec:se-vis-a} for more viz). 

\begin{figure*}[h]
    \centering
    \begin{subfigure}[t]{0.49\textwidth}
        \centering
        {\small\itshape \textcolor{purple}{"A Red Sports Car."}\par}
        \vspace{0.3ex}
        \begin{minipage}[t]{0.28\linewidth}
            \centering
            {\scriptsize\textbf{\small{FKC}}}\par
            \includegraphics[width=\linewidth]{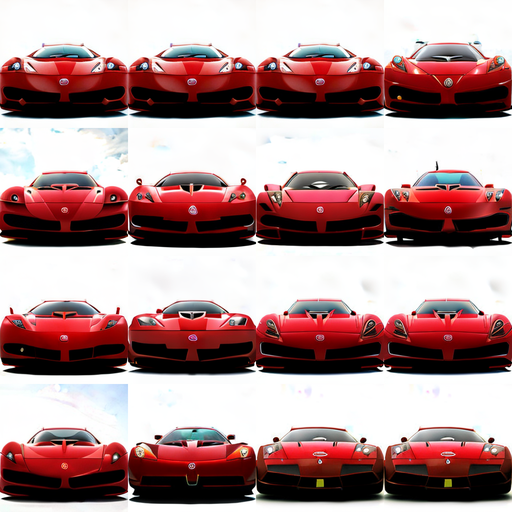}
        \end{minipage}\hfill
        \begin{minipage}[t]{0.28\linewidth}
            \centering
            {\scriptsize\textbf{IMPFM}}\par
            \includegraphics[width=\linewidth]{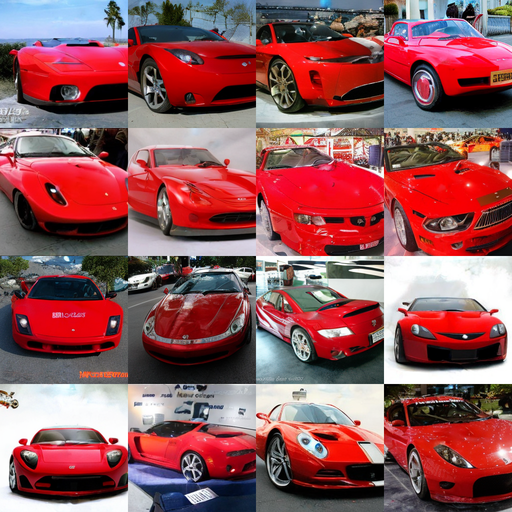}
        \end{minipage}\hfill
        \begin{minipage}[t]{0.28\linewidth}
            \centering
            {\small\textbf{ESS}}\par
            \includegraphics[width=\linewidth]{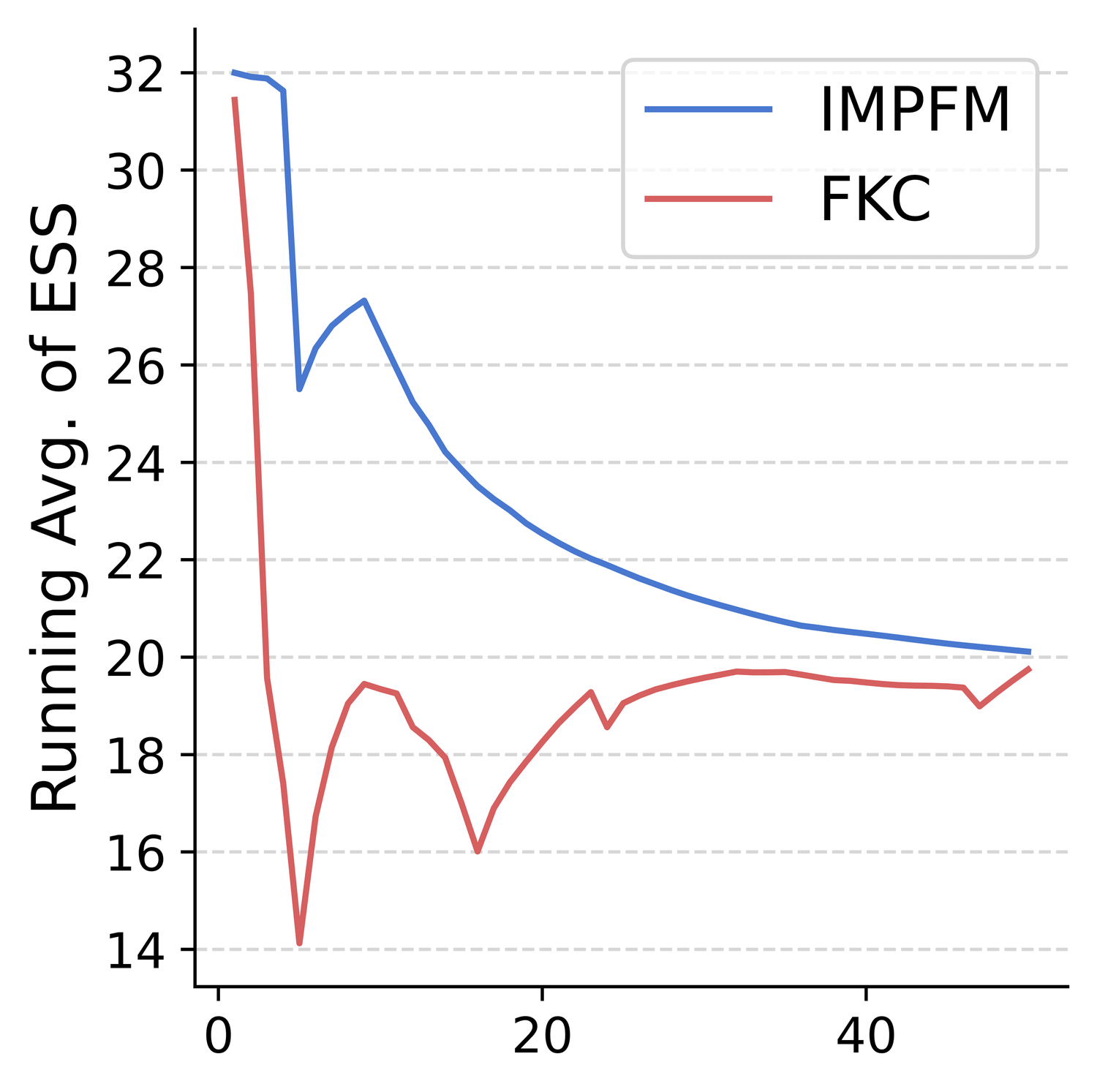}\par
            {\scriptsize\textbf{Resampling Steps}}
        \end{minipage}
        \caption{\small{Impact of Particle Interaction on FKC.}}
        \label{fig:fkc-reward}
    \end{subfigure}%
    \hfill
    \begin{subfigure}[t]{0.49\textwidth}
        \centering
        {\small\itshape \textcolor{purple}{"A Red Daisy"}\par}
        \vspace{0.3ex}
        \begin{minipage}[t]{0.235\linewidth}
            \centering
            {\small\textbf{MFM}}\par
            \includegraphics[width=\linewidth]{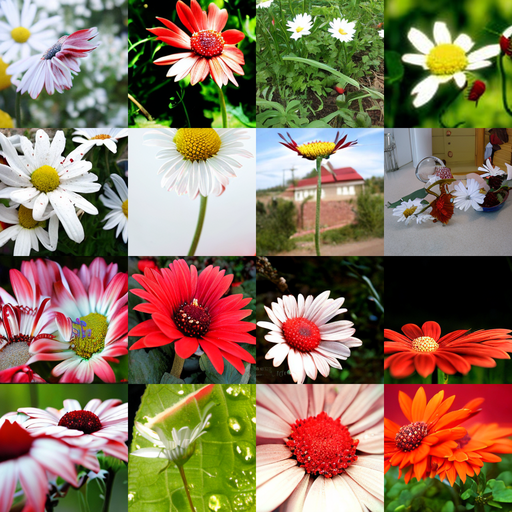}
        \end{minipage}\hfill
        \begin{minipage}[t]{0.235\linewidth}
            \centering
            {\small\textbf{MPC}}\par
            \includegraphics[width=\linewidth]{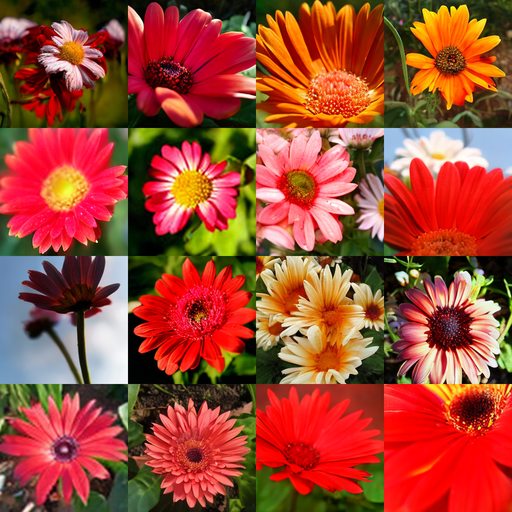}
        \end{minipage}\hfill
        \begin{minipage}[t]{0.245\linewidth}
            \centering
            {\small\textbf{Reward}}\par
            \includegraphics[width=\linewidth]{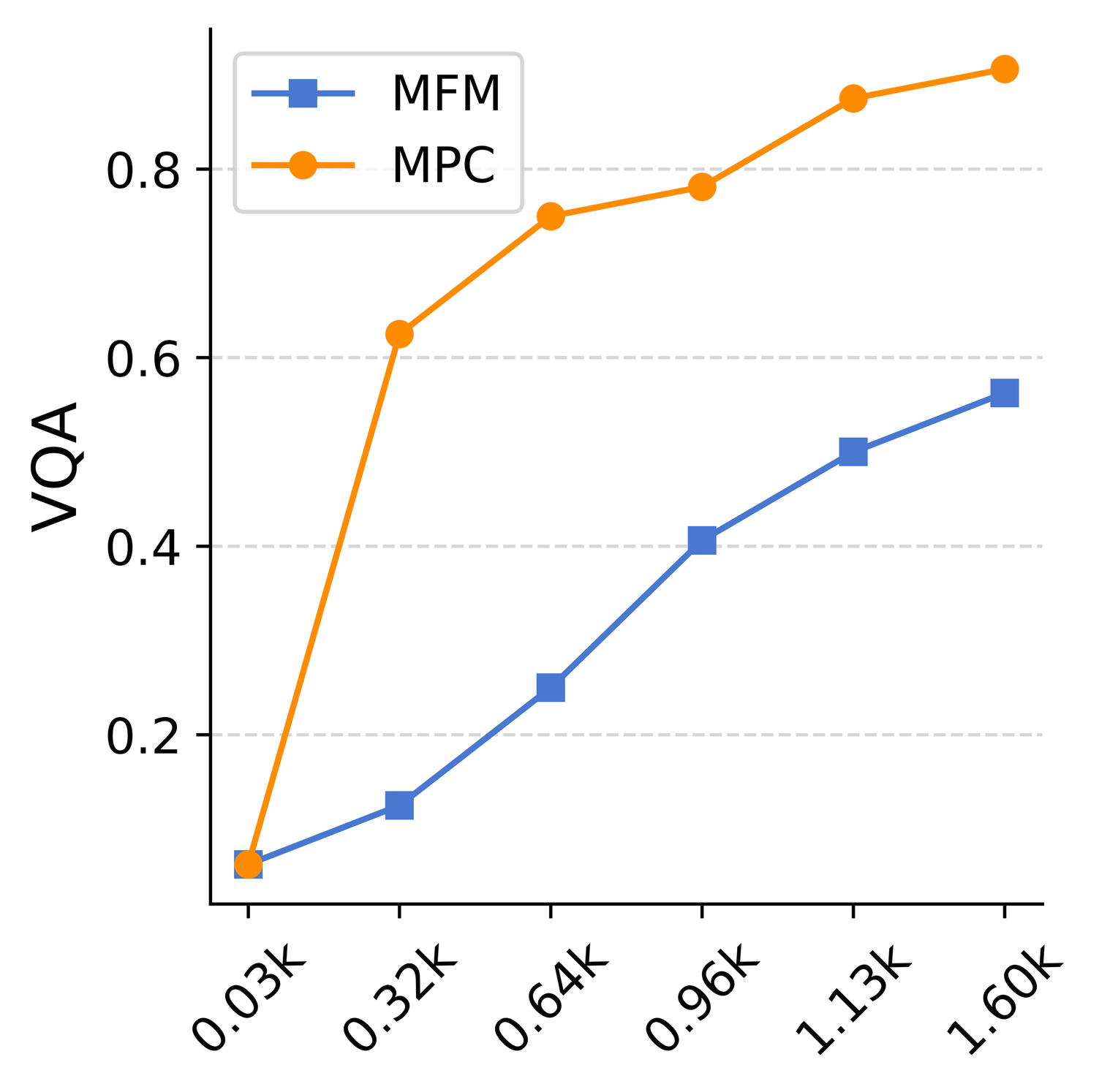}\par
            {\scriptsize\textbf{Feedback}}
        \end{minipage}\hfill
        \begin{minipage}[t]{0.245\linewidth}
            \centering
            {\small\textbf{Diversity}}\par
            \includegraphics[width=\linewidth]{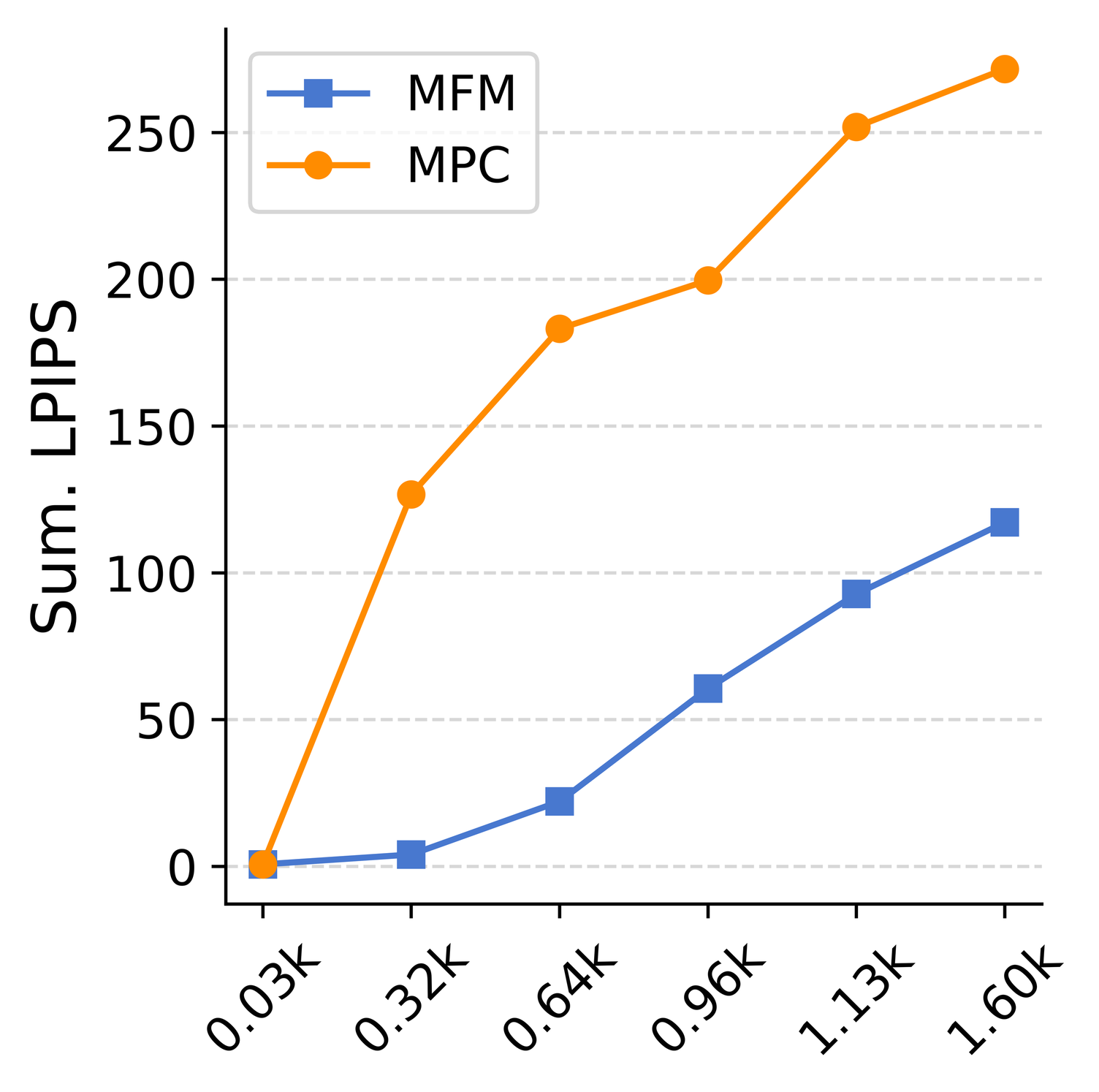}\par
            {\scriptsize\textbf{Feedback}}
        \end{minipage}
        
        \vspace{7pt}
        \caption{\small{Impact of Particle Interaction on Optimal Control.}}
        \label{fig:oc}
    \end{subfigure}
    \vspace{-6pt}
    \label{fig:impfm_ablation}
\end{figure*}
\paragraph{Emergent Stability: Eradicating Weight Degeneracy via Particle Interaction}
\vspace{-7pt}
Weight degeneracy remains a fundamental bottleneck in traditional SMC/FKC frameworks. To isolate the contribution of particle interaction, we conduct two ablation tests. First, we remove the interaction term from both the drift correction (Eq.~\ref{eq:dynamics}) and reweighting (Eq.~\ref{eq:weight}), reducing the sampler to a standard reward-tilted FKC. Second, we take a purely value-guided optimal control sampler — MFM — and equip it with our proposed interactive drift correction mechanism (Eq.~\ref{eq:dynamics}), forming a variant we refer to as MPC, to assess the performance gains attributable to interaction alone. As shown in Fig.~\ref{fig:fkc-reward}, the interaction-aware formulation in both drift correction and reweighting is critical: it promotes substantially greater sample diversity, sustains a persistently high ESS across all correction steps— failure modes endemic to the FKC. Similarly, comparative results between MFM and MPC, as depicted in Fig.~\ref{fig:oc}, confirm that particle interaction-based drift correction (Eq.~\ref{eq:dynamics}) is key to overcoming mode collapse and reward over-optimization even within a purely value-based optimal control setting, as evidenced by richer sample diversity and consistently higher VQA scores. Additional analysis is in the Appendix~\ref{sec:wd-a},~\ref{sec:int-fkc}.
\paragraph{Importance of Reweighting Mechanism}
\begin{wrapfigure}{r}{0.48\textwidth}
    \vspace{-10pt}
    \centering
    \vspace{-1.25\baselineskip}
    {\scriptsize\itshape \textcolor{purple}{"A green bell pepper."}\par}
    \begin{minipage}[t]{0.30\linewidth}
        \centering
        {\scriptsize\textbf{MPC}}\par
        \includegraphics[width=\linewidth]{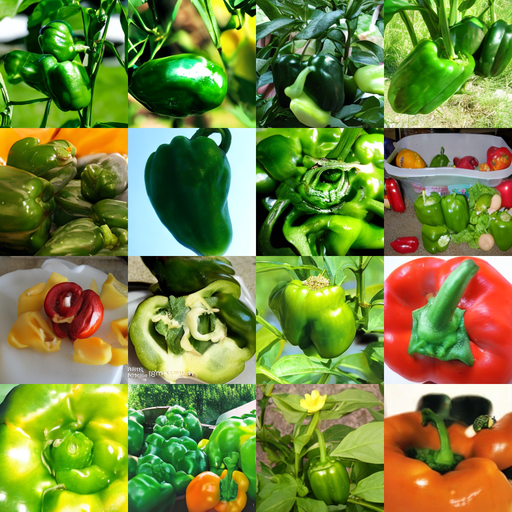}
    \end{minipage}\hspace{3pt}%
    \begin{minipage}[t]{0.30\linewidth}
        \centering
        {\scriptsize\textbf{IMPFM}}\par
        \includegraphics[width=\linewidth]{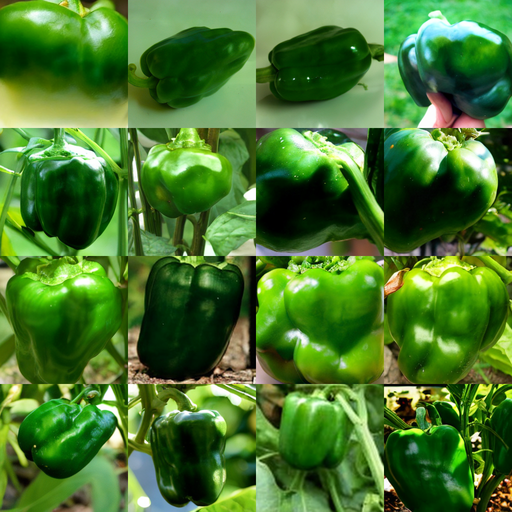}
    \end{minipage}\hspace{3pt}%
    \begin{minipage}[t]{0.34\linewidth}
        \centering
        {\scriptsize\textbf{Reward}}\par
        \includegraphics[width=\linewidth]{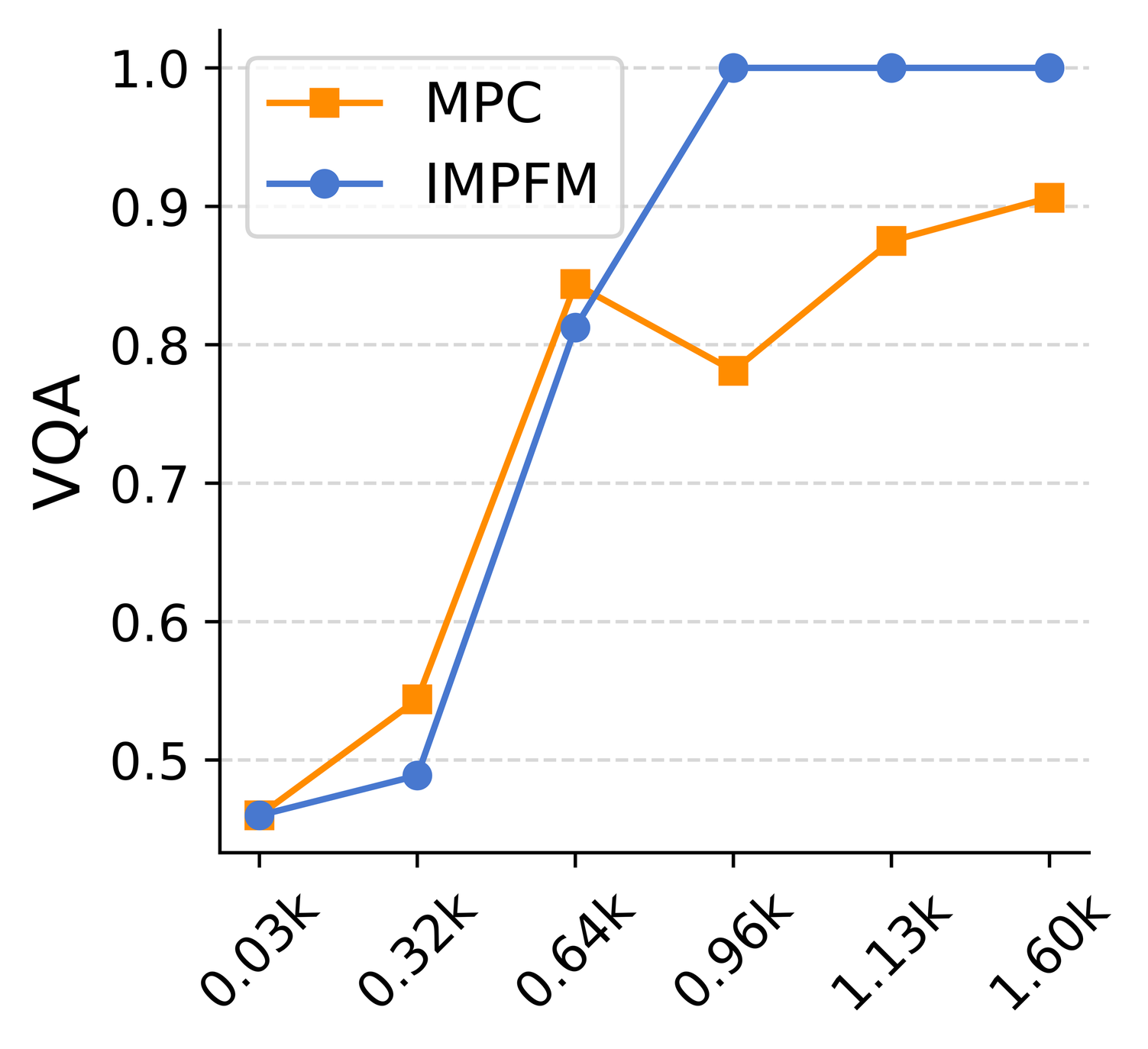}\par
        {\scriptsize\textbf{Feedback Budgets}}
    \end{minipage}
    \caption{\small{Impact of Reweighting.}}
    \label{fig:reweight}
    \vspace{-10pt}
\end{wrapfigure}

We isolate the effect of the reweighting step by comparing the full IMPFM framework against an ablated variant, referred to as Multi Particle Control (MPC), in which only the reweighting mechanism is removed. As reported in~\ref{fig:reweight}, the reweighting step proves critical — it drives faster convergence to higher reward scores (evaluated using the VQA score) across all feedback budgets. As shown in~\ref{fig:reweight}, MPC yields some suboptimal samples because it lacks explicit constraints to remain on the true data manifold. Our reweighting mechanism effectively corrects this, underscoring its necessity in the IMPFM framework.
\vspace{-2pt}
\paragraph{The Role of Sufficient Statistic (SS)}\label{sec:ss}
\begin{wrapfigure}{r}{0.33\textwidth} 
\vspace{-2ex} 
\centering
{\scriptsize\itshape \textcolor{purple}{``A red sports car.''}\par}
\vspace{1pt}
\begin{minipage}[t]{0.48\linewidth}
  \vspace{-4pt}
  \centering
  {\scriptsize\textbf{SS (IMPFM)}}\par
  \includegraphics[width=\linewidth]{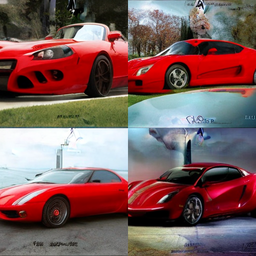}
\end{minipage}\hfill%
\begin{minipage}[t]{0.48\linewidth}
  \vspace{-4pt}
  \centering
  {\scriptsize\textbf{SS (IMPFM)}}\par
  \includegraphics[width=\linewidth]{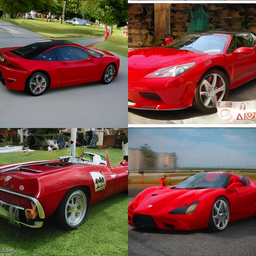}
\end{minipage}
\caption{\small{Efficacy of SS on IMPFM. (left) $\mathcal{B}$ = 20; (right)  $\mathcal{B}$ = 80}}
\label{fig:ss-double}
\vspace{-2ex} 
\end{wrapfigure}
SS serves a dual purpose: it introduces DDPM-like stochasticity to enable exploration while retaining ODE-based generative fidelity, and \emph{supports arbitrary-duration stochastic transitions essential for budget-efficient search.} 
 A direct comparison across varying feedback budgets (Fig.~\ref{fig:ss-double}) reveals that the SS-based formulation yields consistently good sample quality — even under severe feedback-budget constraints — confirming its importance for efficient feedback-driven search. Additional analysis on the role of Sufficient Statistic on search is provided in the Appendix~\ref{sec:ss-a}.
\vspace{-6pt}
\paragraph{Alignment Experimental Setting:}
To assess the effectiveness of IMPFM feedback-driven alignment problems, we focus on two challenging tasks. First, for compositional alignment, we evaluate using 200 randomly selected prompts from GenAI-Bench~\citep{jiang2024genai} that involve at least three advanced compositional skills. Second, for quantity-aware alignment, we evaluate using 200 prompts randomly sampled from the numeracy category of T2I-CompBench++~\citep{huang2025t2i}. 
For the quantity-aware alignment task, the reward is also evaluated as the negative Residual Sum of Squares (RSS) between the target object counts and the detected counts, where detections are obtained using DINO~\citep{liu2024grounding}. 
We employ an off-the-shelf consistency model~\citep{chen2025sana} as the prior and sample from the posterior ($p_{1|t}(\cdot|x_t)$) using the proposed iterative posterior sampling scheme. We provide a detailed derivation (see App.~\ref{sec:sana-sch-a}) of the reparametrization of the noise schedule used in~\citep{chen2025sana} to facilitate its integration within the IMPFM framework.
\begin{figure*}[t]
\centering
\newcommand{\triplepanel}[3]{%
  \setlength{\tabcolsep}{2pt}%
  \begin{tabular}{@{}c@{\hspace{1pt}}c@{\hspace{1pt}}c@{}}
    \includegraphics[width=0.305\linewidth]{#1} &
    \includegraphics[width=0.305\linewidth]{#2} &
    \includegraphics[width=0.305\linewidth]{#3}
  \end{tabular}%
}
\newcommand{\panelblock}[4]{%
  \begin{minipage}[t]{\linewidth}
    \centering
    \triplepanel{#2}{#3}{#4}
    \par\vspace{0.4ex}
    {\itshape ``#1''\par}
  \end{minipage}%
}
\newcommand{\topheaders}{%
  {\fontsize{9.5}{10.5}\selectfont
  \begin{tabular}{@{}c@{\hspace{1pt}}c@{\hspace{1pt}}c@{}}
    \makebox[0.305\linewidth][c]{\textbf{FKS}} &
    \makebox[0.305\linewidth][c]{\textbf{MFM}} &
    \makebox[0.305\linewidth][c]{\textbf{IMPFM}}
  \end{tabular}%
  }
}
\noindent
\begin{minipage}[t]{0.48\textwidth}
\centering
\topheaders
\end{minipage}\hfill%
\begin{minipage}[t]{0.48\textwidth}
\centering
\topheaders
\end{minipage}

\begin{minipage}[t]{0.48\textwidth}
\centering
\panelblock{Two people and two helicopters.}{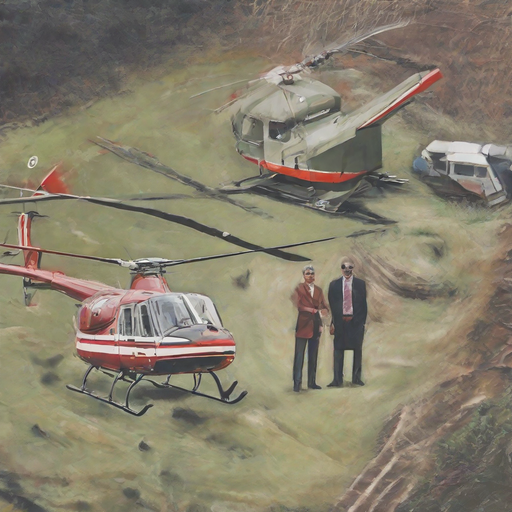}{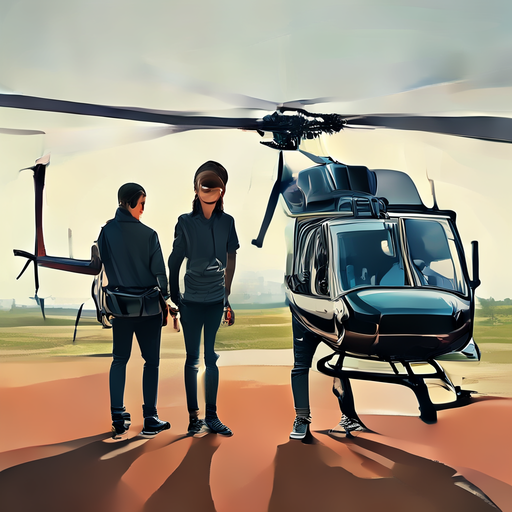}{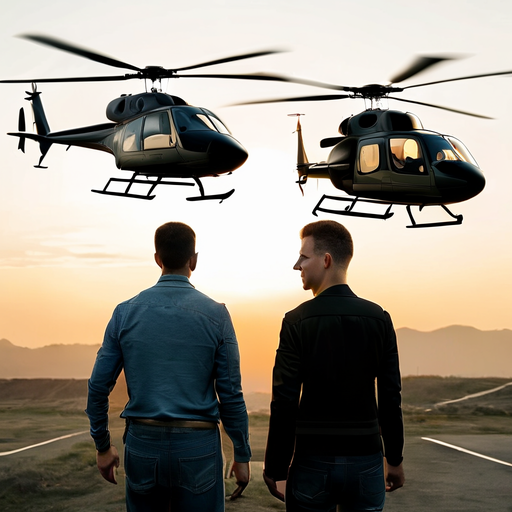}
\end{minipage}\hfill%
\begin{minipage}[t]{0.48\textwidth}
\centering
\panelblock{A baker pulling freshly baked bread out of an oven in a bakery.}{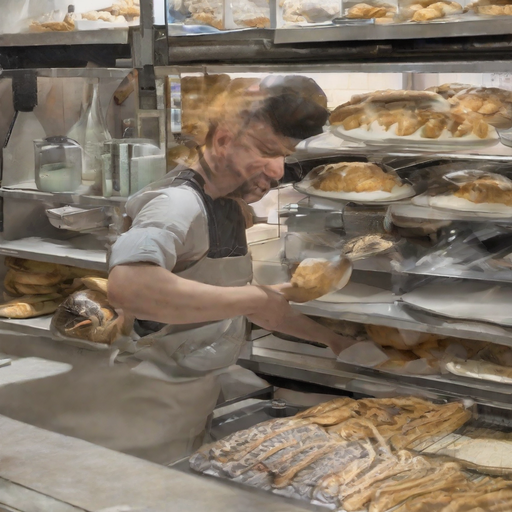}{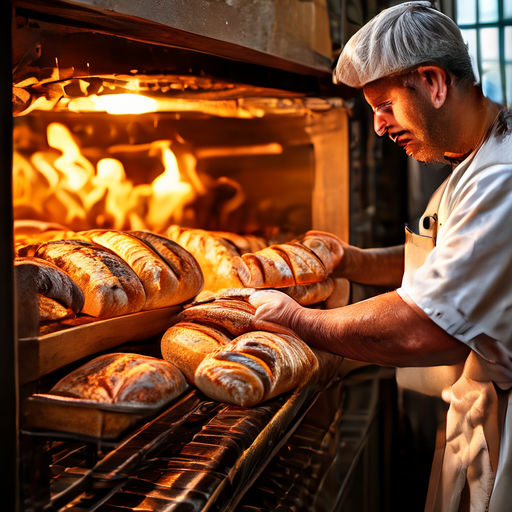}{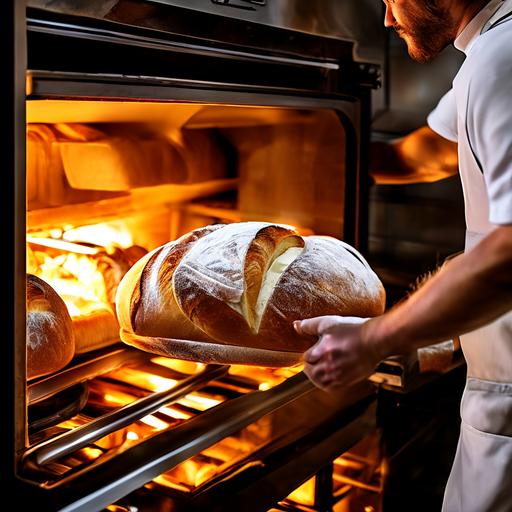}
\end{minipage}

\noindent\rule{\textwidth}{0.6pt}
\caption{\small{Alignment Visualization. (left) Quantity Prompt. (right) Compositional Prompt.}}
\label{fig:align-visu}
\vspace{-4pt}
\end{figure*}
\begin{figure*}[t]
\vspace{-4pt}
\centering

\newcommand{\legenditem}[4]{%
  \begin{tikzpicture}[baseline=-0.6ex]
    \draw[#1, line width=1.8pt, #2] (0,0) -- (0.9,0);
    \node[text=#1, fill=white, inner sep=0.6pt] at (0.45,0) {\scriptsize $#3$};
  \end{tikzpicture}\,#4%
}

\newcommand{\methodlegend}{%
  {\footnotesize
  \setlength{\tabcolsep}{4pt}%
  \begin{tabular}{@{}ccccc@{}}
    \legenditem{cyan!70!black}{solid}{\circ}{DAS} &
    \legenditem{red!75!black}{dashed}{\square}{FKS} &
    \legenditem{purple!80!black}{dotted}{\diamond}{MFM} &
    \legenditem{green!60!black}{solid}{\blacktriangledown}{\textbf{IMPFM (Our)}}
  \end{tabular}%
  }%
}

\newcommand{\barpanel}[2]{%
  \begin{minipage}[t]{0.185\textwidth}
    \centering
    \includegraphics[width=\linewidth]{#1}\par
    \vspace{-0.3ex}
    {\scriptsize #2}
  \end{minipage}%
}

\methodlegend
\vspace{0.1ex}
\noindent\rule{\textwidth}{0.2pt}

\barpanel{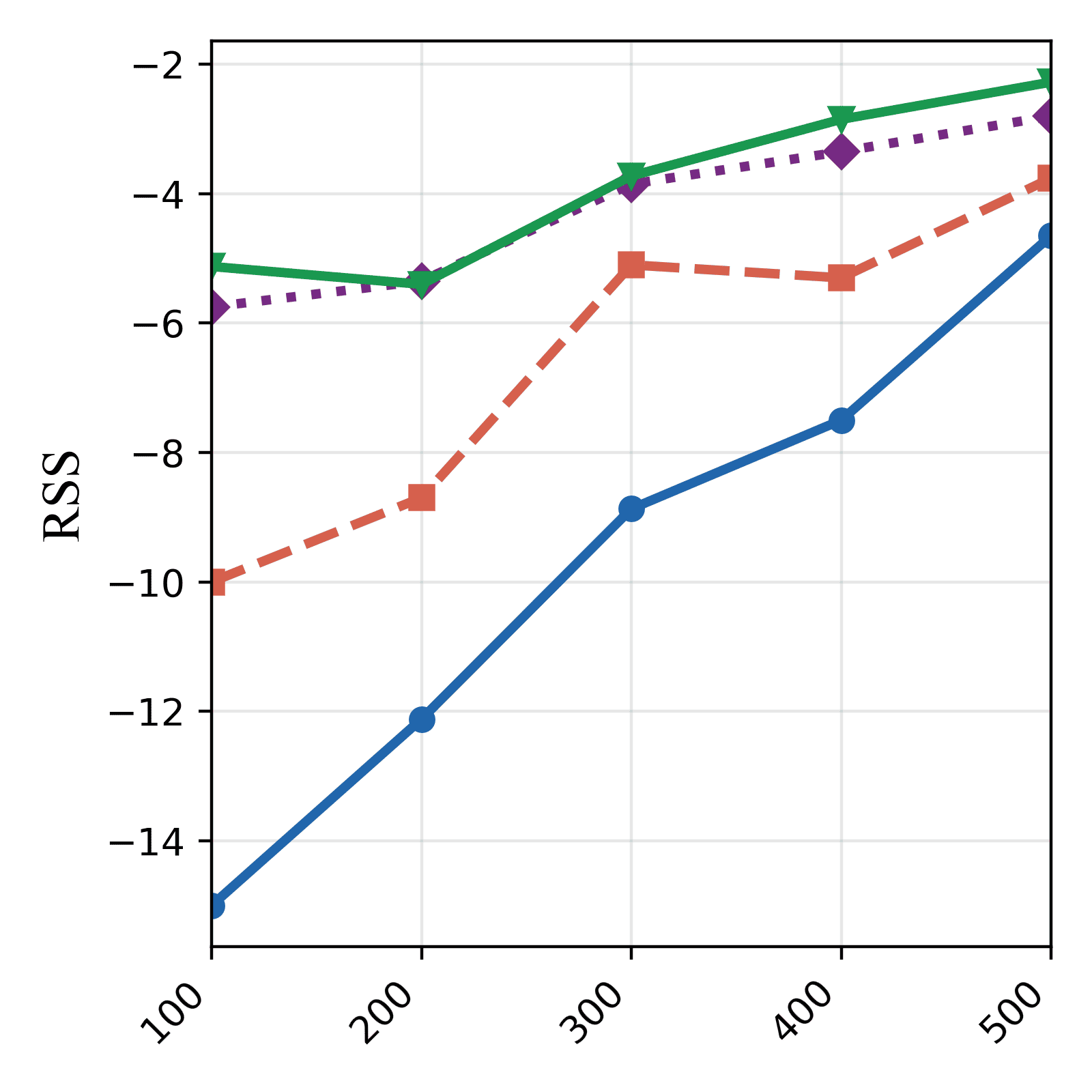}{Feedback Budgets}\hspace{0.006\textwidth}%
\barpanel{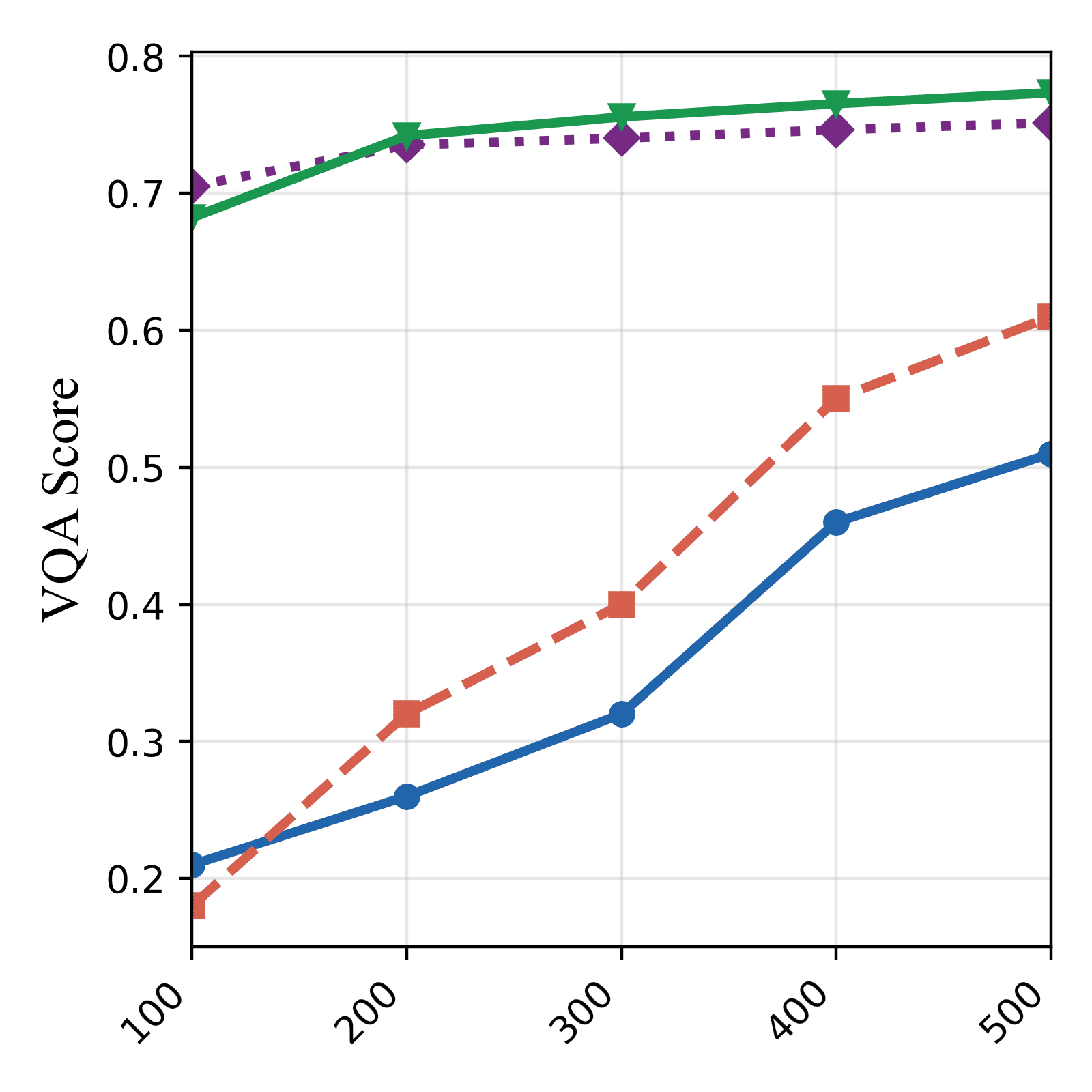}{Feedback Budgets}\hspace{0.006\textwidth}%
\barpanel{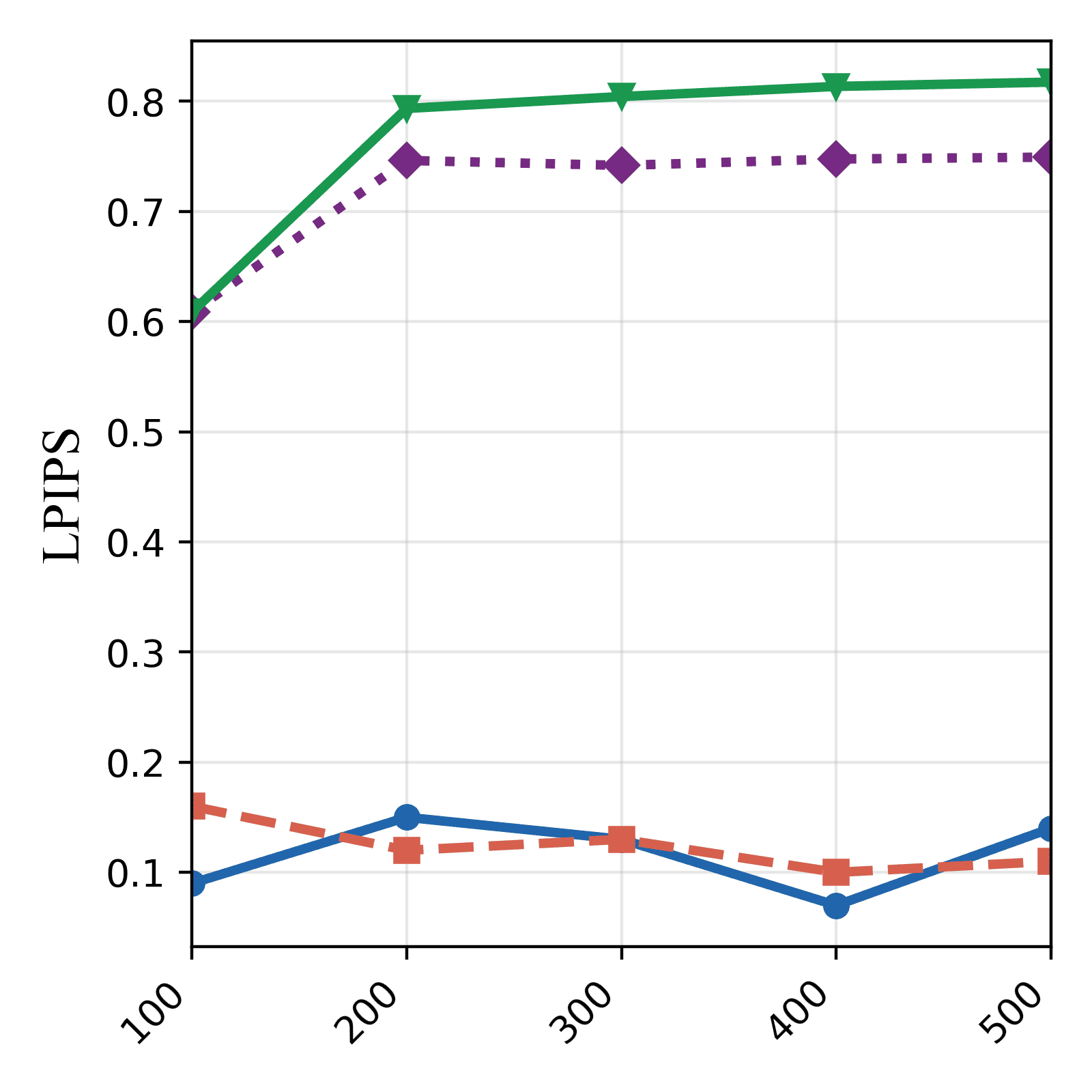}{Feedback Budgets}\hspace{0.006\textwidth}%
{\color{black}\rule{0.8pt}{2.6cm}}\hspace{0.006\textwidth}%
\barpanel{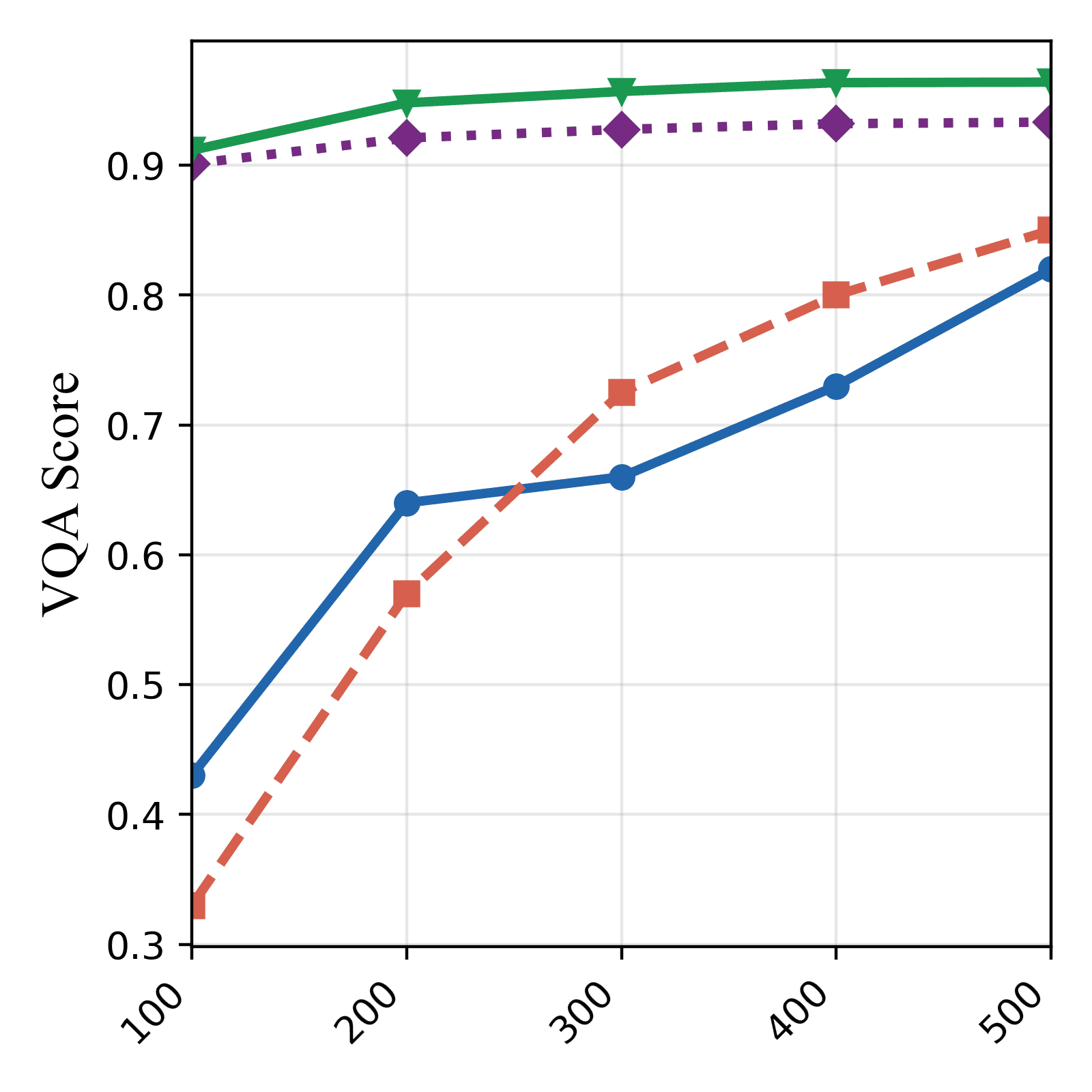}{Feedback Budgets}\hspace{0.006\textwidth}%
\barpanel{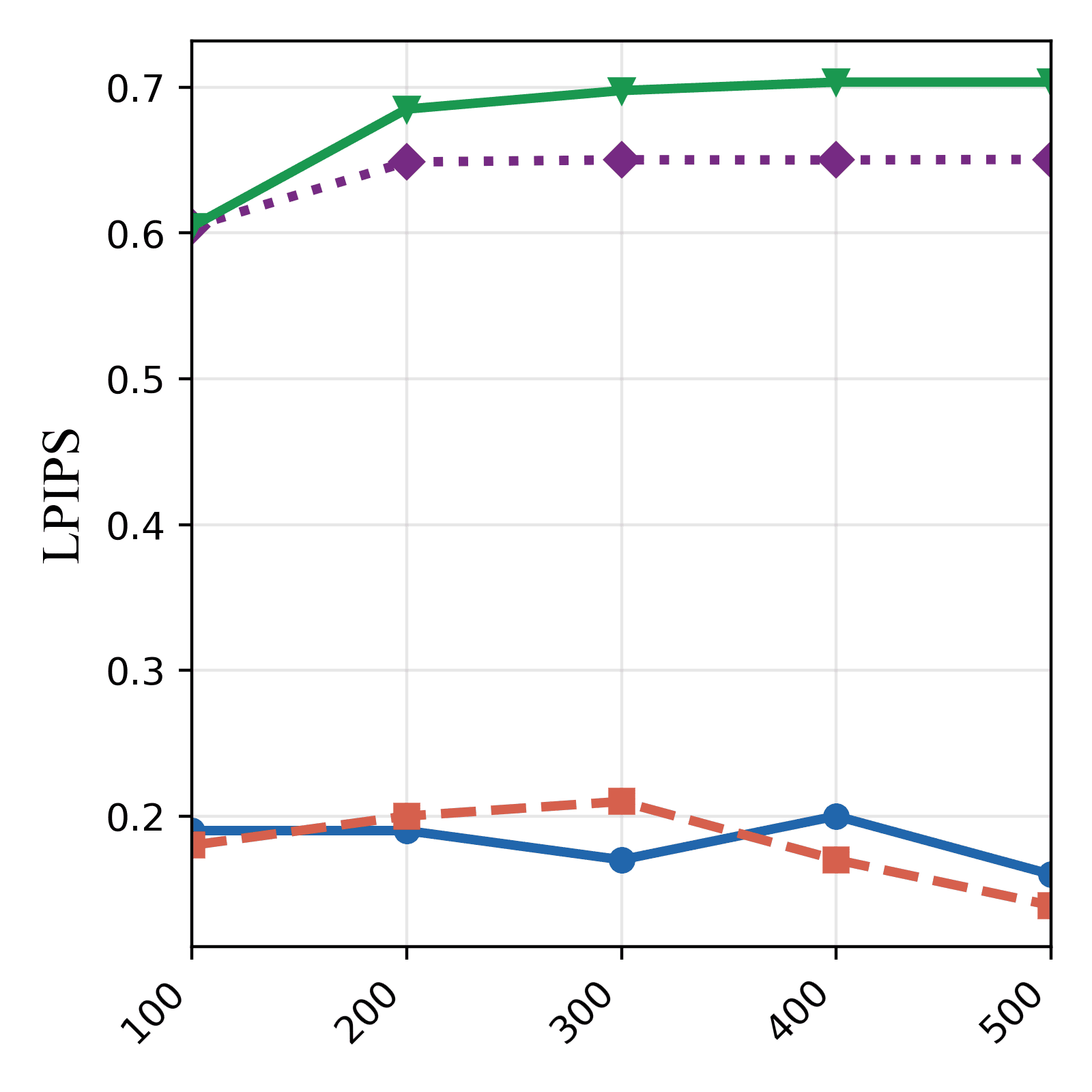}{Feedback Budgets}

\caption{\small{Analysis of Competitive Approaches on (right) Compositional, (left ) Quantity-aware Alignment.}}
\label{fig:alignment-result}
\vspace{-14pt}
\end{figure*}
\vspace{-8pt}
\paragraph{Alignment Result:}
The Alignment results (Fig.~\ref{fig:alignment-result}) reveal a similar trend. IMPFM converges to high reward scores with significantly fewer feedback interactions than competing approaches, while simultaneously maintaining greater sample diversity. The rapid convergence further underscores the effectiveness of the particle interaction mechanism, which enables maximum information utilization for both drift correction and reweighting. The sustained diversity, in turn, confirms that IMPFM does not suffer from the weight degeneracy issues that commonly afflict practical inference — a property made possible by the explicit repulsive force between particles, which actively discourages collapse and promotes population-level diversity. Qualitative results are visualized in Fig.~\ref{fig:align-visu} (See~\ref{sec:align-add-a}). Furthermore, these findings underscore the effectiveness of the iterative posterior sampling scheme. 

\vspace{-6pt}
\paragraph{Conclusion}
In this work, we introduce IMPFM, a principled, interactive Feynman-Kac corrector (FKC) sampler designed for sample-efficient feedback-driven search and alignment. Our rigorous empirical analysis demonstrates its scalability and efficacy, particularly in high-dimensional settings. 
In the future, we aim to extend the applicability of IMPFM across diverse scientific domains. 

\subsection*{AI use statement}
We used large language models (LLMs) solely as a writing assistant for language polishing and
improving clarity of presentation. The LLMs were not involved in research ideation, methodological design, experimental execution, or result analysis. All scientific contributions and substantive
writing were carried out by the authors.

\subsection*{Reproducibility statement}
To facilitate reproducibility, we have provided full implementation details, hyperparameter settings, and architectural specifications. Our complete source code, including training scripts and evaluation pipelines, is available via the zip file included in the supplementary material. Upon acceptance, we will make all code publicly available to facilitate easy reproduction.

\bibliography{main}

\begin{thebibliography}{38}
\providecommand{\natexlab}[1]{#1}
\providecommand{\url}[1]{\texttt{#1}}
\expandafter\ifx\csname urlstyle\endcsname\relax
  \providecommand{\doi}[1]{doi: #1}\else
  \providecommand{\doi}{doi: \begingroup \urlstyle{rm}\Url}\fi

\bibitem[Beirami et~al.(1879)Beirami, Agarwal, Berant, D’Amour, Eisenstein, Nagpal, and Suresh]{beirami1879theoretical}
Ahmad Beirami, Alekh Agarwal, Jonathan Berant, Alex D’Amour, Jacob Eisenstein, Chirag Nagpal, and Ananda~Theertha Suresh.
\newblock Theoretical guarantees on the best-of-n alignment policy. 2024.
\newblock \emph{URL https://api. semanticscholar. org/CorpusID}, 266741736\penalty0 (3), 1879.

\bibitem[Boffi et~al.(2025)Boffi, Albergo, and Vanden-Eijnden]{boffi2025build}
Nicholas~M Boffi, Michael~S Albergo, and Eric Vanden-Eijnden.
\newblock How to build a consistency model: Learning flow maps via self-distillation.
\newblock \emph{arXiv preprint arXiv:2505.18825}, 2025.

\bibitem[Cardoso et~al.(2023)Cardoso, Idrissi, Corff, and Moulines]{cardoso2023monte}
Gabriel Cardoso, Yazid Janati~El Idrissi, Sylvain~Le Corff, and Eric Moulines.
\newblock Monte carlo guided diffusion for bayesian linear inverse problems.
\newblock \emph{arXiv preprint arXiv:2308.07983}, 2023.

\bibitem[Chen et~al.(2025)Chen, Xue, Zhao, Yu, Paul, Chen, Cai, Han, and Xie]{chen2025sana}
Junsong Chen, Shuchen Xue, Yuyang Zhao, Jincheng Yu, Sayak Paul, Junyu Chen, Han Cai, Song Han, and Enze Xie.
\newblock Sana-sprint: One-step diffusion with continuous-time consistency distillation.
\newblock In \emph{Proceedings of the IEEE/CVF International Conference on Computer Vision}, pp.\  16185--16195, 2025.

\bibitem[Dai et~al.(2023)Dai, Li, Li, Tiong, Zhao, Wang, Li, Fung, and Hoi]{dai2023instructblip}
Wenliang Dai, Junnan Li, Dongxu Li, Anthony Tiong, Junqi Zhao, Weisheng Wang, Boyang Li, Pascale~N Fung, and Steven Hoi.
\newblock Instructblip: Towards general-purpose vision-language models with instruction tuning.
\newblock \emph{Advances in neural information processing systems}, 36:\penalty0 49250--49267, 2023.

\bibitem[Dai~Pra(1991)]{dai1991stochastic}
Paolo Dai~Pra.
\newblock A stochastic control approach to reciprocal diffusion processes.
\newblock \emph{Applied mathematics and Optimization}, 23\penalty0 (1):\penalty0 313--329, 1991.

\bibitem[Del~Moral et~al.(2006)Del~Moral, Doucet, and Jasra]{del2006sequential}
Pierre Del~Moral, Arnaud Doucet, and Ajay Jasra.
\newblock Sequential monte carlo samplers.
\newblock \emph{Journal of the Royal Statistical Society Series B: Statistical Methodology}, 68\penalty0 (3):\penalty0 411--436, 2006.

\bibitem[Denker et~al.(2024)Denker, Vargas, Padhy, Didi, Mathis, Dutordoir, Barbano, Mathieu, Komorowska, and Lio]{denker2024deft}
Alexander Denker, Francisco Vargas, Shreyas Padhy, Kieran Didi, Simon Mathis, Vincent Dutordoir, Riccardo Barbano, Emile Mathieu, Urszula~J Komorowska, and Pietro Lio.
\newblock Deft: Efficient fine-tuning of diffusion models by learning the generalised $ h $-transform.
\newblock \emph{Advances in Neural Information Processing Systems}, 37:\penalty0 19636--19682, 2024.

\bibitem[Dong et~al.(2023)Dong, Xiong, Goyal, Zhang, Chow, Pan, Diao, Zhang, Shum, and Zhang]{dong2023raft}
Hanze Dong, Wei Xiong, Deepanshu Goyal, Yihan Zhang, Winnie Chow, Rui Pan, Shizhe Diao, Jipeng Zhang, Kashun Shum, and Tong Zhang.
\newblock Raft: Reward ranked finetuning for generative foundation model alignment.
\newblock \emph{arXiv preprint arXiv:2304.06767}, 2023.

\bibitem[Dou \& Song(2024)Dou and Song]{dou2024diffusion}
Zehao Dou and Yang Song.
\newblock Diffusion posterior sampling for linear inverse problem solving: A filtering perspective.
\newblock In \emph{The Twelfth International Conference on Learning Representations}, 2024.

\bibitem[Holderrieth et~al.(2025)Holderrieth, Singer, Jaakkola, Chen, Lipman, and Karrer]{holderrieth2025glass}
Peter Holderrieth, Uriel Singer, Tommi Jaakkola, Ricky~TQ Chen, Yaron Lipman, and Brian Karrer.
\newblock Glass flows: Transition sampling for alignment of flow and diffusion models.
\newblock \emph{arXiv preprint arXiv:2509.25170}, 2025.

\bibitem[Holderrieth et~al.(2026)Holderrieth, Chen, Eyring, Shah, Anantharaman, He, Akata, Jaakkola, Boffi, and Simchowitz]{holderrieth2026diamond}
Peter Holderrieth, Douglas Chen, Luca Eyring, Ishin Shah, Giri Anantharaman, Yutong He, Zeynep Akata, Tommi Jaakkola, Nicholas~Matthew Boffi, and Max Simchowitz.
\newblock Diamond maps: Efficient reward alignment via stochastic flow maps.
\newblock \emph{arXiv preprint arXiv:2602.05993}, 2026.

\bibitem[Huang et~al.(2025)Huang, Duan, Sun, Xie, Li, and Liu]{huang2025t2i}
Kaiyi Huang, Chengqi Duan, Kaiyue Sun, Enze Xie, Zhenguo Li, and Xihui Liu.
\newblock T2i-compbench++: An enhanced and comprehensive benchmark for compositional text-to-image generation.
\newblock \emph{IEEE Transactions on Pattern Analysis and Machine Intelligence}, 47\penalty0 (5):\penalty0 3563--3579, 2025.

\bibitem[Jain et~al.(2025)Jain, Sareen, Pedramfar, and Ravanbakhsh]{jain2025diffusion}
Vineet Jain, Kusha Sareen, Mohammad Pedramfar, and Siamak Ravanbakhsh.
\newblock Diffusion tree sampling: Scalable inference-time alignment of diffusion models.
\newblock \emph{arXiv preprint arXiv:2506.20701}, 2025.

\bibitem[Jiang et~al.(2024)Jiang, Ku, Li, Ni, Sun, Fan, and Chen]{jiang2024genai}
Dongfu Jiang, Max Ku, Tianle Li, Yuansheng Ni, Shizhuo Sun, Rongqi Fan, and Wenhu Chen.
\newblock Genai arena: An open evaluation platform for generative models.
\newblock \emph{Advances in Neural Information Processing Systems}, 37:\penalty0 79889--79908, 2024.

\bibitem[Kim et~al.(2025)Kim, Kim, and Park]{kim2025test}
Sunwoo Kim, Minkyu Kim, and Dongmin Park.
\newblock Test-time alignment of diffusion models without reward over-optimization.
\newblock \emph{arXiv preprint arXiv:2501.05803}, 2025.

\bibitem[Kirstain et~al.(2023)Kirstain, Polyak, Singer, Matiana, Penna, and Levy]{kirstain2023pick}
Yuval Kirstain, Adam Polyak, Uriel Singer, Shahbuland Matiana, Joe Penna, and Omer Levy.
\newblock Pick-a-pic: An open dataset of user preferences for text-to-image generation.
\newblock \emph{Advances in neural information processing systems}, 36:\penalty0 36652--36663, 2023.

\bibitem[Lee et~al.(2025)Lee, Jeha, Frellsen, Lio, Albergo, and Vargas]{lee2025debiasing}
Cheuk~Kit Lee, Paul Jeha, Jes Frellsen, Pietro Lio, Michael~Samuel Albergo, and Francisco Vargas.
\newblock Debiasing guidance for discrete diffusion with sequential monte carlo.
\newblock \emph{arXiv preprint arXiv:2502.06079}, 2025.

\bibitem[Lin et~al.(2024)Lin, Pathak, Li, Li, Xia, Neubig, Zhang, and Ramanan]{lin2024evaluating}
Zhiqiu Lin, Deepak Pathak, Baiqi Li, Jiayao Li, Xide Xia, Graham Neubig, Pengchuan Zhang, and Deva Ramanan.
\newblock Evaluating text-to-visual generation with image-to-text generation.
\newblock In \emph{European Conference on Computer Vision}, pp.\  366--384. Springer, 2024.

\bibitem[Lipman et~al.(2022)Lipman, Chen, Ben-Hamu, Nickel, and Le]{lipman2022flow}
Yaron Lipman, Ricky~TQ Chen, Heli Ben-Hamu, Maximilian Nickel, and Matt Le.
\newblock Flow matching for generative modeling.
\newblock \emph{arXiv preprint arXiv:2210.02747}, 2022.

\bibitem[Liu(2017)]{liu2017stein}
Qiang Liu.
\newblock Stein variational gradient descent as gradient flow.
\newblock \emph{Advances in neural information processing systems}, 30, 2017.

\bibitem[Liu \& Wang(2016)Liu and Wang]{liu2016stein}
Qiang Liu and Dilin Wang.
\newblock Stein variational gradient descent: A general purpose bayesian inference algorithm.
\newblock \emph{Advances in neural information processing systems}, 29, 2016.

\bibitem[Liu et~al.(2016)Liu, Lee, and Jordan]{liu2016kernelized}
Qiang Liu, Jason Lee, and Michael Jordan.
\newblock A kernelized stein discrepancy for goodness-of-fit tests.
\newblock In \emph{International conference on machine learning}, pp.\  276--284. PMLR, 2016.

\bibitem[Liu et~al.(2024)Liu, Zeng, Ren, Li, Zhang, Yang, Jiang, Li, Yang, Su, et~al.]{liu2024grounding}
Shilong Liu, Zhaoyang Zeng, Tianhe Ren, Feng Li, Hao Zhang, Jie Yang, Qing Jiang, Chunyuan Li, Jianwei Yang, Hang Su, et~al.
\newblock Grounding dino: Marrying dino with grounded pre-training for open-set object detection.
\newblock In \emph{European conference on computer vision}, pp.\  38--55. Springer, 2024.

\bibitem[Nakano et~al.(2022)Nakano, Hilton, Balaji, Wu, Ouyang, Kim, Hesse, Jain, Kosaraju, Saunders, et~al.]{nakano2022webgpt}
Reiichiro Nakano, Jacob Hilton, Suchir Balaji, Jeff Wu, Long Ouyang, Christina Kim, Christopher Hesse, Shantanu Jain, Vineet Kosaraju, William Saunders, et~al.
\newblock Webgpt: Browser-assisted question-answering with human feedback, 2022.
\newblock \emph{URL https://arxiv. org/abs/2112.09332}, 35, 2022.

\bibitem[Ouyang et~al.(2022)Ouyang, Wu, Jiang, Almeida, Wainwright, Mishkin, Zhang, Agarwal, Slama, Ray, et~al.]{ouyang2022training}
Long Ouyang, Jeffrey Wu, Xu~Jiang, Diogo Almeida, Carroll Wainwright, Pamela Mishkin, Chong Zhang, Sandhini Agarwal, Katarina Slama, Alex Ray, et~al.
\newblock Training language models to follow instructions with human feedback.
\newblock \emph{Advances in neural information processing systems}, 35:\penalty0 27730--27744, 2022.

\bibitem[Phillips et~al.(2024)Phillips, Dau, Hutchinson, De~Bortoli, Deligiannidis, and Doucet]{phillips2024particle}
Angus Phillips, Hai-Dang Dau, Michael~John Hutchinson, Valentin De~Bortoli, George Deligiannidis, and Arnaud Doucet.
\newblock Particle denoising diffusion sampler.
\newblock \emph{arXiv preprint arXiv:2402.06320}, 2024.

\bibitem[Potaptchik et~al.(2026)Potaptchik, Saravanan, Mammadov, Prat, Albergo, and Teh]{potaptchik2026meta}
Peter Potaptchik, Adhi Saravanan, Abbas Mammadov, Alvaro Prat, Michael~S Albergo, and Yee~Whye Teh.
\newblock Meta flow maps enable scalable reward alignment.
\newblock \emph{arXiv preprint arXiv:2601.14430}, 2026.

\bibitem[Singhal et~al.(2025)Singhal, Horvitz, Teehan, Ren, Yu, McKeown, and Ranganath]{singhal2025general}
Raghav Singhal, Zachary Horvitz, Ryan Teehan, Mengye Ren, Zhou Yu, Kathleen McKeown, and Rajesh Ranganath.
\newblock A general framework for inference-time scaling and steering of diffusion models.
\newblock \emph{arXiv preprint arXiv:2501.06848}, 2025.

\bibitem[Skreta et~al.(2025)Skreta, Akhound-Sadegh, Ohanesian, Bondesan, Aspuru-Guzik, Doucet, Brekelmans, Tong, and Neklyudov]{skreta2025feynman}
Marta Skreta, Tara Akhound-Sadegh, Viktor Ohanesian, Roberto Bondesan, Al{\'a}n Aspuru-Guzik, Arnaud Doucet, Rob Brekelmans, Alexander Tong, and Kirill Neklyudov.
\newblock Feynman-kac correctors in diffusion: Annealing, guidance, and product of experts.
\newblock \emph{arXiv preprint arXiv:2503.02819}, 2025.

\bibitem[Song et~al.(2020{\natexlab{a}})Song, Meng, and Ermon]{song2020denoising}
Jiaming Song, Chenlin Meng, and Stefano Ermon.
\newblock Denoising diffusion implicit models.
\newblock \emph{arXiv preprint arXiv:2010.02502}, 2020{\natexlab{a}}.

\bibitem[Song et~al.(2020{\natexlab{b}})Song, Sohl-Dickstein, Kingma, Kumar, Ermon, and Poole]{song2020score}
Yang Song, Jascha Sohl-Dickstein, Diederik~P Kingma, Abhishek Kumar, Stefano Ermon, and Ben Poole.
\newblock Score-based generative modeling through stochastic differential equations.
\newblock \emph{arXiv preprint arXiv:2011.13456}, 2020{\natexlab{b}}.

\bibitem[Stiennon et~al.(2020)Stiennon, Ouyang, Wu, Ziegler, Lowe, Voss, Radford, Amodei, and Christiano]{stiennon2020learning}
Nisan Stiennon, Long Ouyang, Jeffrey Wu, Daniel Ziegler, Ryan Lowe, Chelsea Voss, Alec Radford, Dario Amodei, and Paul~F Christiano.
\newblock Learning to summarize with human feedback.
\newblock \emph{Advances in neural information processing systems}, 33:\penalty0 3008--3021, 2020.

\bibitem[Touvron et~al.(2023)Touvron, Martin, Stone, Albert, Almahairi, Babaei, Bashlykov, Batra, Bhargava, Bhosale, et~al.]{touvron2023llama}
Hugo Touvron, Louis Martin, Kevin Stone, Peter Albert, Amjad Almahairi, Yasmine Babaei, Nikolay Bashlykov, Soumya Batra, Prajjwal Bhargava, Shruti Bhosale, et~al.
\newblock Llama 2: Open foundation and fine-tuned chat models.
\newblock \emph{arXiv preprint arXiv:2307.09288}, 2023.

\bibitem[Trippe et~al.(2022)Trippe, Yim, Tischer, Baker, Broderick, Barzilay, and Jaakkola]{trippe2022diffusion}
Brian~L Trippe, Jason Yim, Doug Tischer, David Baker, Tamara Broderick, Regina Barzilay, and Tommi Jaakkola.
\newblock Diffusion probabilistic modeling of protein backbones in 3d for the motif-scaffolding problem.
\newblock \emph{arXiv preprint arXiv:2206.04119}, 2022.

\bibitem[Uehara et~al.(2024)Uehara, Zhao, Black, Hajiramezanali, Scalia, Diamant, Tseng, Levine, and Biancalani]{uehara2024feedback}
Masatoshi Uehara, Yulai Zhao, Kevin Black, Ehsan Hajiramezanali, Gabriele Scalia, Nathaniel~Lee Diamant, Alex~M Tseng, Sergey Levine, and Tommaso Biancalani.
\newblock Feedback efficient online fine-tuning of diffusion models.
\newblock \emph{arXiv preprint arXiv:2402.16359}, 2024.

\bibitem[Wu et~al.(2023)Wu, Trippe, Naesseth, Blei, and Cunningham]{wu2023practical}
Luhuan Wu, Brian Trippe, Christian Naesseth, David Blei, and John~P Cunningham.
\newblock Practical and asymptotically exact conditional sampling in diffusion models.
\newblock \emph{Advances in Neural Information Processing Systems}, 36:\penalty0 31372--31403, 2023.

\bibitem[Xu et~al.(2023)Xu, Liu, Wu, Tong, Li, Ding, Tang, and Dong]{xu2023imagereward}
Jiazheng Xu, Xiao Liu, Yuchen Wu, Yuxuan Tong, Qinkai Li, Ming Ding, Jie Tang, and Yuxiao Dong.
\newblock Imagereward: Learning and evaluating human preferences for text-to-image generation.
\newblock \emph{Advances in Neural Information Processing Systems}, 36:\penalty0 15903--15935, 2023.

\end{thebibliography}
\bibliographystyle{iclr2027_conference}

\newpage
\appendix
\newcommand{\ldotsfill}{\leavevmode\leaders\hbox to .5em{\hss.\hss}\hfill\kern0pt}

\section*{Sampling Meets Interaction: Sequentially Controlled Multi-Particle Flow-Maps For Efficient Inference-Time Search(Appendix)}
\vspace{3pt}

\section*{Appendix Table of Contents}
\hrule
\vspace{0.4cm}

\begin{description}
    \item[\textbf{Appendix: Theoretical Derivations}] \hfill \textbf{Page}
    \begin{itemize}
        \item[1] Derivation of Exploration-Augmented Drift Correction via Interactive Particle Dynamics \dotfill \pageref{sec:prop1-a}
        \item[2] IMPFM Yields Fynman-Kac Corector to KL-Tilted Target \dotfill \pageref{sec:prop2-a}
    \end{itemize}

    \item[\textbf{Appendix: Additional Quantitative Results on Search Task}] \hfill
    \begin{itemize}
        \item[1] Performance Comparison in Longer Horizon Search Tasks \dotfill \pageref{sec:se-quant}
    \end{itemize}

    \item[\textbf{Appendix: Additional Ablation Studies and Qualitative Analysis of IMPFM}] \hfill
    \begin{itemize}
        \item[1] Additional Analysis on the Impact of the Interaction-Aware Feynman-Kac Corrector \dotfill \pageref{sec:int-fkc}
        \item[2] Visualizing the Impact of Sufficient Statistic
        \dotfill 
        \pageref{sec:ss-a}
        \item[3] Additional Qualitative Analysis: Taming Weight Degeneracy Through Particle Interaction-Guided Drift Correction
        \dotfill \pageref{sec:wd-a}
    \end{itemize}

    \item[\textbf{Appendix: Additional Details and Results on Search}] \hfill
    \begin{itemize}
        \item[1] Details of Class and Prompt Detail in the Search Experiment \dotfill \pageref{sec:se-class}
        \item[2] Additional Comparative Visualization on Search Task \dotfill \pageref{sec:se-vis-a}
    \end{itemize}

    \item[\textbf{Appendix: Additional Implementation Details, Background, and Qualitative Results on Alignment Task}] \hfill
    \begin{itemize}
        \item[1] Details about the Evaluation Dataset for Alignment Tasks \dotfill \pageref{sec:align-eval-data}
        \item[2] Implementation Details \dotfill \pageref{sec:imp-a}
        \item[3] Derivation of Noise Schedule to Adapt IMPFM on Alignment Tasks \dotfill \pageref{sec:sana-sch-a}
        \item[4] Sufficient Statistic Computation  \dotfill \pageref{sec:ss-comp}
        \item[5] Extended Related Work \dotfill \pageref{sec:ex-rel}
        \item[6] Background: Feynman-Kac Correctors \dotfill \pageref{sec:fkc-back}
        \item[7] Additional Visualizations on the Alignment Task \dotfill \pageref{sec:align-add-a}
    \end{itemize}

\end{description}

\vspace{0.2cm}
\hrule

\newpage

\section*{Sampling Meets Interaction: Sequentially Controlled Multi-Particle Flow-Maps For Efficient Inference-Time Search (Appendix)}
\vspace{3pt}

\section{Proof of Proposition 4.2}\label{sec:prop1-a}

\begin{tcolorbox}[colback=blue!5!white, colframe=gray!80!black, title={\textbf{Proposition 4.1} (Exploration-Augmented Drift Correction via Interactive Particle Dynamics))}]
Let $q(x)$ denote the empirical distribution induced by the interacting particles, and define the Boltzmann distribution as $p_{\text{boltz}}(x) \propto e^{V(x)}$. The steepest descent direction that minimizes the Kullback-Leibler divergence, i.e., $\nabla_x D_{\text{KL}}(q \parallel p_{\text{boltz}})$ within the unit ball of a Reproducing Kernel Hilbert Space (RKHS) is given by:
\begin{equation}
    \mathbb{E}_{x^i \sim q} \left[ \nabla_{x^i} V(x^{i}) + \frac{1}{n-1} \sum_{j=1, j \neq i}^n \left[ k(x^j, x^{i}) \nabla_{x^j} V(x^j) + \nabla_{x^j} k(x^j, x^{i}) \right] \right]
\end{equation}
\end{tcolorbox} 

\begin{proof}

We aim to derive the steepest direction minimizing the KL-Divergence within the unit ball of an RKHS.
\paragraph{Step 1: Setup the Perturbation} 
Let $q(x)$ be the current distribution of particles. We apply a smooth transformation $T(x) = x + \epsilon \phi(x)$, where $\epsilon > 0$ is a small step size and $\phi \in \mathcal{H}^d$ is the velocity field in a Reproducing Kernel Hilbert Space (RKHS) with kernel $k(\cdot, \cdot)$. 
Let $q_{[\epsilon\phi]}$ denote the pushforward distribution of $q$ under $T$.
\paragraph{Step 2: Derivative of the KL Divergence}
We want to find the direction $\phi$ that maximizes the decrease of $D_{KL}(q_{[\epsilon\phi]} \parallel p_{boltz})$. By the results of~\cite{liu2016kernelized}, the first variation of the KL divergence at $\epsilon = 0$ evaluates to the expectation of the Stein operator:$$\left. \frac{d}{d\epsilon} D_{KL}(q_{[\epsilon\phi]} \parallel p_{boltz}) \right|_{\epsilon=0} = - \mathbb{E}_{x \sim q} \big[ \text{trace}(\mathcal{A}_{p_{boltz}} \phi(x)) \big]$$where the Stein operator $\mathcal{A}_p$ is defined as:$$\mathcal{A}_{p_{boltz}} \phi(x) = \phi(x)^\top \nabla_x \log p_{boltz}(x) + \nabla_x \cdot \phi(x)$$Given $p_{boltz}(x) \propto e^{V(x)}$, we have $\nabla_x \log p_{boltz}(x) = \nabla_x V(x)$. 
Substituting this in yields:$$\left. \frac{d}{d\epsilon} D_{KL} \right|_{\epsilon=0} = - \mathbb{E}_{x \sim q} \big[ \phi(x)^\top \nabla_x V(x) + \nabla_x \cdot \phi(x) \big]$$
\paragraph{Step 3: Applying the RKHS Reproducing Property}
By the reproducing property of the RKHS, any $\phi \in \mathcal{H}^d$ satisfies $\phi(x) = \langle \phi, k(x, \cdot) \rangle_{\mathcal{H}}$. We can pass the linear inner product outside the expectation:$$\mathbb{E}_{x \sim q} \big[ \phi(x)^\top \nabla_x V(x) + \nabla_x \cdot \phi(x) \big] = \left\langle \phi(\cdot), \; \mathbb{E}_{x \sim q} \big[ k(x, \cdot) \nabla_x V(x) + \nabla_x k(x, \cdot) \big] \right\rangle_{\mathcal{H}}$$
\paragraph{Step 4: The Steepest Descent Direction}
To find the steepest descent direction within the unit ball of the RKHS, we solve:$$\phi^* = \arg\max_{\|\phi\|_{\mathcal{H}} \le 1} \left( -\left. \frac{d}{d\epsilon} D_{KL} \right|_{\epsilon=0} \right)$$By the Cauchy-Schwarz inequality in Hilbert spaces ($\langle \phi, u \rangle \le \|\phi\| \|u\|$), the linear functional is maximized when $\phi$ is strictly proportional to the second argument of the inner product. Therefore, the optimal vector field is:$$\phi^*(\cdot) = \mathbb{E}_{x \sim q} \big[ k(x, \cdot) \nabla_x V(x) + \nabla_x k(x, \cdot) \big]$$
\paragraph{Step 5: Empirical Particle Approximation}
Given $n$ interacting particles $\{x^1, \dots, x^n\}$, evaluating the optimal velocity field $\phi^*$ at a specific particle $x^i$ using the empirical distribution $q(x) \approx \frac{1}{n} \sum_{j=1}^n \delta(x - x^j)$ yields:$$\phi^*(x^i) = \frac{1}{n} \sum_{j=1}^n \big[ k(x^j, x^i) \nabla_{x_j} V(x^j) + \nabla_{x_j} k(x^j, x^i) \big]$$
To match the proposed formulation in Proposition 4.1, we isolate the self-interaction term ($j=i$). For standard translation-invariant kernels (like the RBF kernel), $k(x^i, x^i) = 1$ and $\nabla_{x_i} k(x^i, x^i) = 0$.
By separating the $j=i$ term and replacing the generic $\frac{1}{n}$ weight of the interaction sum with the unbiased leave-one-out U-statistic estimator $\frac{1}{n-1}$ for the other particles, the update direction at $x^i$ becomes exactly:$$\nabla_{x_i} V(x^i) + \frac{1}{n-1} \sum_{j \neq i} \big[ k(x^j, x^i) \nabla_{x_j} V(x^j) + \nabla_{x_j} k(x^j, x^i) \big]$$Taking the expectation over all particles $x^i \sim q$ yields the exact expression as stated in Proposition 4.1. 

\end{proof}
\section{Proof of Proposition 4.3}\label{sec:prop2-a}

\begin{tcolorbox}[colback=blue!5!white, colframe=gray!80!black, title={\textbf{Proposition 4.3} (IMPFM Yields Fynman-Kac Corector to KL-Tilted Target)}]\label{prop:3}
Consider a Flow model with drift $b_t(x)$ trained to sample from $q_t(x)$. Sampling from the KL-tilted marginals $p_t^{\text{kl}}(x) \propto q_t(x) \exp\Big( -D_{\text{KL}}\big( q_t(x) \parallel p_{\text{boltz}}(x) \big) \Big)$ is performed by an weighted SDE with an equivalent corresponding weighted ODE:
\begin{align}
    dx^i_t &= b_t(x^i_t)\>dt  + \frac{\sigma_t^2}{2}\frac{1}{n} \sum_{j=1}^n \left[ k(x^j_t, x^{i}_t) \nabla_{x^j_t} V(x^j_t) \> dt + \nabla_{x^j_t} k(x^j_t, x^{i}_t) \> dt \right]
\end{align}
\vspace{-15pt}
\begin{align}
    dw^i_t = \frac{1}{n} \sum_{j=1}^n \left[ \bm{\langle} k(x_t^j, x_t^{i}) \nabla_{x_t^j} V(x_t^j), b_t(x_t^i) \bm{\rangle} \> dt + \bm{\langle} \nabla_{x_t^j} k(x_t^j, x_t^{i}), b_t(x_t^i) \bm{\rangle}\> dt \right] 
\end{align}
\end{tcolorbox} 

\begin{proof}

We first show the weighted SDE form that corresponds to sampling from $p_t^{kl}(x)$, then we convert the resulting weighted SDE to the corresponding weighted ODE. 
\paragraph{Step 1:}
Assume $f^{kl}(x) = -D_{KL}(q_t(x) || p_{boltz}(x))$.
The generation process can be defined as the family of denoising SDEs in the opposite time direction as below: 
\begin{equation}\label{eq:sde_1}
    dx_t = \left( -f_t(x_t) + \sigma_t^2 \nabla \log q_t(x_t) \right) dt + \sigma_t dW_t
\end{equation}
The SDE in Eq.~\ref{eq:sde_1} can be viewed as the composition of a flow and diffusion terms, where the corresponding
Fokker-Planck PDE describes the combined evolution:
\begin{equation}
    \frac{\partial q_t(x)}{\partial t} = -\underbrace{\langle \nabla, q_t(x) v_t(x) \rangle}_{\text{Continuity Eqn. }} + \underbrace{\frac{\sigma_t^2}{2} \Delta q_t(x)}_{\text{Diffusion Eqn.}}.
\end{equation}
We want to find the PDE for the reward-tilted density:
\begin{equation}\label{eq:marginal_kl}
    \hat{p_t}^{kl}(x) = \frac{q_t(x) \exp(\beta_t f^{kl}(x))}{\int dx \, q_t(x) \exp(\beta_t f^{kl}(x))}.
\end{equation}
By applying the logarithm and then taking the partial derivative with respect to $t$, we get:
\begin{equation}
    \frac{\partial}{\partial t} \log \hat{p_t}^{kl}(x) = \frac{\partial}{\partial t} \log q_t(x) + \frac{\partial \beta_t}{\partial t} f^{kl}(x) - \int dx \, \hat{p_t}^{kl}(x) \left[ \frac{\partial}{\partial t} \log q_t(x) + \frac{\partial \beta_t}{\partial t} f^{kl}(x) \right].
\end{equation}
For the first term, we have:
\begin{align}
    \frac{\partial}{\partial t} \log q_t(x) &= -\langle \nabla, v_t(x) \rangle - \langle \nabla \log q_t(x), v_t(x) \rangle + \frac{\sigma_t^2}{2} \Delta \log q_t(x) + \frac{\sigma_t^2}{2} \|\nabla \log q_t(x)\|^2 \nonumber \\
    &= -\langle \nabla, v_t(x) \rangle - \langle \nabla \log \hat{p_t}^{kl}(x), v_t(x) \rangle + \frac{\sigma_t^2}{2} \Delta \log \hat{p_t}^{kl}(x) + \frac{\sigma_t^2}{2} \|\nabla \log \hat{p_t}^{kl}(x)\|^2 \nonumber \\
    &\quad + \left\langle \beta_t \nabla f^{kl}(x), v_t(x) - \sigma_t^2 \nabla \log q_t(x) - \frac{\sigma_t^2}{2} \beta_t \nabla f^{kl}(x) \right\rangle - \beta_t \frac{\sigma_t^2}{2} \Delta f^{kl}(x).
\end{align}
The last equality is obtained using the fact that $$\nabla \log q_t(x) = \nabla \log \hat{p_t}^{kl}(x) - \beta_t \nabla f^{kl}(x) \quad$$
and $$\Delta \log q_t(x) = \Delta \log \hat{p_t}^{kl}(x) - \beta_t \Delta f^{kl}(x) \quad $$ Both of which follow from~\ref{eq:marginal_kl}.

Thus, we have:
\begin{equation}
    \frac{\partial \hat{p_t}^{kl}(x)}{\partial t} = -\langle \nabla, \hat{p_t}^{kl}(x) v_t(x) \rangle + \frac{\sigma_t^2}{2} \Delta \hat{p_t}^{kl}(x) + \hat{p_t}^{kl}(x) \big( g_t(x) - \mathbb{E}_{\hat{p_t}^{kl}(x)}[g_t(x)] \big),
\end{equation}
\begin{equation}
    g_t(x) = \left\langle \beta_t \nabla f^{kl}(x), v_t(x) - \sigma_t^2 \nabla \log q_t(x) - \frac{\sigma_t^2}{2} \beta_t \nabla f^{kl}(x) \right\rangle - \beta_t \frac{\sigma_t^2}{2} \Delta f^{kl}(x) + \frac{\partial \beta_t}{\partial t} f^{kl}(x).
\end{equation}
Furthermore, we can add the gradient of the tilted function as an additional drift term $a \nabla f^{kl}(x)$, i.e.,
\begin{equation}
    \frac{\partial \hat{p_t}^{kl}(x)}{\partial t} = -\langle \nabla, \hat{p_t}^{kl}(x) (v_t(x) + a \nabla f^{kl}(x)) \rangle 
    + \frac{\sigma_t^2}{2} \Delta \hat{p_t}^{kl}(x) + \hat{p_t}^{kl}(x) \big( g_t(x) - \mathbb{E}_{\hat{p_t}^{kl}(x)}[g_t(x)] \big),
\end{equation}
\begin{align}
    g_t(x) &= a \Delta f^{kl}(x) + a \langle \nabla \log p_t(x), \nabla f^{kl}(x) \rangle - \beta_t \frac{\sigma_t^2}{2} \Delta f^{kl}(x) + \frac{\partial \beta_t}{\partial t} f^{kl}(x) \nonumber \\
    &\quad + \left\langle \beta_t \nabla f^{kl}(x), v_t(x) - \sigma_t^2 \nabla \log q_t(x) - \frac{\sigma_t^2}{2} \beta_t \nabla f^{kl}(x) \right\rangle.
\end{align}
Taking $v_t(x) = -f_t(x) + \sigma_t^2 \nabla \log q_t(x)$ and $a = \beta_t \sigma_t^2 / 2$, we have:
\begin{align*}
    \frac{\partial \hat{p_t}^{kl}(x)}{\partial t} = -\left\langle \nabla, \hat{p_t}^{kl}(x) \left( -f_t(x) + \sigma_t^2 \nabla \log q_t(x) + \beta_t \frac{\sigma_t^2}{2} \nabla f^{kl}(x) \right) \right\rangle \\
    + \frac{\sigma_t^2}{2} \Delta \hat{p_t}^{kl}(x) + \hat{p_t}^{kl}(x) \big( g_t(x) - \mathbb{E}_{\hat{p_t}^{kl}(x)}[g_t(x)] \big),
\end{align*}
\begin{align}
    g_t(x) &= \left\langle \beta_t \nabla f^{kl}(x), \frac{\sigma_t^2}{2} \nabla \log \hat{p_t}^{kl}(x) \right\rangle + \frac{\partial \beta_t}{\partial t} f^{kl}(x) + \left\langle \beta_t \nabla f^{kl}(x), -f_t(x) - \frac{\sigma_t^2}{2} \beta_t \nabla f^{kl}(x) \right\rangle \nonumber \\
    &= \frac{\partial \beta_t}{\partial t} f^{kl}(x) + \left\langle \beta_t \nabla f^{kl}(x), \frac{\sigma_t^2}{2} \nabla \log q_t(x) - f_t(x) \right\rangle.
\end{align}
This can be simulated as:
\begin{equation}
    dx_t = \left( -f_t(x_t) + \sigma_t^2 \nabla \log q_t(x_t) + \beta_t \frac{\sigma_t^2}{2} \nabla f^{kl}(x_t) \right) dt + \sigma_t dW_t,
\end{equation}
\begin{equation}
    dw_t = \left[ \frac{\partial \beta_t}{\partial t} f^{kl}(x_t) + \left\langle \beta_t \nabla f^{kl}(x_t), \frac{\sigma_t^2}{2} \nabla \log q_t(x_t) - f_t(x_t) \right\rangle \right] dt.
\end{equation}

\paragraph{Step 2:}
Assuming $\beta_t = 1$, the above weighted SDE can be written as:
\begin{equation}
    dx_t = \left( -f_t(x_t) + \sigma_t^2 \nabla \log q_t(x_t) + \frac{\sigma_t^2}{2} \nabla f^{kl}(x_t) \right) dt + \sigma_t dW_t,
\end{equation}
\begin{equation}
    dw_t = \left[  \left\langle \nabla f^{kl}(x_t), \frac{\sigma_t^2}{2} \nabla \log q_t(x_t) - f_t(x_t) \right\rangle \right] dt.
\end{equation}

By assuming $b_t(x_t)= -f_t(x_t) + \frac{\sigma^2_t}{2} \nabla \log q_t(x_t)$, we can write the following as,
\begin{equation}
    dx_t = \left( b_t(x_t) + \frac{\sigma_t^2}{2} \nabla \log q_t(x_t) + \frac{\sigma_t^2}{2} \nabla f^{kl}(x_t) \right) dt + \sigma_t dW_t,
\end{equation}
\begin{equation}
    dw_t = \left[  \left\langle \nabla f^{kl}(x_t), b_t(x_t) \right\rangle \right] dt.
\end{equation}

Now, by following the standard SDE to ODE conversion rule, we can write as follows:
\begin{equation}
    dx_t = \left( b_t(x_t) + \frac{\sigma_t^2}{2} \nabla f^{kl}(x_t) \right) dt,
\end{equation}
\begin{equation}
    dw_t = \left[  \left\langle \nabla f^{kl}(x_t), b_t(x_t) \right\rangle \right] dt.
\end{equation}

Now, utilizing the results from Proposition 4.2 regarding  $- \nabla D_{KL}(q_t(x) || p_{boltz}(x))$ and utilizing the formulation of $f^{kl}(x_t)= -D_{KL}(q_t(x) || p_{boltz}(x))$, we get the final expression for the $i$-th particle as:
\begin{align}
    dx^i_t &= b_t(x^i_t)\>dt  + \frac{\sigma_t^2}{2}\frac{1}{n} \sum_{j=1}^n \left[ k(x^j_t, x^{i}_t) \nabla_{x^j_t} V(x^j_t) \> dt + \nabla_{x^j_t} k(x^j_t, x^{i}_t) \> dt \right]
\end{align}
\vspace{-15pt}
\begin{align}
    dw^i_t = \frac{1}{n} \sum_{j=1}^n \left[ \bm{\langle} k(x_t^j, x_t^{i}) \nabla_{x_t^j} V(x_t^j), b_t(x_t^i) \bm{\rangle} \> dt + \bm{\langle} \nabla_{x_t^j} k(x_t^j, x_t^{i}), b_t(x_t^i) \bm{\rangle}\> dt \right] 
\end{align}
This completes the Proof.

\end{proof}

\section{Performance Comparison in Longer Horizon Search Tasks}\label{sec:se-quant}
\begin{figure*}[h]
\centering

\newcommand{\legenditem}[4]{%
  \begin{tikzpicture}[baseline=-0.6ex]
    \draw[#1, line width=1.8pt, #2] (0,0) -- (0.9,0);
    \node[text=#1, fill=white, inner sep=0.6pt] at (0.45,0) {\scriptsize $#3$};
  \end{tikzpicture}\,#4%
}

\newcommand{\methodlegend}{%
  {\footnotesize
  \setlength{\tabcolsep}{4pt}%
  \begin{tabular}{@{}ccccc@{}}
    \legenditem{cyan!70!black}{solid}{\circ}{DAS} &
    \legenditem{red!75!black}{dashed}{\square}{FKS} &
    \legenditem{orange!90!black}{dash dot}{\triangle}{BoN} &
    \legenditem{purple!80!black}{dotted}{\diamond}{MFM} &
    \legenditem{green!60!black}{solid}{\blacktriangledown}{\textbf{IMPFM (Our)}}
  \end{tabular}%
  }%
}

\newcommand{\barpanel}[2]{%
  \begin{minipage}[t]{0.31\textwidth}
    \centering
    \includegraphics[width=\linewidth]{#1}\par
    \vspace{-0.3ex}
    {\scriptsize #2}
  \end{minipage}%
}

\methodlegend
\vspace{0.1ex}
\noindent\rule{\textwidth}{0.5pt}
\vspace{0.3ex}
{\begin{minipage}[t]{0.24\textwidth}
  \centering
  \includegraphics[width=\linewidth]{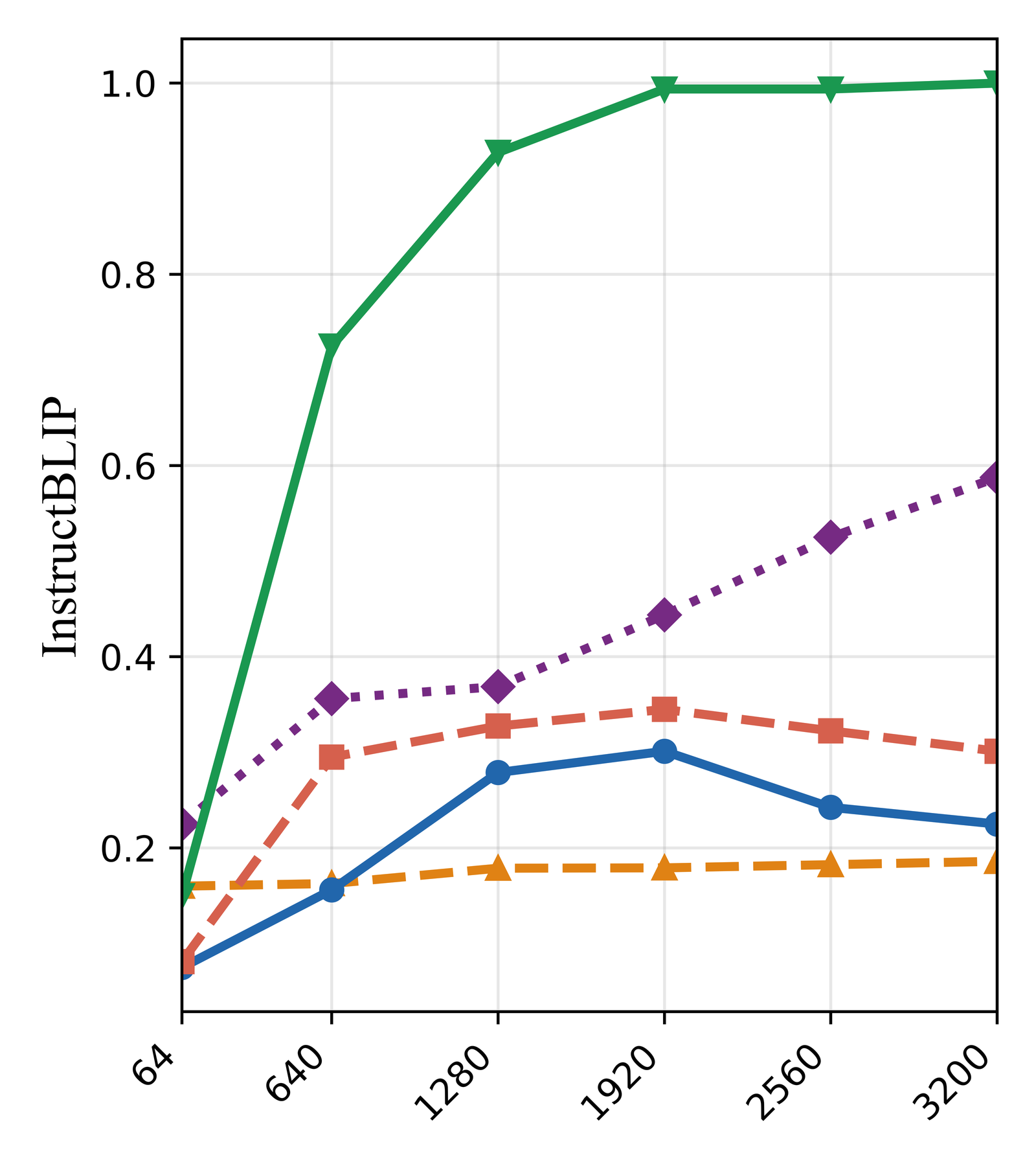}\par
  \vspace{-0.3ex}
  {\scriptsize Feedback Budgets}
\end{minipage}}\hspace{0.035\textwidth}%
{\begin{minipage}[t]{0.24\textwidth}
  \centering
  \includegraphics[width=\linewidth]{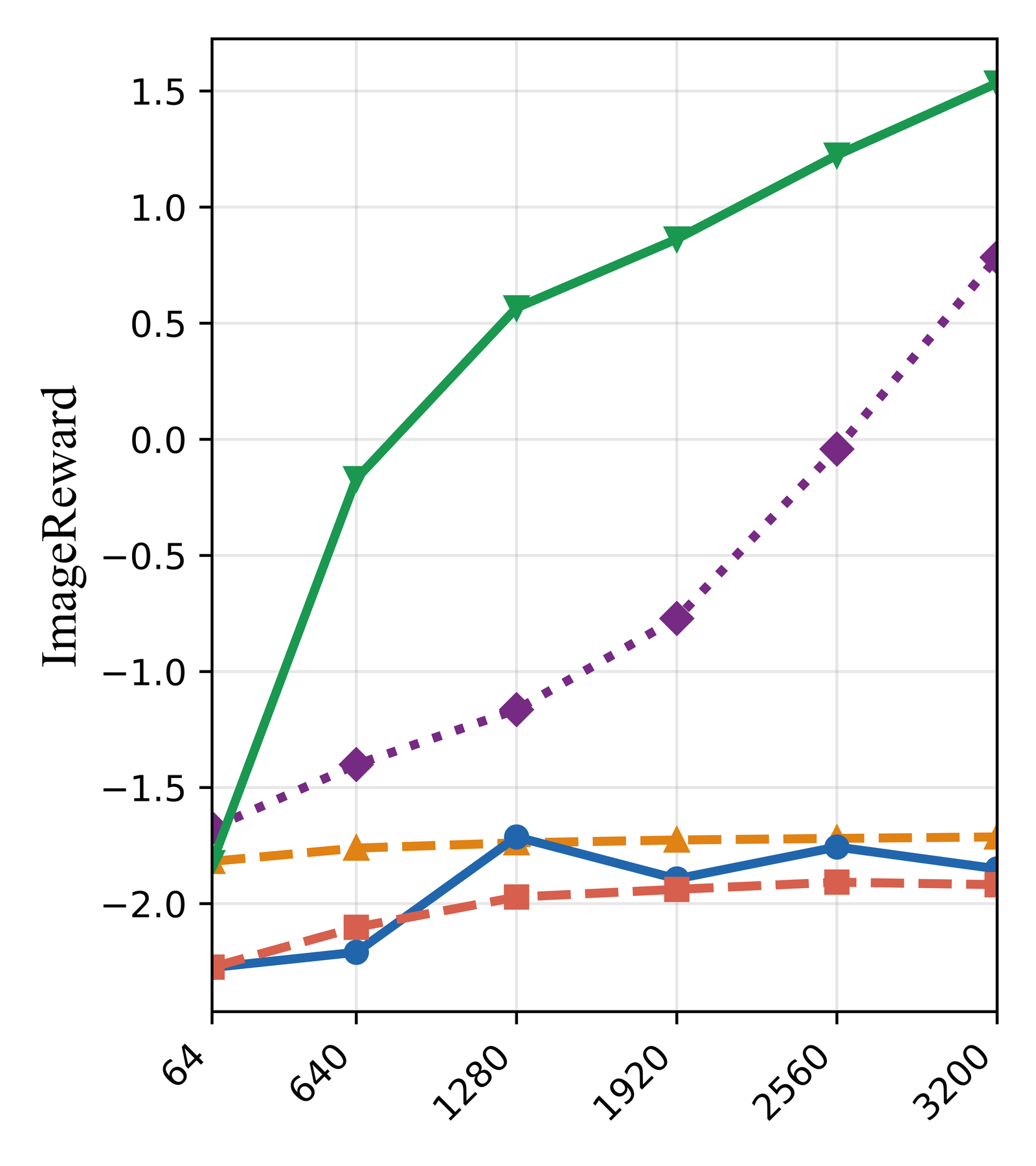}\par
  \vspace{-0.3ex}
  {\scriptsize Feedback Budgets}
\end{minipage}}\hspace{0.035\textwidth}%
{\begin{minipage}[t]{0.24\textwidth}
  \centering
  \includegraphics[width=\linewidth]{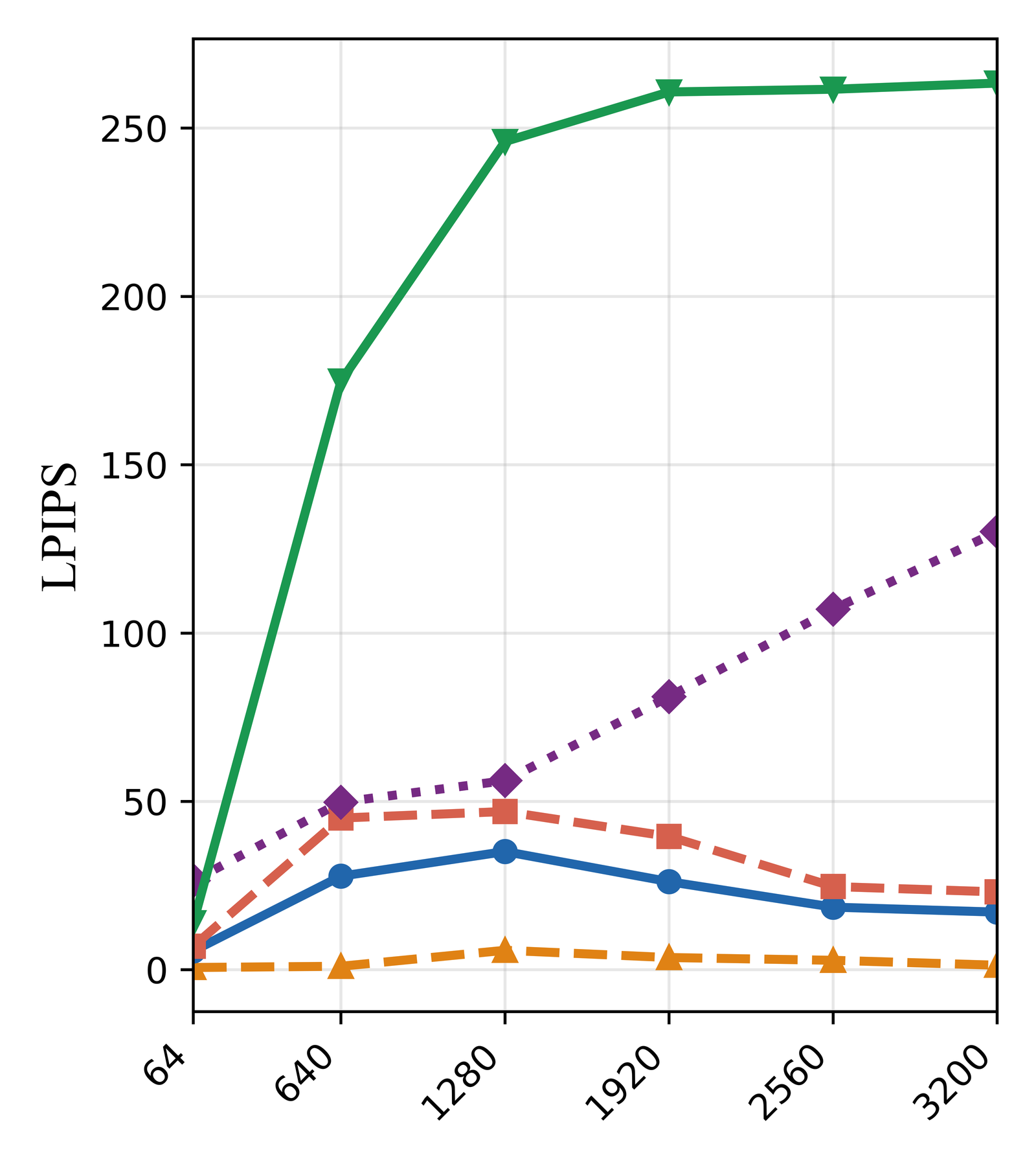}\par
  \vspace{-0.3ex}
  {\scriptsize Feedback Budgets}
\end{minipage}}
\vspace{-3pt}
\caption{\small{Search Performance Analysis of Competitive Approaches on Imagenet Target Classes.}}
\label{fig:search-result-longer}
\end{figure*}

While in the main paper, we evaluate IMPFM on feedback-efficient search tasks with relatively short horizons, this section extends our comparative analysis to longer-horizon tasks, with findings reported in Figure~\ref{fig:search-result-longer}. Notably, the performance trends in these extended settings mirror those observed in the shorter-horizon tasks. These additional results reinforce that IMPFM is highly sample-efficient—achieving higher rewards with fewer interactions—while simultaneously maintaining a robust search capability over longer horizons that consistently outperforms the baselines.

\section{Additional Analysis on the Impact of the Interaction-Aware Feynman-Kac Corrector}\label{sec:int-fkc}
\begin{figure*}[h]
\centering
\newcommand{\figprompt}[1]{%
  \begin{minipage}[t]{0.27\textwidth}
    \centering\itshape ``#1''\par
  \end{minipage}%
}
\newcommand{\figsample}[1]{%
  \begin{minipage}[t]{0.27\textwidth}
    \centering
    \includegraphics[width=\linewidth]{#1}
  \end{minipage}%
}
\noindent\rule{\textwidth}{0.6pt}
\vspace{0.4ex}
\begin{center}
\setlength{\tabcolsep}{4pt}%
\renewcommand{\arraystretch}{1.12}%
\begin{tabular}{@{}c@{\hspace{6pt}}c@{\hspace{8pt}}c@{\hspace{8pt}}c@{}}
  & \figprompt{\textcolor{purple}{A Red Sports Car.}}
  & \figprompt{\textcolor{purple}{A Black Cat.}}
  & \figprompt{\textcolor{purple}{A green bell pepper.}} \\[0.6ex]
  \raisebox{10.1ex}{\textbf{\scriptsize FKC (Reward Tilted)}}
  & \figsample{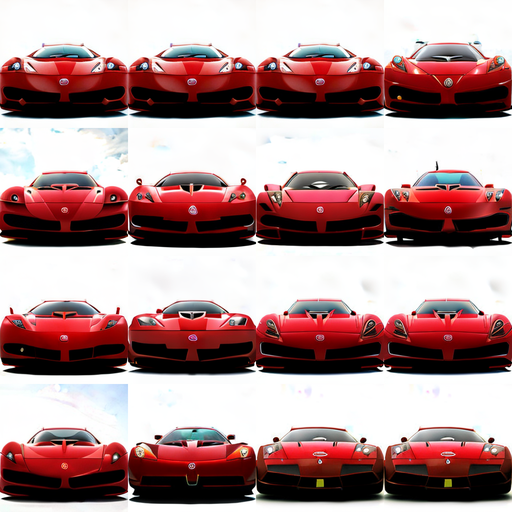}
  & \figsample{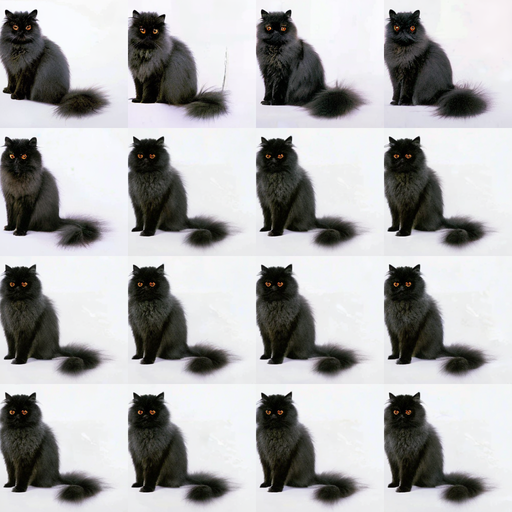}
  & \figsample{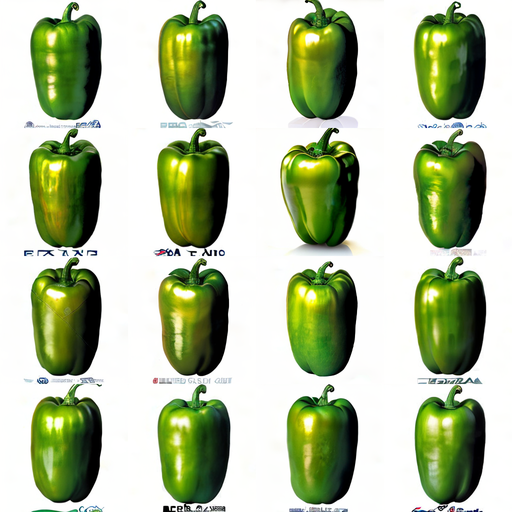} \\[0.6ex]
  \multicolumn{4}{@{}c@{}}{\rule{\textwidth}{0.5pt}} \\[0.6ex]
  \raisebox{10.1ex}{\textbf{\scriptsize IMPFM}}
  & \figsample{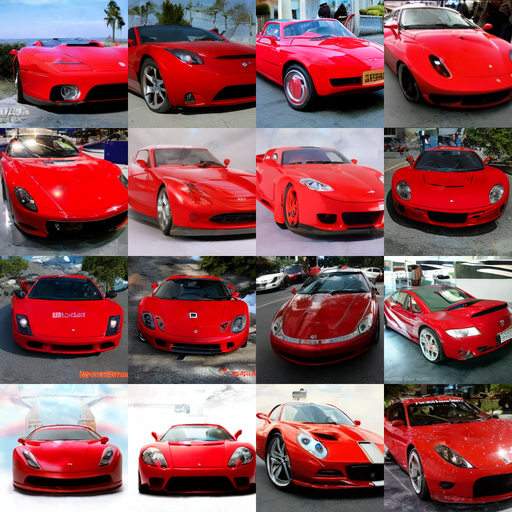}
  & \figsample{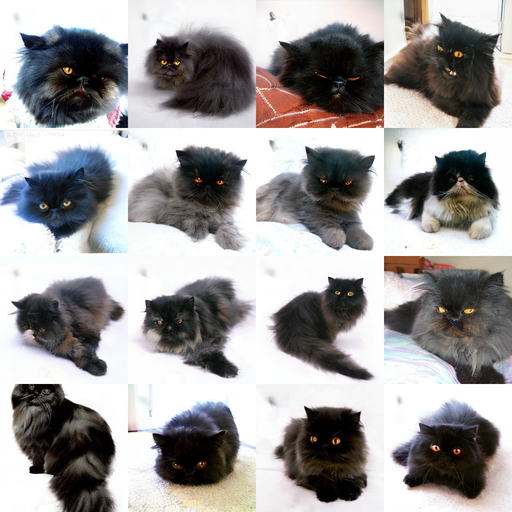}
  & \figsample{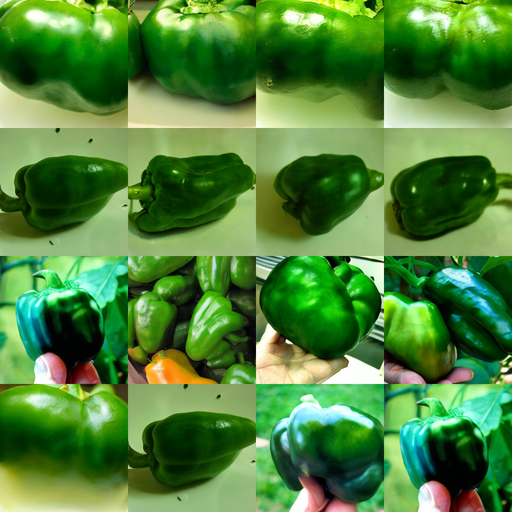}
\end{tabular}
\end{center}
\vspace{0.6ex}
\noindent\rule{\textwidth}{0.6pt}
\caption{\small{Impact of Particle Interaction in Mitigating Reward Over-optimization on FKC.}}
\label{fig:fkc-comp}
\end{figure*}

\begin{figure*}[h]
\centering
\newcommand{\figpromptoc}[1]{%
  \begin{minipage}[t]{0.27\textwidth}
    \centering\itshape ``#1''\par
  \end{minipage}%
}
\newcommand{\figsampleoc}[1]{%
  \begin{minipage}[t]{0.27\textwidth}
    \centering
    \includegraphics[width=\linewidth]{#1}
  \end{minipage}%
}
\noindent\rule{\textwidth}{0.6pt}
\begin{center}
\setlength{\tabcolsep}{4pt}%
\renewcommand{\arraystretch}{1.12}%
\begin{tabular}{@{}c@{\hspace{6pt}}c@{\hspace{8pt}}c@{\hspace{8pt}}c@{}}
  & \figpromptoc{\textcolor{purple}{A green bell pepper.}}
  & \figpromptoc{\textcolor{purple}{A Red Sports Car.}}
  & \figpromptoc{\textcolor{purple}{A dog in the snow.}} \\[0.6ex]
  \raisebox{10.1ex}{\textbf{\scriptsize MFM}}
  & \figsampleoc{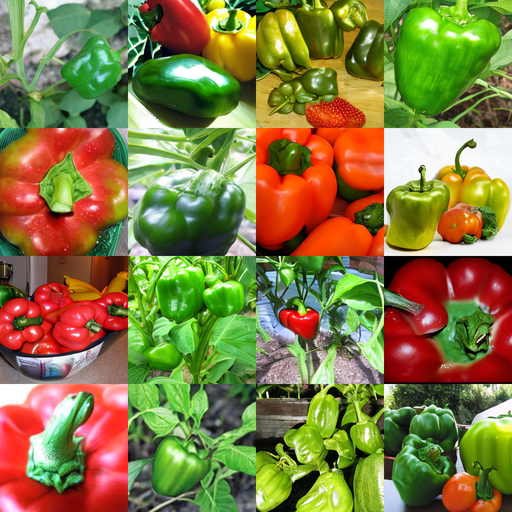}
  & \figsampleoc{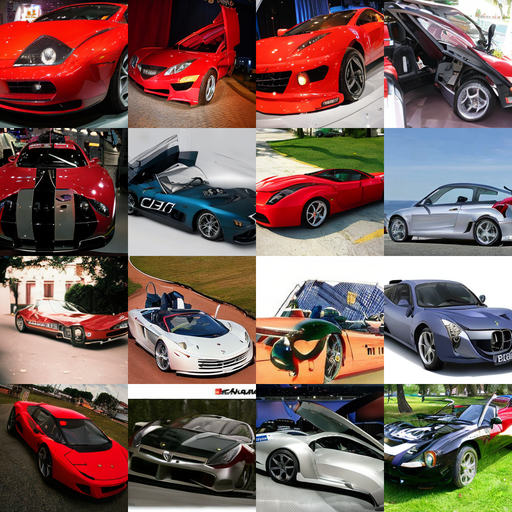}
  & \figsampleoc{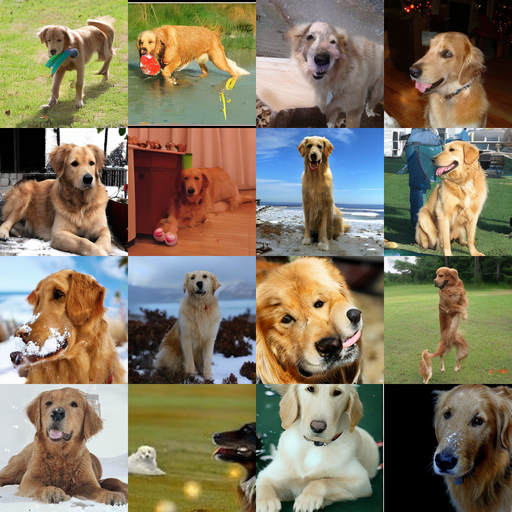} \\[0.02ex]
  \multicolumn{4}{@{}c@{}}{\rule{\textwidth}{0.5pt}} \\[0.6ex]
  \raisebox{10.1ex}{\textbf{\scriptsize MPC}}
  & \figsampleoc{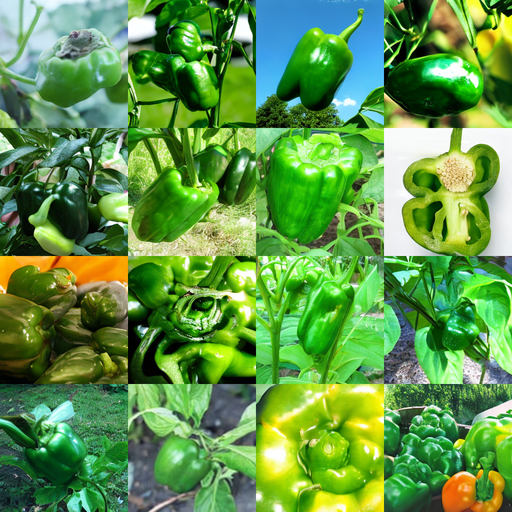}
  & \figsampleoc{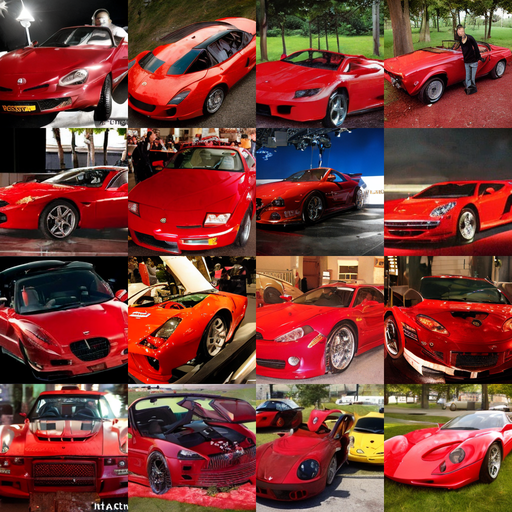}
  & \figsampleoc{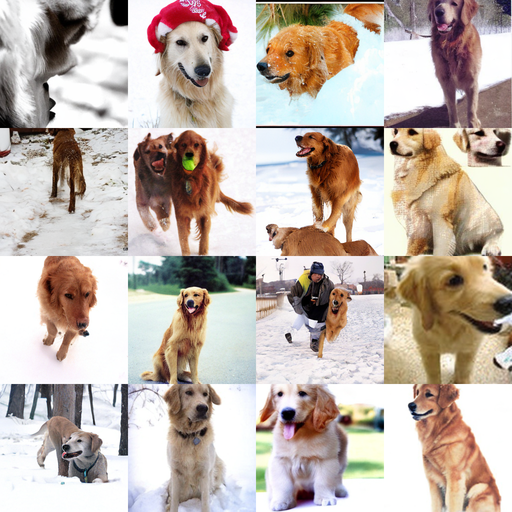}
\end{tabular}
\end{center}
\vspace{0.02ex}
\noindent\rule{\textwidth}{0.6pt}
\caption{\small{Impact of Particle Interaction in Mitigating Reward Over-optimization on Optimal Control.}}
\label{fig:oc-mpc}
\end{figure*}

Building on the qualitative and quantitative analysis presented in the main paper, we provide here additional results to further substantiate the impact of the interaction-aware Feynman-Kac corrector. Following the same ablation protocol, we remove the interaction terms — comprising both the repulsive term and the kernel-weighted value gradient — from the drift correction and reweighting mechanisms, yielding a standard reward-tilted Feynman-Kac corrector, referred to as FKC-Reward-Tilted. We benchmark this against our full Interaction-Aware Feynman-Kac corrector and report the results in Fig.~\ref{fig:fkc-comp}.
As illustrated, the interaction-aware formulation substantially enhances sample diversity, directly alleviating mode collapse. The key limitation of the FKC-Reward-Tilted sampler lies in its purely pointwise optimization objective: by neglecting global distributional constraints, it greedily pursues high-reward regions and consequently becomes trapped in local neighborhoods, rendering it susceptible to reward over-optimization. In contrast, the interaction-aware Feynman-Kac corrector incorporates a KL-divergence tilting factor that enforces distributional alignment, while the explicit inter-particle interaction mechanism actively promotes global coverage by aggregating information across the entire particle ensemble. This dual mechanism naturally steers particles away from local optima, fostering both diversity and robustness against reward over-optimization, as clearly reflected in Fig.~\ref{fig:fkc-comp}. Furthermore, we analyze the specific impact of the interaction term within the drift correction mechanism (Eq.~\ref{eq:dynamics}). To isolate this effect, we visually compare the samples generated by our fully interactive formulation (MPC) against an ablated, non-interacting baseline (MFM). As shown in Figure~\ref{fig:oc-mpc}, incorporating particle interaction significantly improves both sample diversity and target alignment. This highlights the interaction term as a critical component for overcoming the severe mode collapse and reward over-optimization typical of standard optimal control frameworks.
In a nutshell, these additional qualitative analyses reinforce the importance of the interaction-aware Feynman-Kac corrector for feedback-efficient search.
\vspace{4pt}
\vspace{-22pt}
\section{Visualizing the Impact of Sufficient Statistic}\label{sec:ss-a}
\begin{figure*}[h]
\centering
\newcommand{\imagewithessb}[2]{%
  \begin{minipage}[t]{0.185\linewidth}
    \centering
    \includegraphics[width=\linewidth]{#1}\par
    \vspace{-0.2ex}
    {\scriptsize Feedback Steps = #2}
  \end{minipage}%
}
\newcommand{\methodrowb}[6]{%
  \begin{minipage}[c][0.205\textwidth][c]{0.12\textwidth}
    \centering
    \raisebox{2.0ex}[0pt][0pt]{\scriptsize\shortstack[c]{#1}}
  \end{minipage}%
  \hspace{0.01\textwidth}%
  \begin{minipage}[t]{0.86\textwidth}
    \centering
    \imagewithessb{#2}{20}\hspace{0.008\linewidth}%
    \imagewithessb{#3}{40}\hspace{0.008\linewidth}%
    \imagewithessb{#4}{60}\hspace{0.008\linewidth}%
    \imagewithessb{#5}{80}\hspace{0.008\linewidth}%
    \imagewithessb{#6}{100}
  \end{minipage}%
}
{\itshape ``\textcolor{purple}{A red Sports Car.
}''\par}
\vspace{0.02ex}
\noindent\rule{\textwidth}{0.6pt}
\vspace{0.03ex}
\methodrowb{SDE (IMPFM)}{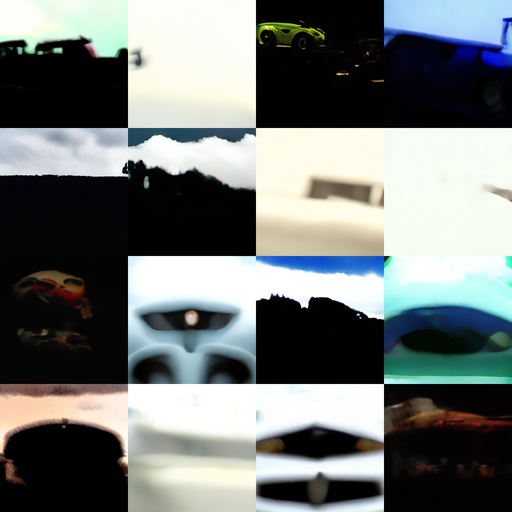}{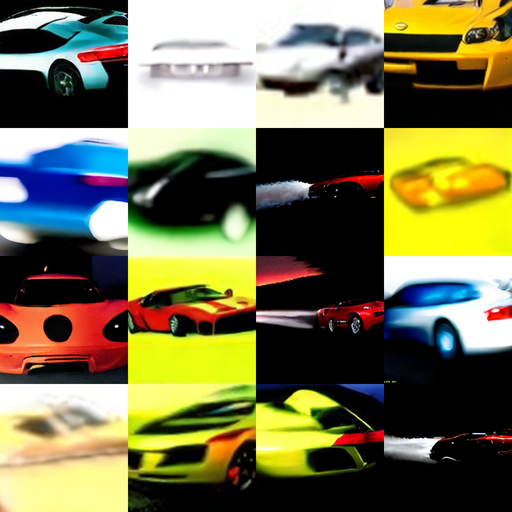}{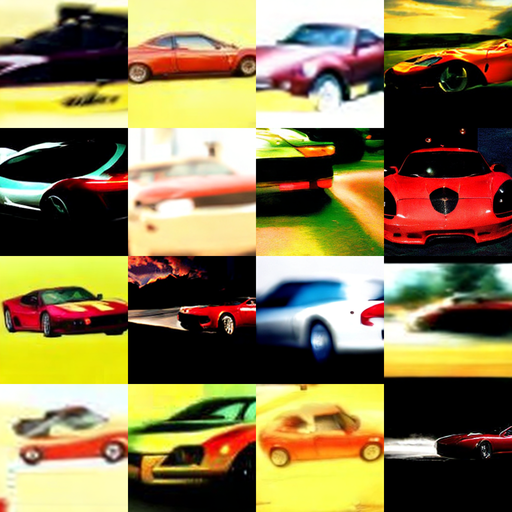}{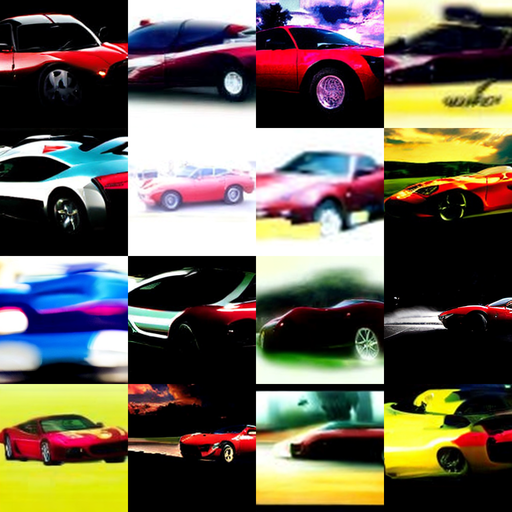}{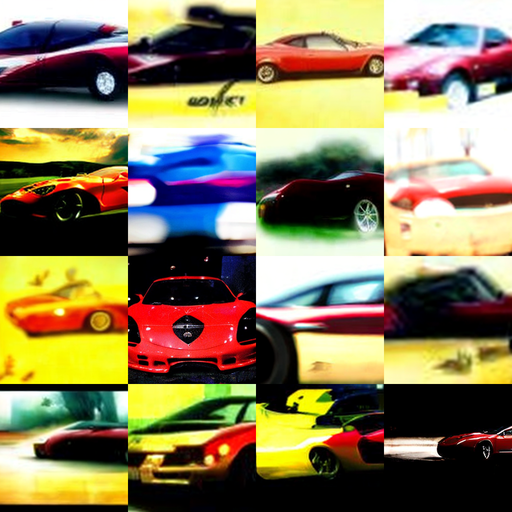}
\vspace{0.02ex}
\methodrowb{SS (IMPFM)}{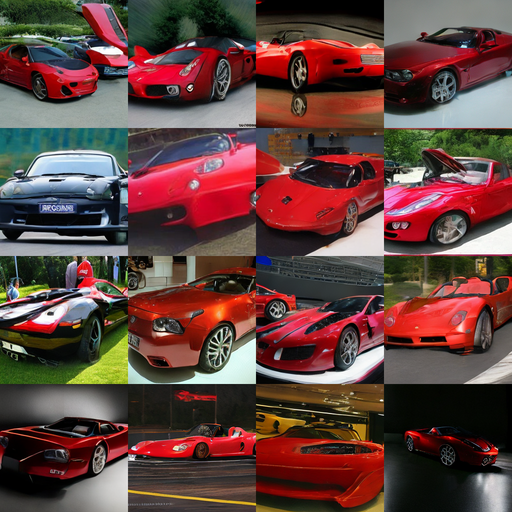}{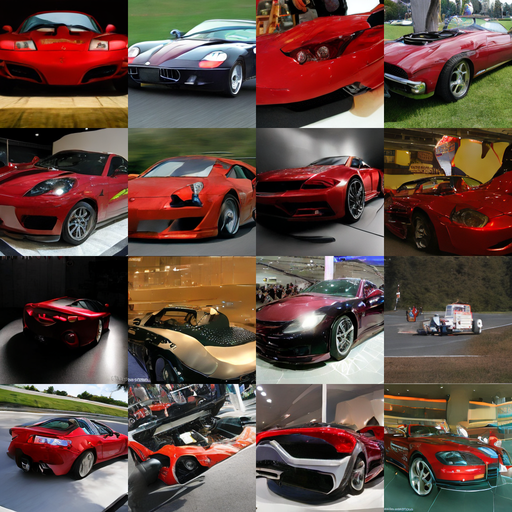}{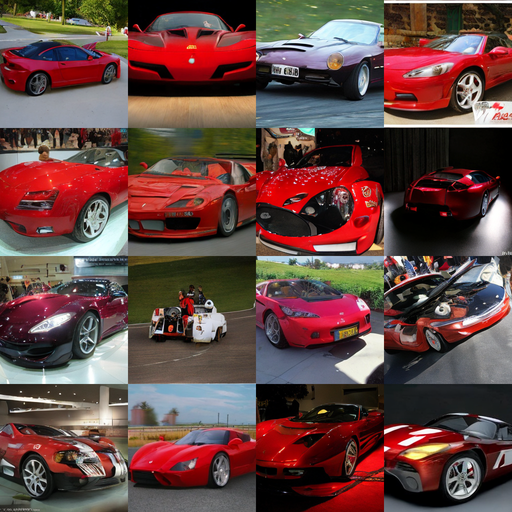}{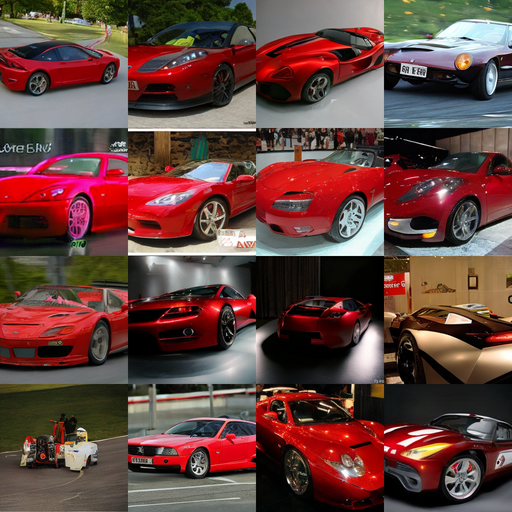}{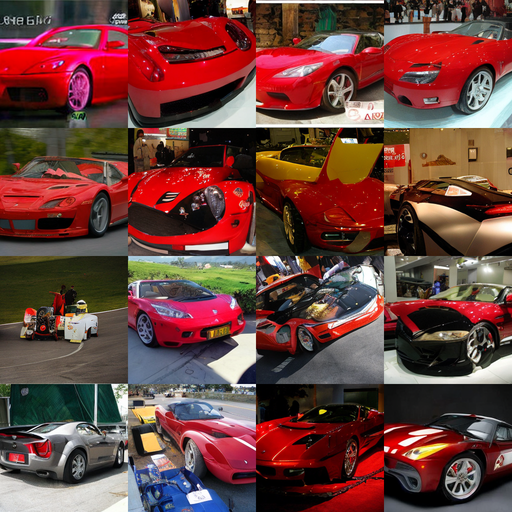}
\vspace{-0.1ex}
\noindent\rule{\textwidth}{0.1pt}
\caption{\small{Visualizing the Impact of Sufficient Statistics for Lower Feedback Budget. }}
\label{fig:ss-additional}
\end{figure*}

\vspace{-12pt}
In the main paper (Section~\ref{sec:ss}), we analyze the impact of the Sufficient Statistic (SS). Here, we present additional visualizations highlighting that the SS is crucial for enabling feedback-efficient search, especially under severe budget constraints. As depicted in Fig.~\ref{fig:ss-additional}, when the feedback step is less than 50, the alternative approach fails to generate high-quality samples. In contrast, \emph{by leveraging dynamic transition steps enabled via the SS mechanism, IMPFM successfully generates high-quality, target-aligned samples even with a very small feedback step (e.g., 20)}.

\section{Additional Qualitative Analysis: Taming Weight Degeneracy Through Particle Interaction-Guided Drift Correction}\label{sec:wd-a}
\vspace{-3pt}
\begin{figure}[h]
    \centering
    \includegraphics[width=0.6\linewidth]{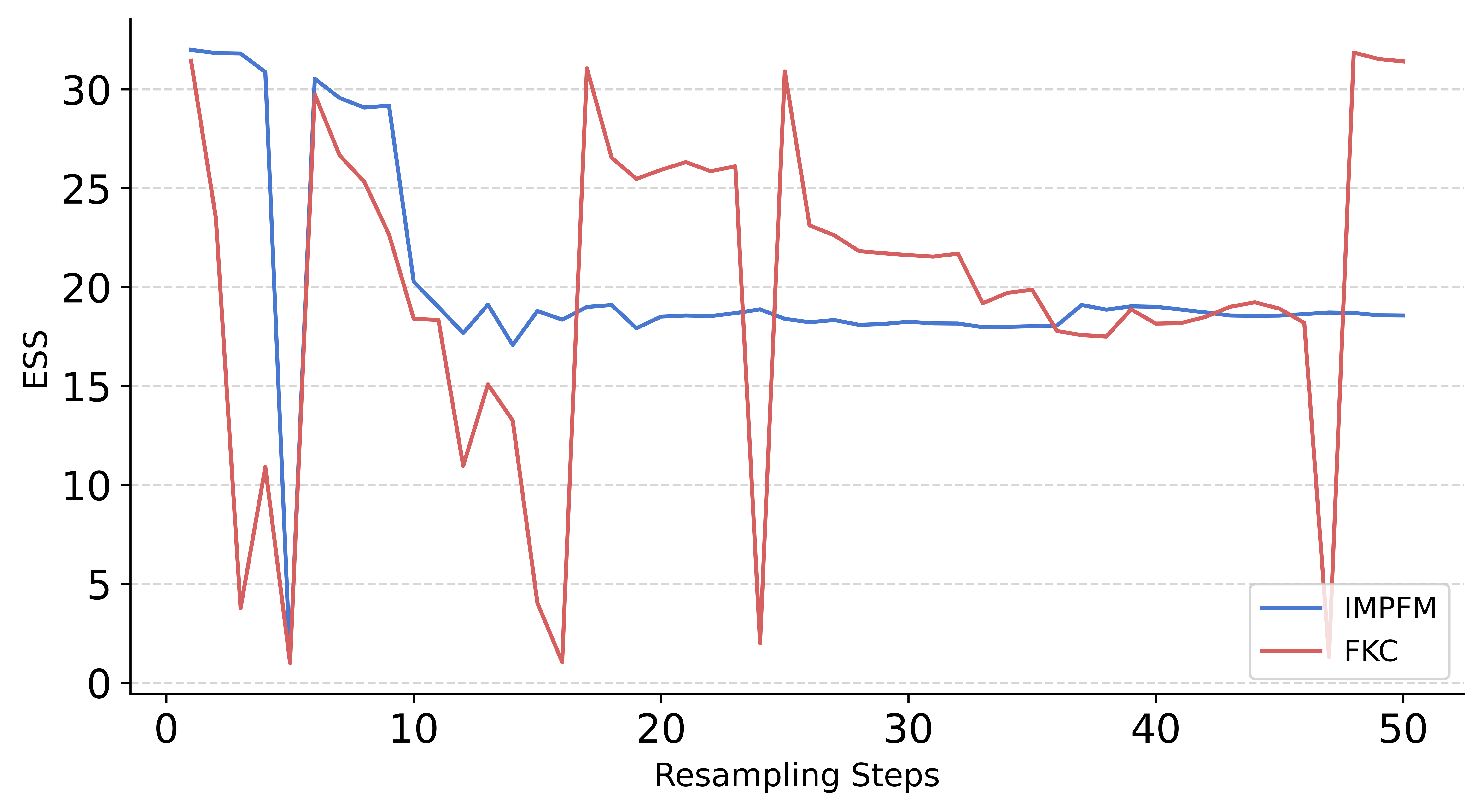}
    \caption{Efficacy of Interactive Particle mechanism on Mitigating Weight Degeneracy on FKC Sampler.}
    \label{fig:ess__}
\end{figure}
\begin{figure*}[h]
\centering
\newcommand{\imagewithess}[2]{%
  \begin{minipage}[t]{0.235\linewidth}
    \centering
    \includegraphics[width=\linewidth]{#1}\par
    \vspace{-0.2ex}
    {\scriptsize Resampling Step: #2}
  \end{minipage}%
}
{\itshape ``\textcolor{purple}{a red sports car.}''\par}
\vspace{0.4ex}
\noindent\rule{\textwidth}{0.6pt}
\vspace{0.5ex}
\begin{minipage}[c][0.205\textwidth][c]{0.12\textwidth}
  \centering
  \raisebox{2.0ex}[0pt][0pt]{\scriptsize\shortstack[c]{Value based\\Isolated Particle\\Drift Correction}}
\end{minipage}%
\hspace{0.01\textwidth}%
\begin{minipage}[t]{0.87\textwidth}
  \centering
  \imagewithess{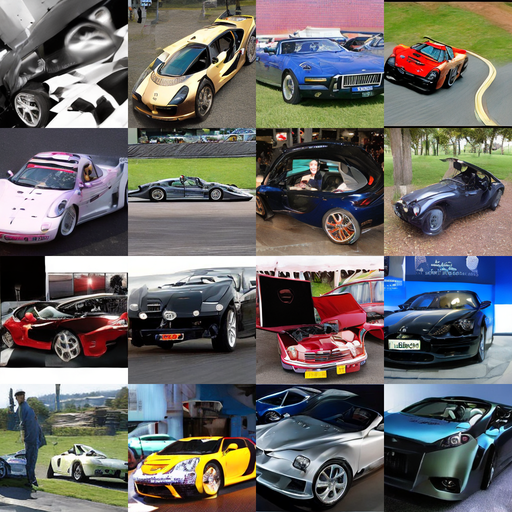}{0}\hspace{0.012\linewidth}%
  \imagewithess{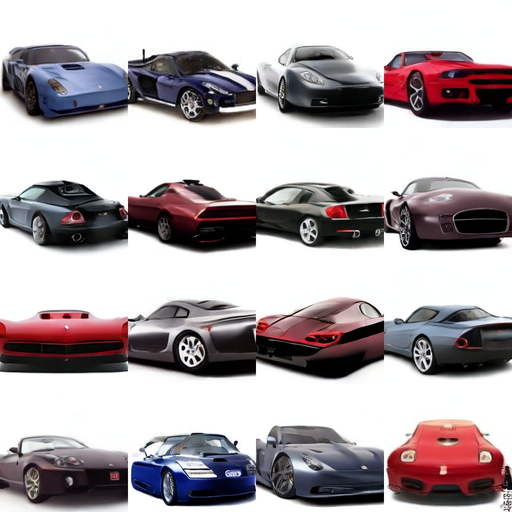}{10}\hspace{0.012\linewidth}%
  \imagewithess{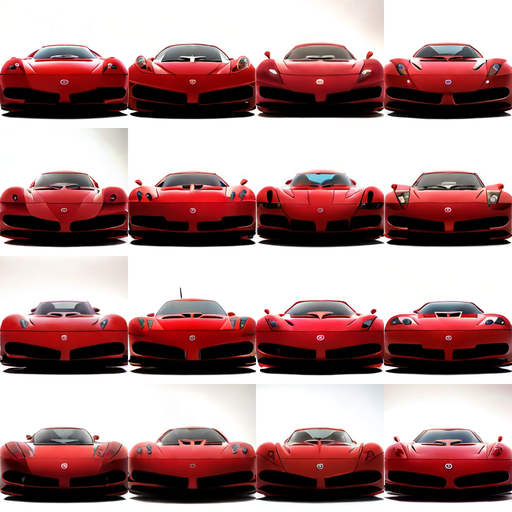}{30}\hspace{0.012\linewidth}%
  \imagewithess{figs/fkc10.png}{50}
\end{minipage}%
\vspace{0.8ex}
\begin{minipage}[c][0.205\textwidth][c]{0.12\textwidth}
  \centering
  \raisebox{2.0ex}[0pt][0pt]{\scriptsize\shortstack[c]{Interactive\\Multi-Particle\\Drift correction}}
\end{minipage}%
\hspace{0.01\textwidth}%
\begin{minipage}[t]{0.87\textwidth}
  \centering
  \imagewithess{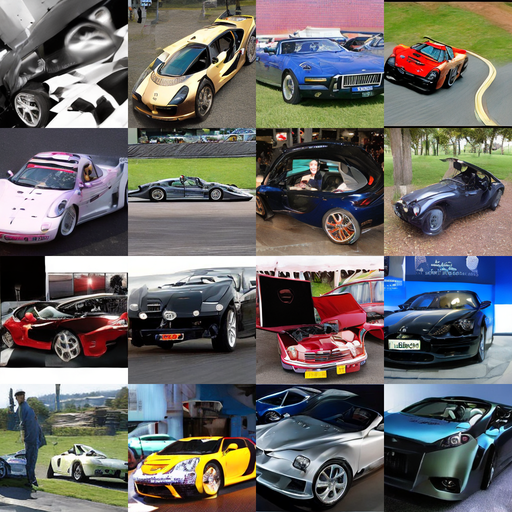}{0}\hspace{0.012\linewidth}%
  \imagewithess{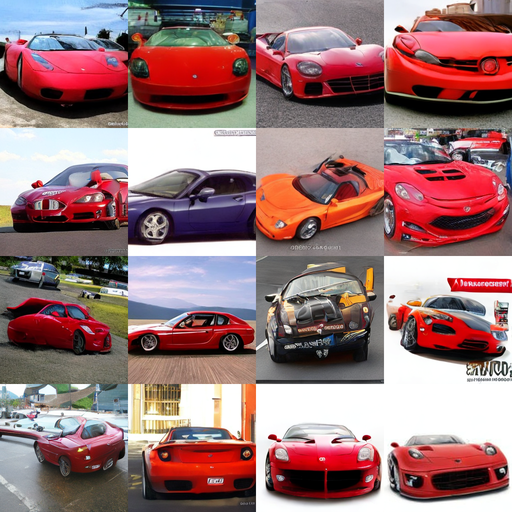}{10}\hspace{0.012\linewidth}%
  \imagewithess{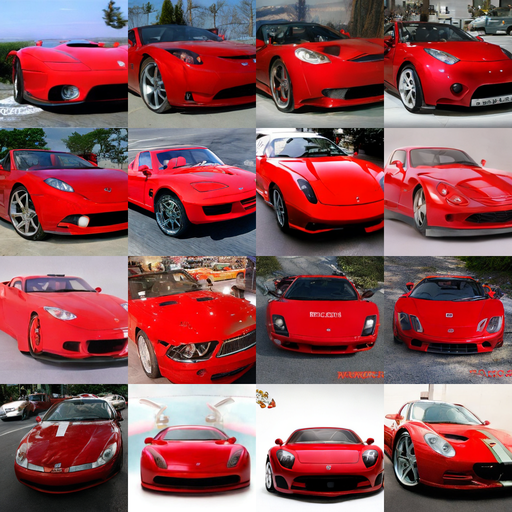}{30}\hspace{0.012\linewidth}%
  \imagewithess{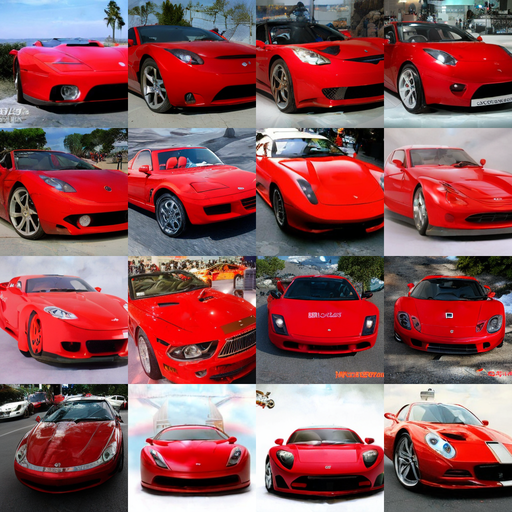}{50}
\end{minipage}%
\vspace{0.6ex}
\noindent\rule{\textwidth}{0.6pt}
\caption{\small{Impact of Interactive Multi-Particle Drift Correction Mechanism on Mitigating Weight Degeneracy. }}
\label{fig:search-visu_ESS}
\end{figure*}
In this section, we present a deeper qualitative investigation into how the proposed multi-particle interaction-driven drift correction mechanism combats weight degeneracy.  To achieve this, we visualize the denoised samples generated via flow-map for each particle in the batch and compute the Effective Sample Size (ESS) at every single transition step. By contrasting our fully interacting dynamics (Eq.~\ref{eq:dynamics}) against a naive baseline—where particles are treated in strict isolation and corrected purely based on individual value—we directly highlight the stabilizing power of our approach. We report the findings in Figure~\ref{fig:ess__} and~\ref{fig:search-visu_ESS}. 
Crucially, we demonstrate that multi-particle interaction acts as a fundamental driver of sample diversity. Unlike isolated particle dynamics—which rapidly collapse into semantically redundant outputs—our interacting system sustains rich, diverse generation across every correction step. This structural stability is corroborated by a drastic, consistent improvement in the Effective Sample Size (ESS), effectively eradicating the pervasive weight degeneracy problem. Importantly, this interactive framework is made computationally viable by flow-map; by enabling highly efficient evaluation of value gradients ($\nabla V_t$), flow-maps unlock the ability to compute complex, multi-particle interactions without prohibitive overhead.

\section{Details of Class and Prompt Detail in the Search Experiment}\label{sec:se-class}
In the following table~\ref{tb:st}, we detail the randomly selected ImageNet target classes and their fine-grained specifications used to evaluate the competing methods on the feedback-efficient search task.
\begin{table}[h]
    \centering
    \begin{tabular}{|p{0.22\textwidth}|p{0.68\textwidth}|}
        \hline
        \textbf{Target Class} & \textbf{Fine-Grained Prompt} \\
        \hline
        Car & A Red Sports Car \\
        \hline
        Cat & A Black Cat. \\
        \hline
        Dog & A Dog in the Snow. \\
        \hline
        Daisy & A Red Daisy. \\
        \hline
        Pepper & A Green Bell Pepper. \\
        \hline
    \end{tabular}
    \vspace{2pt}
    \caption{Target classes and their corresponding fine-grained prompts for the Feedback-efficient Search task.}
    \label{tb:st}
\end{table}

\section{Additional Comparative Visualization on Search Task}\label{sec:se-vis-a}

\begin{figure*}[h]
\centering
\newcommand{\stacktriplepanel}[3]{%
  \setlength{\tabcolsep}{2pt}%
  \begin{tabular}{@{}c@{\hspace{10pt}}c@{\hspace{10pt}}c@{}}
    \includegraphics[width=0.28\linewidth]{#1} &
    \includegraphics[width=0.28\linewidth]{#2} &
    \includegraphics[width=0.28\linewidth]{#3}
  \end{tabular}%
}
\newcommand{\stackpanelblock}[4]{%
  \begin{minipage}[t]{\textwidth}
    \centering
    \stacktriplepanel{#2}{#3}{#4}
    \par\vspace{0.4ex}
    {\itshape ``#1''\par}
  \end{minipage}%
}
\newcommand{\stacktopheaders}{%
  {\fontsize{9.5}{10.5}\selectfont
  \begin{tabular}{@{}c@{\hspace{10pt}}c@{\hspace{10pt}}c@{}}
    \makebox[0.28\linewidth][c]{\textbf{FKS}} &
    \makebox[0.28\linewidth][c]{\textbf{MFM}} &
    \makebox[0.28\linewidth][c]{\textbf{IMPFM}}
  \end{tabular}%
  }
}
\stacktopheaders
\vspace{-1.0ex}

\noindent\rule{\textwidth}{0.6pt}
\vspace{-1.0ex}
\stackpanelblock{\textcolor{purple}{a red sports car.}}{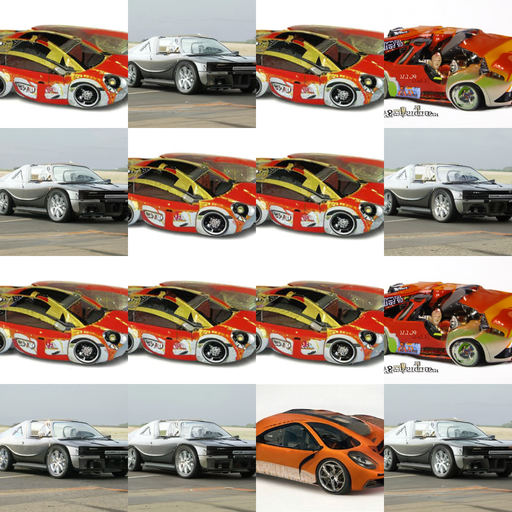}{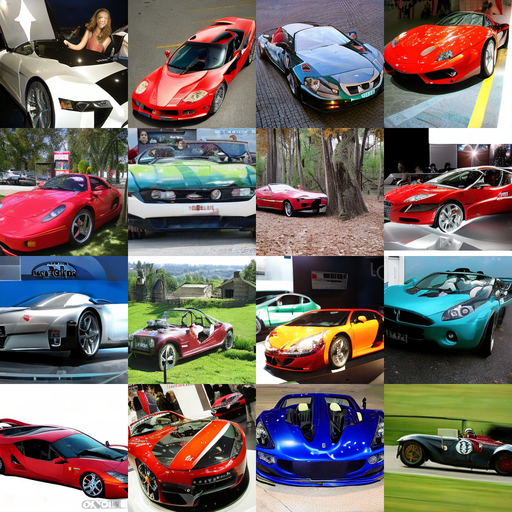}{figs/car_impfm.png}

\vspace{1.2ex}
\stacktopheaders
\vspace{-1.0ex}

\noindent\rule{\textwidth}{0.6pt}
\vspace{-1.0ex}
\stackpanelblock{\textcolor{purple}{a black cat.}}{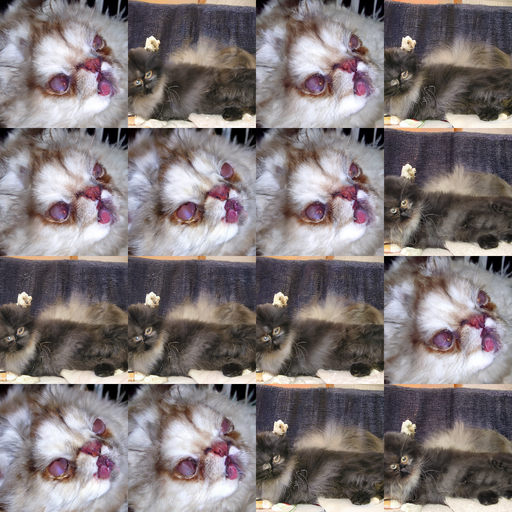}{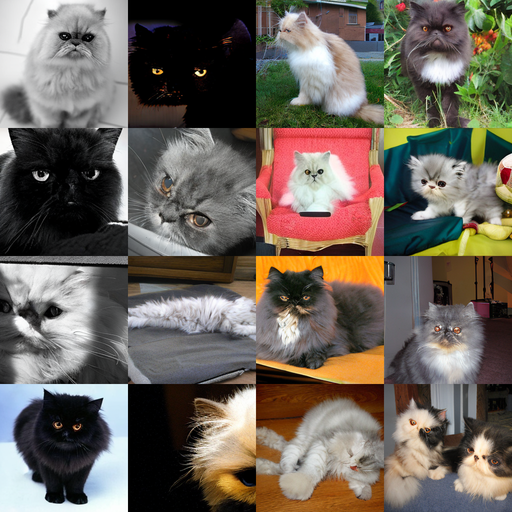}{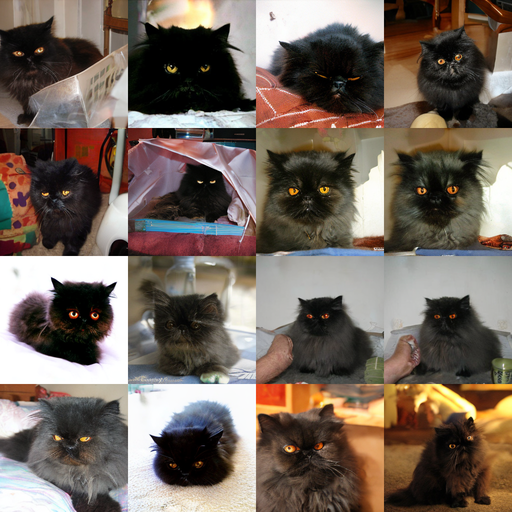}
\vspace{0.6ex}
\caption{\small{Search Visualizations.}}
\label{fig:search-visu11}
\vspace{-10pt}
\end{figure*}

In Fig.~\ref{fig:search-visu11}, we present additional comparative visualizations for search tasks involving diverse classes from ImageNet. These figures further highlight the impact of IMPFM in enabling feedback-efficient search.

\section{Details about the Evaluation Dataset for Alignment Tasks}\label{sec:align-eval-data}
We provide the evaluation prompts for both the compositional and quantity-aware alignment tasks in separate Excel files, available via the anonymous link.


\section{Implementation Details}\label{sec:imp-a}
For all experiments in the feedback-driven search setting, we use 32 particles, with each particle contributing one posterior sample for feedback guidance, resulting in 32 feedback samples per resampling step. Notably, when configured with a single particle, IMPFM reduces to the MFM baseline. However, as the particle count increases, IMPFM outperforms MFM under the same feedback budget due to its posterior sample sharing mechanism. Other implementation details are available in the provided code. Finally, all experiments are
implemented in PyTorch and conducted on 3 NVIDIA A100 GPUs. Our training and inference code will be made public.

\section{Derivation of TrigFlow/SANA~\citep{chen2025sana} Noise Schedule to Adapt IMPFM on Alignment Tasks}\label{sec:sana-sch-a}

In this section, we describe how we adapt the trigonometric noise schedule from TrigFlow~\citep{chen2025sana} and the consistency model (SANA-Sprint)~\citep{chen2025sana} to our IMPFM framework, which operates under a different noise schedule. We 
first review the TrigFlow parameterization, then describe the consistency model 
reparameterization, and finally derive the adaptation of the noise schedule to our setting.
 
\subsection{Background: TrigFlow Transformation}
\label{subsec:SANA-sch}
 
TrigFlow introduces a trigonometric reparameterization of the Flow Matching 
timestep, establishing a bijective mapping between the standard FM timestep 
$t_{FM} \in [0, 1]$ and a cosine-based schedule parameterized by $t_{cm}$. 
Specifically, the mapping is defined as:
\begin{equation}
    t_{FM} = \frac{\sin(t_{cm})}{\sin(t_{cm}) + \cos(t_{cm})},
    \label{eq:tFM}
\end{equation}
where $t_{cm}$ controls the balance between noise and signal through 
trigonometric interpolation. This mapping ensures that $t_{FM} = 0$ 
corresponds to a clean sample and $t_{FM} = 1$ corresponds to pure noise, 
while providing a smoother, geometry-aware interpolation path compared to 
a linear schedule.
 
Given the mapping in Eq.~\eqref{eq:tFM}, the noisy latent $X_{t_{FM}}$ is 
expressed in terms of the original latent $X_{t_{cm}}$ as:
\begin{equation}
    X_{t_{FM}} = \frac{X_{t_{cm}}}{\sigma_d} \sqrt{t_{FM}^2 + (1 - t_{FM})^2},
    \label{eq:Xtfm}
\end{equation}
where $\sigma_d$ is a scaling factor that normalizes the latent representation. 
The factor $\sqrt{t_{FM}^2 + (1-t_{FM})^2}$ accounts for the norm of the trigonometric interpolation coefficients, ensuring consistent scaling 
across timesteps.
 
Under this reparameterization, the score network $\hat{F}_\theta$ is 
reformulated to operate on the rescaled latent $X_{t_{cm}} / \sigma_d$ at timestep $t_{cm}$. The relationship between the TrigFlow score network and the velocity field $V_\theta$ of the standard Flow Matching model is:
\begin{equation}
    \begin{split}
        \hat{F}_\theta\!\left(\frac{X_{t_{cm}}}{\sigma_d},\, t_{cm},\, y\right) 
        &= \frac{1}{\sqrt{t_{FM}^2 + (1 - t_{FM})^2}} \\
        &\quad \times \Big[ (1 - 2t_{FM})\, X_{t,FM} + \left(1 - 2t_{FM} + 2t_{FM}^2\right) V_\theta\!\left(X_{t_{FM}},\, t_{FM},\, y\right) \Big],
    \end{split}
    \label{eq:Ftheta}
\end{equation}
where $y$ is the conditioning input (e.g., a text prompt). This 
reformulation shows that the TrigFlow score network can be expressed as a linear combination of the noisy latent and the predicted velocity field, weighted by time-dependent coefficients that are derived analytically from the trigonometric schedule.
 
\subsection{SANA-Sprint Reparameterization}
\label{subsec:sana_sprint}
 
SANA-Sprint builds upon TrigFlow by introducing a consistency-style denoising function $f_\theta$ that directly maps a noisy latent $X_t$ to a clean estimate $\hat{X}_0$ in a single step. Under the TrigFlow time parameterization, this function takes the form:
\begin{equation}
    f_\theta(X_t, t) = \cos(t_{cm}) \cdot X_t 
    - \sin(t_{cm}) \cdot \sigma_d \cdot F_\theta\!\left(\frac{X_t}{\sigma_d},\, t_{cm}\right),
    \label{eq:ftheta_sana}
\end{equation}
where $X_{cm} = X_t / \text{Scale}$ is the rescaled latent. Intuitively, 
Eq.~\eqref{eq:ftheta_sana} can be understood as a trigonometric 
decomposition: the $\cos(t_{cm})$ term retains the signal component of $X_t$, while the $\sin(t_{cm})$ term removes the noise component as predicted by $F_\theta$. This yields the denoised estimate:
\begin{equation}
    \hat{X}_0 = \cos(t_{cm})\, X_{t_{cm}} - \sin(t_{cm})\, \hat{F}_\theta,
    \label{eq:X0hat}
\end{equation}
which provides a clean, geometry-consistent reconstruction of the 
original sample from any noisy intermediate state.
 
\subsection{Adapting to Our IMPFM Framework with the Sufficient Statistic ($S$) Schedule}
\label{subsec:schedule_adaptation}
 
Our IMPFM framework uses a different noise schedule compared to TrigFlow. Specifically, we adopt the $S$ schedule, which defines a signal-to-noise ratio (SNR) function that differs from the trigonometric parameterization used in TrigFlow. To adapt the SANA-Sprint framework to our setting, we must establish a consistent mapping between the two 
schedules.
 
\paragraph{Setting up the Schedule Correspondence.}
Under the $S$ schedule, the signal and noise coefficients at timestep 
$t = t_{cm}$ are defined as:
\begin{equation}
    \alpha_t = \cos t = \sqrt{s}, 
    \qquad 
    \sigma_t = \sin t = \sqrt{1 - s},
    \label{eq:alpha_sigma}
\end{equation}
where $s \in [0, 1]$ is the schedule variable. This gives the forward noising process:
\begin{equation}
    X_t = \left(\alpha_t\, \hat{X}_0 + \sigma_t\, \frac{\varepsilon}{\sigma_d}\right) \cdot \sigma_d
        = \cos(t)\, \hat{X}_0 \cdot \sigma_d + \sin(t)\, \varepsilon,
    \label{eq:forward_process}
\end{equation}
which is a standard affine interpolation between the clean sample $\hat{X}_0$ and Gaussian noise $\varepsilon \sim \mathcal{N}(0, I)$, scaled by the latent normalization factor $\sigma_d$.
 
\paragraph{Defining the SNR Function.}
Under the $S$ schedule, the SNR function $g(s)$ and its transformed counterpart $\tilde{g}(\alpha_t)$ are defined as:
\begin{equation}
    g(s) = \frac{1}{s} - 1,
    \qquad
    \tilde{g}(\alpha_t) = \frac{1}{\alpha_t} - 1 
    = \frac{1}{\cos^2(t_{cm})} - 1,
    \label{eq:snr_functions}
\end{equation}
where $\tilde{g}(\alpha_t)$ is the SNR expressed as a function of the signal coefficient $\alpha_t = \cos(t_{cm})$. This function is monotonically 
decreasing in $t_{cm}$, consistent with the intuition that noise increases as the timestep progresses.
 
\paragraph{Deriving the Optimal Transition Timestep $r^* = r^*_{cm}$ for computing DDPM transition via Sufficient Statistic.}
A key step in adapting the noise schedule is determining the optimal 
intermediate timestep $r^*$ at which to perform the consistency transition 
between two noise levels $t_{cm}$ and $t_{cm'}$ (with $t_{cm'} < t_{cm}$, 
i.e., $t_{cm'}$ is closer to the clean data). The optimal $r^*$ is derived 
by matching the SNR levels under the $S$ schedule, yielding:
\begin{equation}
    \tilde{g}(r^*) = \frac{\tilde{g}(\cos t_{cm'}) \cdot \tilde{g}(\cos t_{cm})}
    {\tilde{g}(\cos t_{cm}) - \tilde{g}(\cos t_{cm'})},
    \label{eq:r_star_snr}
\end{equation}
which can be solved by applying the inverse of $g$ to both sides:
\begin{equation}
    \hat{r}^* = g^{-1}\!\left(
    \frac{\tilde{g}(\cos t_{cm'}) \cdot \tilde{g}(\cos t_{cm})}
    {\tilde{g}(\cos t_{cm}) - \tilde{g}(\cos t_{cm'})}
    \right).
    \label{eq:r_star}
\end{equation}
Eq.~\eqref{eq:r_star} gives the optimal SNR level $r^*$ in the $s$-domain. 
To recover the corresponding trigonometric timestep $r^*_{cm}$, we use 
the relation $\alpha_t = \cos(t_{cm}) = \sqrt{s}$, which gives:
\begin{equation}
    \cos(r^*_{cm}) = d\sqrt{r^*} 
    \quad \Longrightarrow \quad 
    r^*_{cm} = \arccos\!\left(\sqrt{r^*}\right).
    \label{eq:r_star_cm}
\end{equation}
This recovers the optimal intermediate timestep $r^*_{cm}$ in the trigonometric domain, which can be directly used within the TrigFlow/SANA-Sprint framework.
 
\paragraph{Computing the DDPM Transition via Sufficient Statistic}
Given the optimal transition timestep $r^*_{cm}$, the DDPM transition sample 
$X_{t_{cm'}}$ at the new timestep $t_{cm'}$ is obtained by applying 
the Sufficient Statistic scheme as defined in Equation 9 in the main paper. 

\section{Sufficient Statistic Computation}\label{sec:ss-comp}
Sufficient Statistic computation requires to evaluate $\bar{\gamma}$, $\rho$ and $\sigma$. Following~\citep{holderrieth2025glass}, it can be computed as
$ \quad \bar{\gamma} = \rho\sigma_{t^{child}}\sigma_t^{-1}, \quad \bar{\alpha} = \alpha_{t^{child}} - \bar{\gamma}\alpha_t, \quad \bar{\sigma}^2 = \sigma_{t^{child}}^2(1 - \rho^2)$. $\rho$ is the correlation parameter~\citep{holderrieth2025glass}.
\section{Extended Related Work}
\label{sec:ex-rel}

\paragraph{Fine-Tune Generative model with RL:}
Fine-tuning generative models with human feedback, such as user preferences, has become increasingly prevalent~\citep{ouyang2022training,touvron2023llama}.
However, these approaches typically assume a static reward model, treating rewards as fixed ground truth and not accommodating online queries. In contrast, we study an online, interactive setting in which target properties are initially unknown and are gradually revealed through sequential feedback, enabling continual target discovery rather than one-shot alignment to a fixed reward proxy.
\citep{dong2023raft} introduces a general online learning framework for aligning generative models, but it is not tailored to diffusion models. More recently,  ~\citep{uehara2024feedback} proposed a feedback-efficient online fine-tuning method for diffusion models; however, it learns a separate online reward model to guide sampling, which amplifies bias in the early stages of fine-tuning, and due to the KL-regularized fine-tuning objective, it remains vulnerable to mode collapse.
While tree-based methods~\cite{jain2025diffusion} provide a gradient-free alternative for search, they lack a mechanism for active trajectory correction. By relying strictly on prior dynamics, these samplers fail to leverage acquired feedback effectively, resulting in inefficient search under limited budget constraints. 

\section{Background: Feynman-Kac Correctors}
\label{sec:fkc-back}

Feynman-Kac Corrector (FKC) framework has been proposed to effectively sample from modified target distributions at inference time---such as reward-tilted distributions, annealed densities, or geometric averages~\citep{skreta2025feynman}. In standard diffusion models~\citep{song2020score,song2020denoising}, the time-evolution of the sample density $p_t(x)$ under the reverse-time stochastic differential equation (SDE) is governed by the Fokker-Planck partial differential equation (PDE):
\begin{equation}
    \frac{\partial p_t(x)}{\partial t} = -\langle \nabla, p_t(x) v_t(x) \rangle + \frac{\sigma_t^2}{2} \Delta p_t(x)
\end{equation}
where $v_t(x) = -f_t(x) + \sigma_t^2 \nabla \log q_t(x)$ represents the reverse-time drift. However, heuristically modifying this drift to target new compositional distributions fails to track the prescribed intermediate marginals. 

The FKC method corrects this by formulating the exact time-evolution of the target distribution as a Feynman-Kac PDE. This extends the standard Fokker-Planck equation by introducing a spatial reweighting function, $\bar{g}_t(x) = g_t(x) - \int g_t(x) p_t(x) dx$, which rigorously quantifies the mismatch between the simulated heuristic and the true target marginals. The combined evolution is described by:
\begin{equation}
    \frac{\partial p_t^{\text{FK}}(x)}{\partial t} = -\langle \nabla, p_t^{\text{FK}}(x) v_t(x) \rangle + \frac{\sigma_t^2}{2} \Delta p_t^{\text{FK}}(x) + \bar{g}_t(x) p_t^{\text{FK}}(x)
\end{equation}
Consequently, the sampling process is cast as a weighted SDE, where a particle $x_t$ and its accumulated log-weight $w_t$ evolve according to:
\begin{equation}
    dx_t = v_t(x_t)dt + \sigma_t dW_t, \quad dw_t = \bar{g}_t(x_t)dt
\end{equation}

Crucially, this mathematical framework seamlessly extends to reward-tilted sampling, where the objective is to generate samples from $p_t^{\text{reward}}(x) \propto q_t(x)\exp(\beta_t r(x))$ for a differentiable reward function $r(x)$ and inverse temperature $\beta_t$. Under the FKC formulation, the drift is augmented by the reward gradient to proactively steer the simulation, and the reweighting function accumulates mass based on the alignment between the diffusion vector field and the reward gradient. The resulting weighted SDE for reward-tilted sampling is exactly specified as:
\begin{equation}
    dx_t = \left(-f_t(x_t) + \sigma_t^2 \nabla \log q_t(x_t) + \beta_t \frac{\sigma_t^2}{2} \nabla r(x_t)\right)dt + \sigma_t dW_t
\end{equation}
\begin{equation}
    dw_t = \left[ \frac{\partial \beta_t}{\partial t} r(x_t) + \left\langle \beta_t \nabla r(x_t), \frac{\sigma_t^2}{2} \nabla \log q_t(x_t) - f_t(x_t) \right\rangle \right] dt
\end{equation}
To efficiently simulate these weighted equations and prevent weight degeneracy, the FKC framework employs Sequential Monte Carlo (SMC) resampling. By simulating a batch of non-interacting particles and periodically resampling them proportional to their exponentiated weights $\exp(w_t)$, the population dynamically corrects its distribution to faithfully match $p_t^{\text{reward}}(x)$ throughout the reverse trajectory.

\section{Additional Visualizations on the Alignment Task}\label{sec:align-add-a}
\begin{figure*}[h]
\centering
\newcommand{\methodrowpanel}[2]{%
  \begin{minipage}[t]{0.49\textwidth}
    \centering
    \includegraphics[width=\linewidth,height=0.205\textheight,keepaspectratio]{#1}\par
  \end{minipage}%
}
\noindent\rule{\textwidth}{0.6pt}
\vspace{0.2ex}
\begin{center}
\setlength{\tabcolsep}{2pt}%
\renewcommand{\arraystretch}{1.0}%
\begin{tabular}{@{}c@{\hspace{1pt}}c@{\hspace{4pt}\vrule width 0.6pt\hspace{4pt}}c@{}}
  & \small \textcolor{purple}{\shortstack{A crystal tree shimmering under a twilit,\\ starry sky.}}
  & \small \textcolor{purple}{\shortstack{A man shaping clay on a wheel\\ in a cluttered workshop.}} \\[0.2ex]
  \raisebox{9.5ex}{\textbf{\scriptsize FKS}}
  & \methodrowpanel{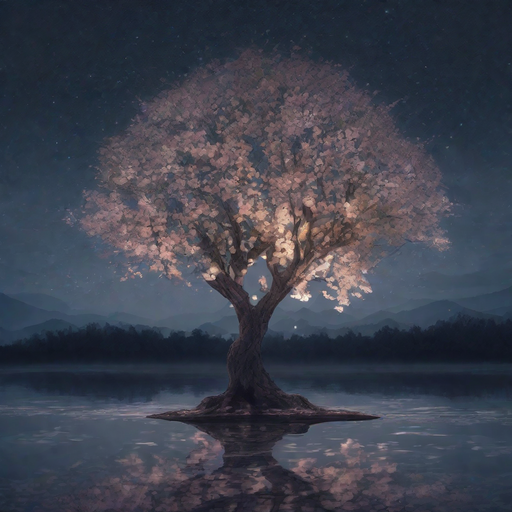}{}
  & \methodrowpanel{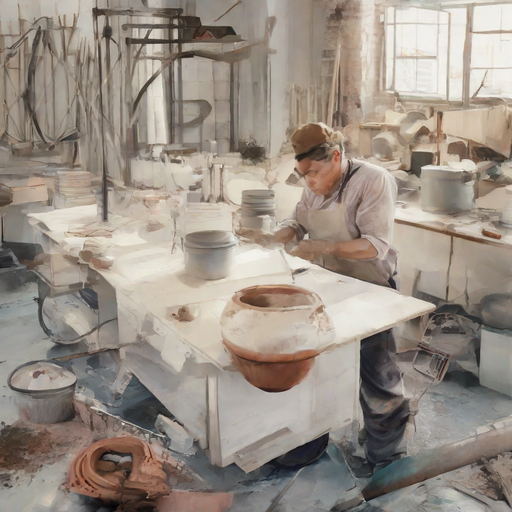}{} \\[0.25ex]
  \raisebox{9.5ex}{\textbf{\scriptsize IMPFM}}
  & \methodrowpanel{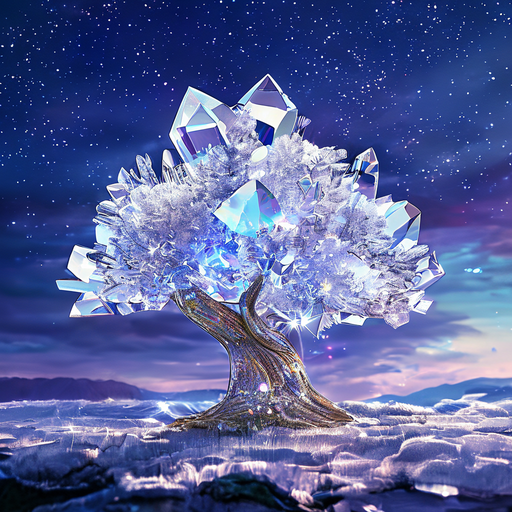}{}
  & \methodrowpanel{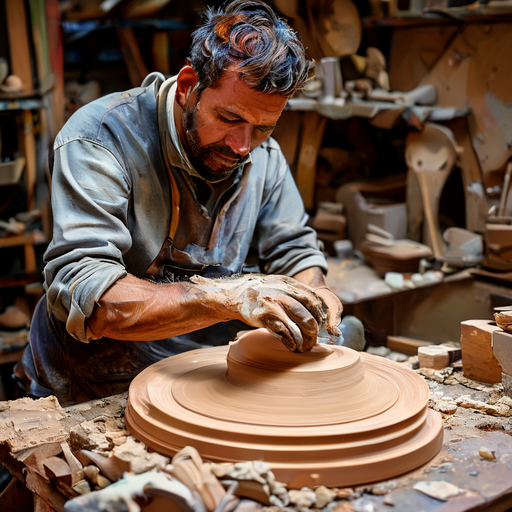}{} \\[0.35ex]
  \multicolumn{3}{@{}c@{}}{\rule{\textwidth}{0.8pt}} \\[0.35ex]
  & \small \textcolor{purple}{\shortstack{A dragon perched majestically on a craggy,\\ smoke-wreathed mountain.}}
  & \small \textcolor{purple}{\shortstack{A ghostly ship sailing on a fog-shrouded,\\ moonlit sea.}} \\[0.2ex]
  \raisebox{9.5ex}{\textbf{\scriptsize FKS}}
  & \methodrowpanel{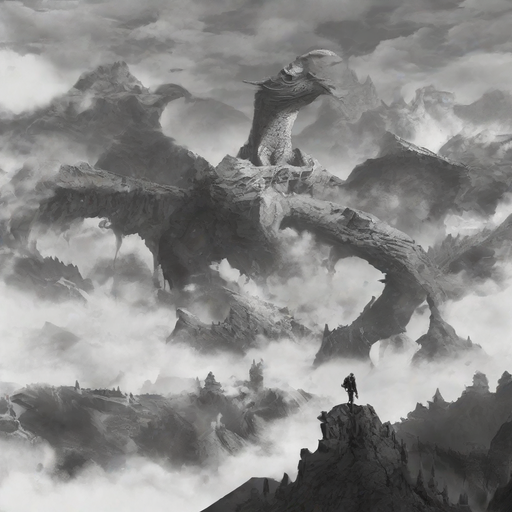}{}
  & \methodrowpanel{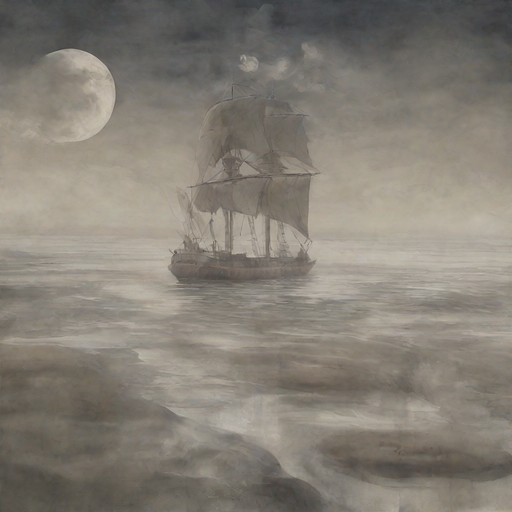}{} \\[0.25ex]
  \raisebox{9.5ex}{\textbf{\scriptsize IMPFM}}
  & \methodrowpanel{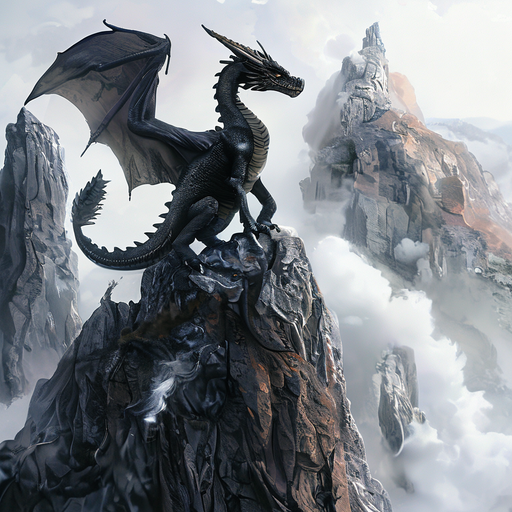}{}
  & \methodrowpanel{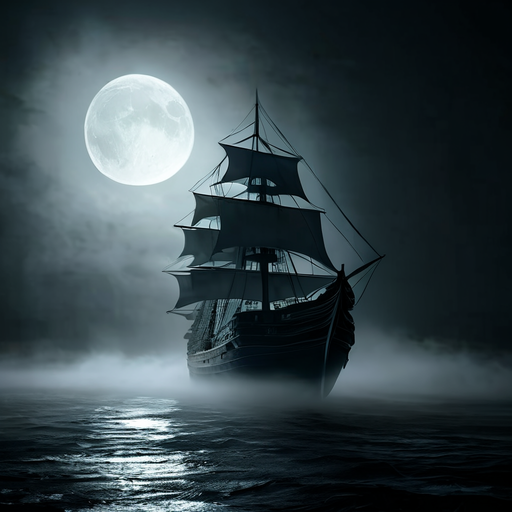}{}
\end{tabular}
\end{center}
\vspace{-0.2ex}
\noindent\rule{\textwidth}{0.6pt}
\caption{\small{Additional Alignment Visualization. }}
\label{fig:align-visu3}
\end{figure*}

\begin{figure*}[h]
\centering
\newcommand{\methodrowpanel}[2]{%
  \begin{minipage}[t]{0.49\textwidth}
    \centering
    \includegraphics[width=\linewidth,height=0.165\textheight,keepaspectratio]{#1}\par
  \end{minipage}%
}
\noindent\rule{\textwidth}{0.6pt}
\vspace{0.2ex}
\begin{center}
\setlength{\tabcolsep}{2pt}%
\renewcommand{\arraystretch}{1.0}%
\begin{tabular}{@{}c@{\hspace{1pt}}c@{\hspace{4pt}\vrule width 0.6pt\hspace{4pt}}c@{}}
  & \small \textcolor{purple}{\shortstack{A sorcerer’s hat casting shadows over a\\ cluttered, enchanted desk.}}
  & \small \textcolor{purple}{Five boys.} \\[0.2ex]
  \raisebox{9.5ex}{\textbf{\scriptsize FKS}}
  & \methodrowpanel{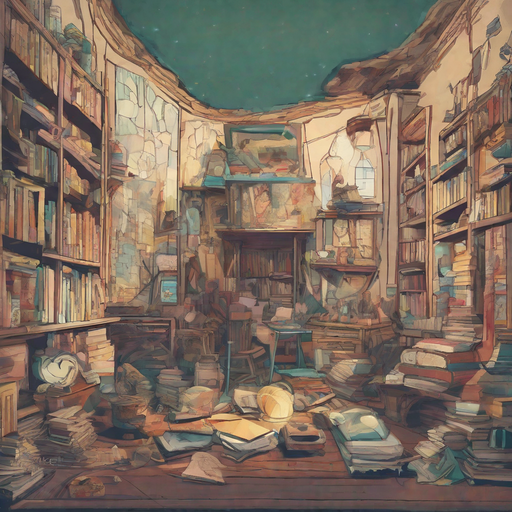}{}
  & \methodrowpanel{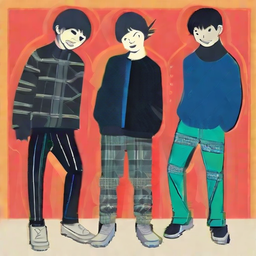}{} \\[0.25ex]
  \raisebox{9.5ex}{\textbf{\scriptsize IMPFM}}
  & \methodrowpanel{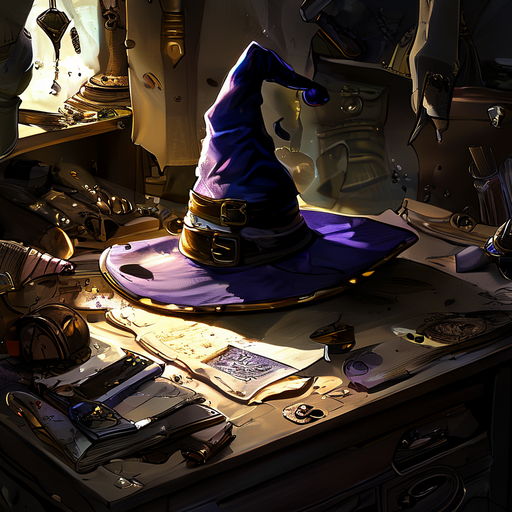}{}
  & \methodrowpanel{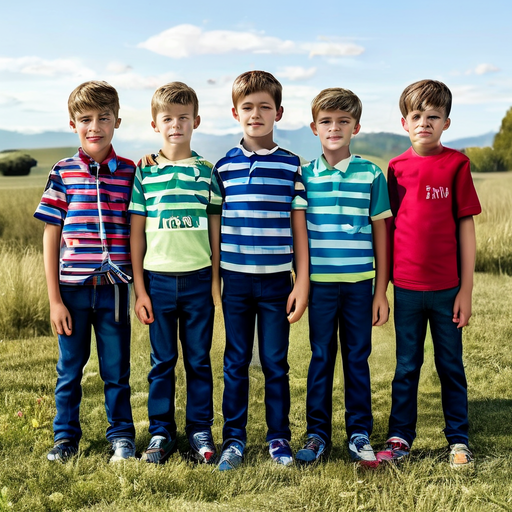}{}
\end{tabular}
\end{center}
\vspace{-0.2ex}
\noindent\rule{\textwidth}{0.6pt}
\caption{\small{Additional Alignment Visualization. }}
\label{fig:align-visunew}
\end{figure*}

\begin{figure*}[h]
\centering
\newcommand{\methodrowpanel}[2]{%
  \begin{minipage}[t]{0.49\textwidth}
    \centering
    \if\relax\detokenize{#2}\relax\else{\scriptsize\itshape ``#2''\par}\vspace{0.15ex}\fi
    \includegraphics[width=\linewidth,height=0.195\textheight,keepaspectratio]{#1}\par
  \end{minipage}%
}
\noindent\rule{\textwidth}{0.6pt}
\vspace{0.2ex}
\begin{center}
\setlength{\tabcolsep}{2pt}%
\renewcommand{\arraystretch}{1.0}%
\begin{tabular}{@{}c@{\hspace{1pt}}c@{\hspace{1pt}}c@{}}
  & \small \textcolor{purple}{An ice castle standing proudly in the midst of a blizzard.}
  & \small \textcolor{purple}{A lantern casting dim light in a haunted forest.} \\[0.2ex]
  \raisebox{9.5ex}{\textbf{\scriptsize IMPFM}}
  & \methodrowpanel{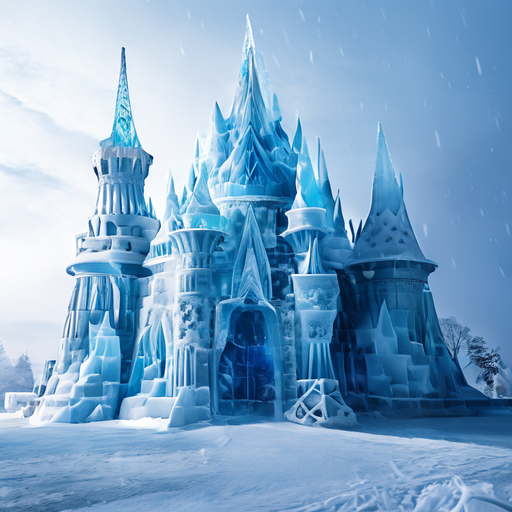}{}
  & \methodrowpanel{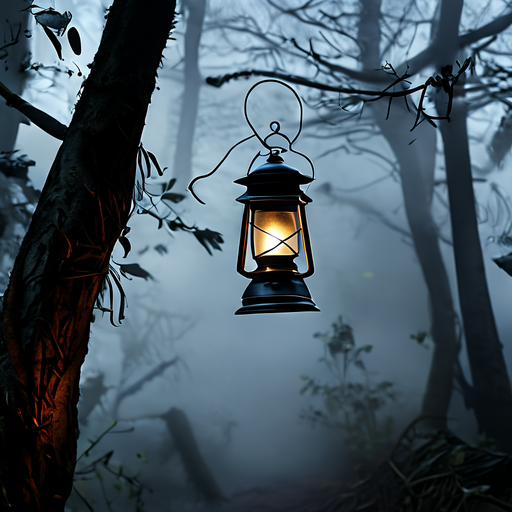}{} \\[0.25ex]
  & \small \textcolor{purple}{A small cat waves a wand.}
  & \small \textcolor{purple}{A phoenix soaring above a city, aglow with golden flames.} \\[0.2ex]
  \raisebox{9.5ex}{\textbf{\scriptsize IMPFM}}
  & \methodrowpanel{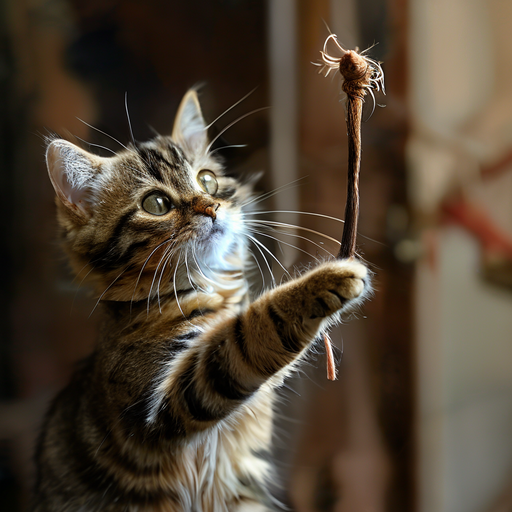}{}
  & \methodrowpanel{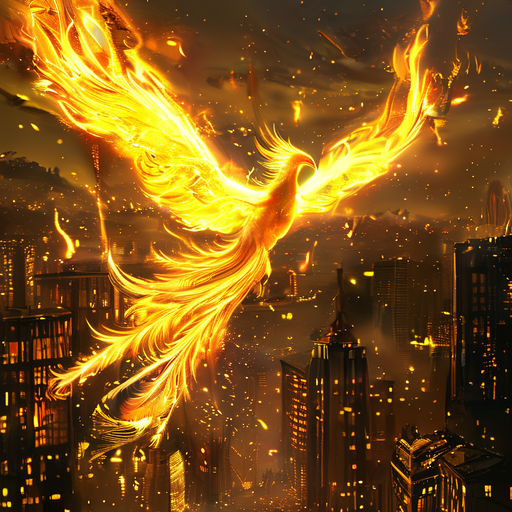}{}
\end{tabular}
\end{center}
\vspace{-0.2ex}
\noindent\rule{\textwidth}{0.6pt}
\caption{\small{Additional Alignment Visualization. }}
\label{fig:align-visu4}
\end{figure*}

In Fig.~\ref{fig:align-visu3},~\ref{fig:align-visunew},~\ref{fig:align-visu4}, we provide additional comparative visualizations for the quantitative alignment tasks. These results underscore the effectiveness of IMPFM in complex, feedback-efficient alignment tasks that demand robust reasoning capabilities. 
\begin{figure*}[!h]
\centering
\newcommand{\triplepanel}[3]{%
  \setlength{\tabcolsep}{2pt}%
  \begin{tabular}{@{}c@{\hspace{1pt}}c@{\hspace{1pt}}c@{}}
    \includegraphics[width=0.305\linewidth]{#1} &
    \includegraphics[width=0.305\linewidth]{#2} &
    \includegraphics[width=0.305\linewidth]{#3}
  \end{tabular}%
}
\newcommand{\panelblock}[4]{%
  \begin{minipage}[t]{\linewidth}
    \centering
    \triplepanel{#2}{#3}{#4}
    \par\vspace{0.4ex}
    {\itshape ``#1''\par}
  \end{minipage}%
}
\newcommand{\topheaders}{%
  {\fontsize{9.5}{10.5}\selectfont
  \begin{tabular}{@{}c@{\hspace{1pt}}c@{\hspace{1pt}}c@{}}
    \makebox[0.305\linewidth][c]{\textbf{FKS}} &
    \makebox[0.305\linewidth][c]{\textbf{MFM}} &
    \makebox[0.305\linewidth][c]{\textbf{IMPFM}}
  \end{tabular}%
  }
}
\noindent
\begin{minipage}[t]{0.48\textwidth}
\centering
\topheaders
\end{minipage}\hfill%
\begin{minipage}[t]{0.48\textwidth}
\centering
\topheaders
\end{minipage}

\begin{minipage}[t]{0.48\textwidth}
\centering
\panelblock{Two people and two helicopters.}{figs/q1_quant.png}{figs/mfm_q1.png}{figs/ours_q1.png}
\end{minipage}\hfill%
\begin{minipage}[t]{0.48\textwidth}
\centering
\panelblock{A baker pulling freshly baked bread out of an oven in a bakery.}{figs/fks_1.png}{figs/mfm_1.png}{figs/ours_1.png}
\end{minipage}%
\vspace{0.6ex}

\begin{minipage}[t]{0.48\textwidth}
\centering
\panelblock{Six bowls.}{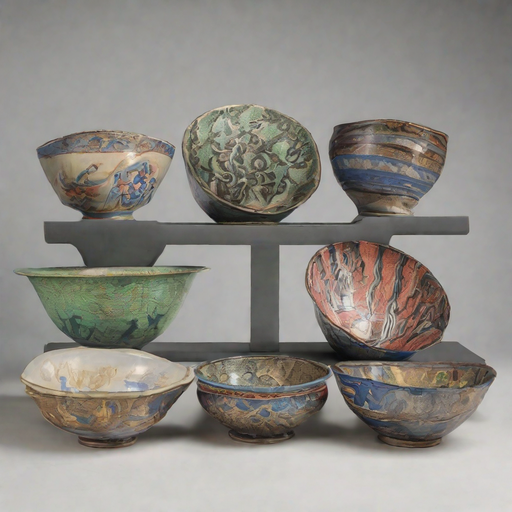}{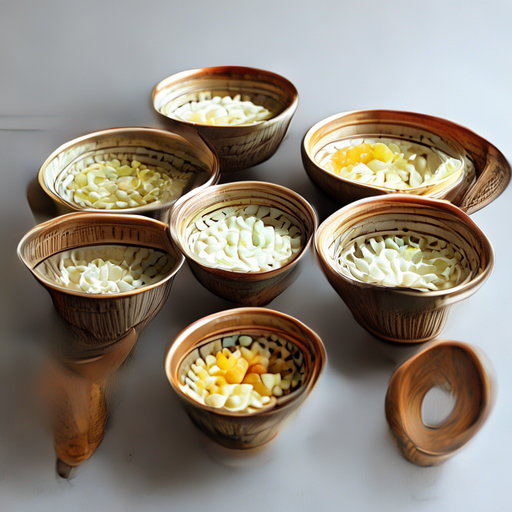}{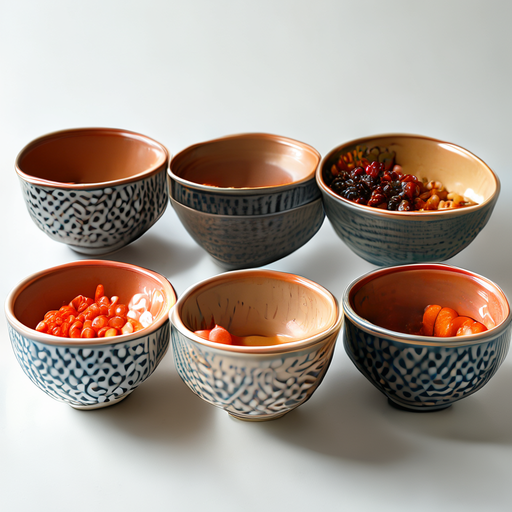}
\end{minipage}\hfill%
\begin{minipage}[t]{0.48\textwidth}
\centering
\panelblock{A gardener tending to flowers in a greenhouse filled with sunlight.}{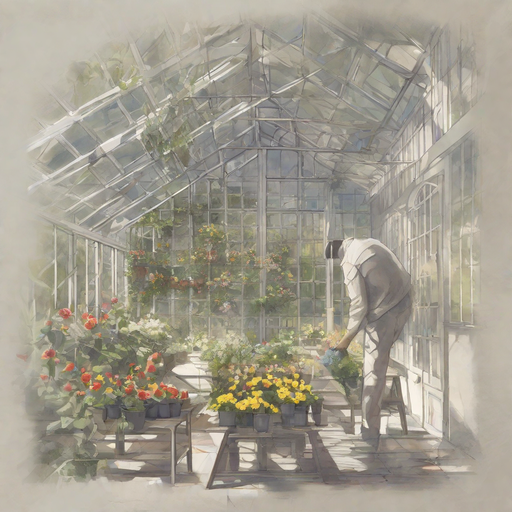}{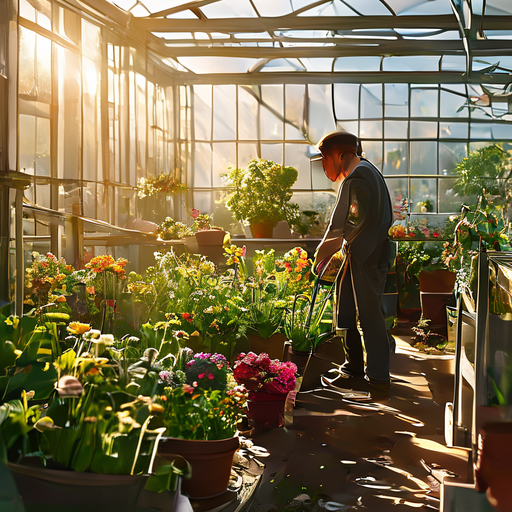}{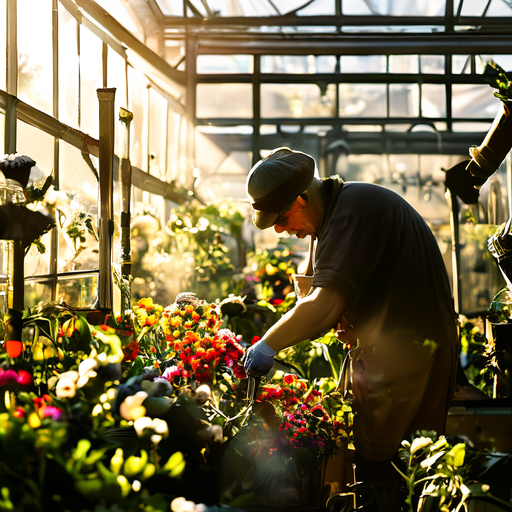}
\end{minipage}

\noindent\rule{\textwidth}{0.6pt}
\caption{\small{Alignment Visualization. (left) Quantity Prompt. (right) Compositional Prompt.}}
\label{fig:align-visu}
\vspace{-4pt}
\end{figure*}


\end{document}